%% file: acl_latex.tex
\pdfoutput=1
\documentclass[11pt]{article}

\usepackage[final]{acl}

\usepackage{times}
\usepackage{latexsym}

\usepackage[T1]{fontenc}
\usepackage[utf8]{inputenc}

\usepackage{microtype}

\usepackage{inconsolata}

\usepackage{graphicx}
\usepackage{dsfont}
\usepackage{pifont}
\usepackage{enumitem}
\usepackage{rotating}

\usepackage{booktabs}
\usepackage{multirow}
\usepackage{subcaption}
\usepackage{tcolorbox}
\usepackage{arydshln}
\usepackage{graphicx}
\usepackage{tabularx} 

\usepackage{amsthm}
\usepackage{amsmath}
\usepackage{amssymb}
\usepackage{xspace}
\newcommand{\narrowtextsc}[1]{\textls[-50]{\textsc{#1}}}
\newcommand{\lm}[1]{\texttt{#1}}
\newcommand{\sys}[1]{\narrowtextsc{#1}}
\newcommand{\data}[1]{\textsf{#1}}

\usepackage{xcolor,colortbl}

\definecolor{Gray}{gray}{0.94}
\definecolor{LightCyan}{rgb}{0.88,1,1}
\newcolumntype{a}{>{\columncolor{Gray}}c}
\newcolumntype{o}{>{\columncolor{white}}c}
\usepackage{soul}
\definecolor{celeste}{cmyk}{0.3922, 0.0353, 0, 0.1}
\definecolor{purple}{cmyk}{0.16, 0.28, 0, 0}
\definecolor{brilliantlavender}{cmyk}{0, 0.2235, 0, 0.1}
\definecolor{LightRed}{RGB}{232, 56, 107} 
\definecolor{LightBlue}{RGB}{116, 232, 226}
\definecolor{Tan}{rgb}{0.8203,0.7031,0.5469}
\definecolor{gblue}{RGB}{81,231,195}

\definecolor{greenblue}{RGB}{142, 207,201}

\definecolor{orange}{RGB}{255, 190, 122}

\definecolor{red}{RGB}{250, 127,111}

\definecolor{blue}{RGB}{224,230,255}

\usepackage{verbatim}

\title{Can Large Language Models Still Explain Themselves? Investigating the Impact of Quantization on Self-Explanations}
\newcommand{\affilsup}[1]{\rlap{\textsuperscript{\normalfont#1}}}

\author{
    Qianli Wang\affilsup{1,5}
    \qquad 
    Nils Feldhus\affilsup{2,5,\footnotemark[1],\footnotemark[2]}
    \qquad
    \textbf{Pepa Atanasova}\affilsup{3,\footnotemark[1]}
    \qquad
    \textbf{Fedor Splitt\affilsup{1}}
    \\
    \textbf{Simon Ostermann\affilsup{4,5,6}}
    \qquad
    \textbf{Sebastian M\"oller\affilsup{1,5}}
    \qquad
    \textbf{Vera Schmitt\affilsup{1}}
    \\
        $^1$Technische Universit\"at Berlin
        \quad
        $^2$University of Groningen
        \quad
        $^3$University of Copenhagen
        \\
        $^4$Saarland Informatics Campus
        \quad
        $^5$German Research Center for Artificial Intelligence (DFKI)
        \\
        $^6$Centre for European Research in Trusted AI (CERTAIN)
    \\
    \footnotesize{\textbf{Correspondence}: \texttt{\href{mailto:qianli.wang@tu-berlin.de}{qianli.wang@tu-berlin.de}}}
}


\begin{document}
\maketitle

\renewcommand{\thefootnote}{\fnsymbol{footnote}}
\footnotetext[1]{Equal contribution and shared second authorship.}
\footnotetext[2]{Work done at TU Berlin.}
\renewcommand*{\thefootnote}{\arabic{footnote}}

\begin{abstract}
% Quantization methods are widely used to accelerate inference and streamline the deployment of large language models (LLMs). While prior research has extensively investigated how quantization affects various LLM capabilities, its effects on self-explanations (SEs) remain unexplored. SEs generated by LLMs to justify their own outputs, are crucial for comprehending model decision-making. Thus, investigating the extent to which this capability is preserved under quantization is imperative. 
Quantization is widely used to accelerate inference and streamline the deployment of large language models (LLMs), yet its effects on self-explanations (SEs) remain unexplored. SEs, generated by LLMs to justify their own outputs, require reasoning about the model's own decision-making process, a capability that may exhibit particular sensitivity to quantization. As SEs are increasingly relied upon for transparency in high-stakes applications, understanding whether and to what extent quantization degrades SE quality and faithfulness is critical.
To address this gap, we examine two types of SEs: \textit{natural language explanations} (NLEs) and \textit{counterfactual examples}, generated by LLMs quantized using three common techniques at distinct bit widths. %We assess the resulting SEs' quality through automated metrics and a user study, while also examining SE faithfulness.
Our findings indicate that quantization typically leads to moderate declines in both SE quality (up to 4.4\%) and faithfulness (up to 3.9\%). The user study further demonstrates that quantization considerably diminishes both the coherence and trustworthiness of SEs (by up to 8.5\%). %Surprisingly, models with more aggressive bit reduction occasionally maintain their SE capabilities more effectively than their full-precision counterparts. 
Compared to smaller models, larger models show limited resilience to quantization in terms of SE quality but maintain more faithfulness. Moreover, no quantization technique consistently excels across task accuracy, SE quality, and faithfulness. Because quantization's impact varies considerably by context and can be sizable in specific cases, we recommend validating SE quality for the intended use case. Despite these sometimes considerable drops, quantization remains an effective compression technique when its impact on SEs is properly validated. \looseness=-1

\end{abstract}

% ------------------------------------------
\begin{figure*}[t!]
\centering
\resizebox{\textwidth}{!}{
\begin{minipage}{\columnwidth}
\includegraphics[width=\columnwidth]{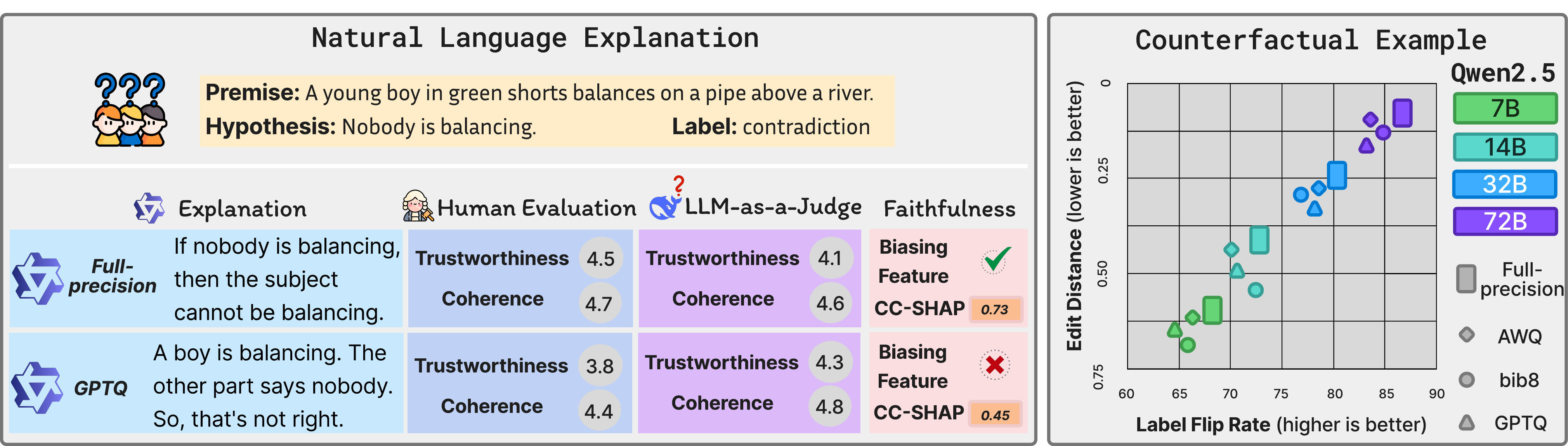}
\end{minipage}
}
\caption{Example effects of quantization on two types of self-explanations, \textit{natural language explanations} (\textbf{left}) and \textit{counterfactual examples} (\textbf{right}), across different models and quantization methods.
Natural language explanations are rated by \ding{192} human annotators in terms of perceived trustworthiness and coherence (\S\ref{subsec:user_study_results}); \ding{193} LLM-as-a-Judge evaluation (\S\ref{subsub:discussion}); \ding{194} faithfulness test by Biasing Feature \cite{turpin2023language} and CC-SHAP \cite{parcalabescu-frank-2024-measuring} (\S\ref{subsubsec:nle}), while counterfactual examples are evaluated on edit distance and label flip rate (\S\ref{subsubsec:res_cfe}).
}
\label{fig:example}
% \vspace{-1em}
\end{figure*}
% ------------------------------------------

\section{Introduction}
% \cite{gholami2022survey, treviso-etal-2023-efficient, zhu-2024-surveymodelcompressionlarge, wan2024efficient, liu-etal-2024-unlocking-data,}
% fairness \cite{ramesh-etal-2023-comparative},
% , zhu-2024-surveymodelcompressionlarge
% , mathematical reasoning \cite{liu2025quantizationhurtsreasoningempirical}
% With the shift towards LLMs and the focus on inference acceleration enabling more accessible and efficient deployment of LLMs, numerous quantization techniques have been developed \cite{dettmers-etal-2022-8bits, frantar-etal-2023-optq, xiao-etal-2023-smooth, lin-etal-2024-awq}. By reducing the precision of a model's parameters and lowering the number of bits used, quantization allows us to decrease model size while mostly preserving its performance \cite{gray-etal-1998-quantization}. To date, the effects of quantization are evaluated across various aspects, including multilinguality \cite{marchisio-etal-2024-quantization}, bias \cite{goncalves-strubell-2023-understanding}, and alignment \cite{jin-etal-2024-comprehensive}. 
% , xiao-etal-2023-smooth,

Deploying LLMs efficiently at scale has motivated extensive research on quantization~\cite{dettmers-etal-2022-8bits, frantar-etal-2023-optq, lin-etal-2024-awq, wang-etal-2026-compressed}. Quantization achieves model compression and efficient deployment (e.g., of on-device LLMs) by reducing parameter precision and bit allocation, delivering substantial size reductions while preserving most functionality \cite{gray-etal-1998-quantization}. Previous work has investigated quantization's influence on various model dimensions, such as multilinguality \cite{marchisio-etal-2024-quantization}, bias \cite{goncalves-strubell-2023-understanding}, and alignment \cite{jin-etal-2024-comprehensive}. An important capability dimension that may be affected by quantization is the capability of a model to explain itself. %Another important model capability is the generation of self-explanations (SEs) that justify its decisions 
Self-explanations (SEs) are statements generated by models to justify their own decisions \cite{agarwal2024faithfulnessvsplausibilityunreliability, madsen-etal-2024-self, wang2026macroenhancingmultilingualcounterfactual}, which are deemed to be an effective and convincing way to deliver explanations to users and enhance the transparency of black-box LLMs~\cite{huang2023largelanguagemodelsexplain, randl-etal-2024-self}. Nevertheless, SE may obfuscate the true reasoning process of LLMs \cite{turpin2023language, tutek-etal-2025-measuring, sun2026investigatinginterplaycontextualparametric}, and we hypothesize that quantization may exacerbate this, since LLMs are directly optimized for task performance but learn to generate faithful SEs more indirectly. %There exist two main types of SEs: natural language explanations \cite{wang-etal-2024-llmcheckup, liu-etal-2024-emergent} and counterfactual examples \cite{bhattacharjee-etal-2024-zero, giorgi2025natural}. 
Moreover, quantized models have been widely adopted in prior work for many types of SE generation \cite{wang-etal-2024-llmcheckup, liu-etal-2024-emergent, bhattacharjee-etal-2024-zero, giorgi2025natural}. However, the impact of quantization on SEs, specifically on whether SEs remain faithful to a model's inner workings and whether their quality can be largely preserved, remains unexplored and has yet to be comprehensively characterized. %\looseness=-1

We bridge this gap through a comprehensive study on how quantization affects both the \textit{quality} and \textit{faithfulness} of SEs. Our study encompasses two distinct types of free-text SEs: \textit{natural language explanations} and \textit{counterfactual examples} (Figure~\ref{fig:example}). \textbf{First}, we perform comprehensive automatic evaluations of SE quality across five datasets and six models of varying sizes under full precision and three quantization approaches with different bit widths (\S\ref{subsec:results_explainability}). We show that different types of SEs exhibit varying levels of sensitivity to quantization, though \textit{quantization generally leads to moderate degradation in SE quality} (up to 4.4\%). Unexpectedly, larger quantized models do not always outperform smaller full-precision models in generating high-quality SEs, nor do LLMs with lower bit precision consistently lag behind their higher bit-precision counterparts.

% \hl{findings to be updated!} We find that quantization leads to information loss within the model, and this trend becomes more pronounced in smaller models. Moreover, larger quantized models do not consistently outperform smaller full-precision models and quantization-induced degradation on explanation quality depends on the specific quantization techniques and models deployed. Generally, integer quantization \texttt{bib8} yields superior explanation quality compared to other quantization techniques across the evaluated methods. Surprisingly, lower-precision LLMs can sometimes surpass their higher-precision counterparts and a quantized model may even generate better explanations than the same model in full precision. 

\textbf{Second}, we assess SE faithfulness through self-consistency checks for counterfactuals and two metrics for natural language explanations. SE faithfulness exhibits \textit{modest average decline under quantization} (up to 3.9\%); however, larger models demonstrate greater robustness in preserving SE faithfulness (\S\ref{subsubsec:nle}). Across our experiments, no quantization method consistently excels across \textit{task performance}, \textit{explanation quality}, and \textit{faithfulness} simultaneously. Furthermore, we observe a distinct trade-off between SE characteristics, namely quality and faithfulness, and overall task performance under quantization. Therefore, SE characteristics should be validated for specific use cases depending on whether task performance or explanation performance is prioritized (\S\ref{subsub:discussion}).

\textbf{Lastly}, we conduct a user study with 48 participants, evaluating trustworthiness and coherence of SEs generated by models at different bit-precision levels. Human evaluators perceive \textit{full-precision models as producing more trustworthy and coherent SEs than those produced by their quantized counterparts} (Figure~\ref{fig:example}). Notably, conducting a similar LLM-as-a-Judge evaluation fails to fully capture the impact of quantization on self-explanation quality, evidenced by the weak or negative, and non-statistically significant correlation observed between human and judge model ratings (\S\ref{subsec:user_study_results}). %Notably, automatic metrics may fail to fully capture the impact of quantization on explanation quality, as evidenced by the discrepancies with human evaluation results (Table~\ref{tab:user_study}).

\textbf{In conclusion}, the modest reductions in SE quality and faithfulness do not diminish quantization's value as an effective model compression strategy. Nevertheless, for high-stakes scenarios requiring optimal explanation reliability, we recommend application-specific validation before deployment. \looseness=-1

\section{Preliminaries and Related Work}
\label{sec:background}
\paragraph{Quantization.} 
Quantization has been demonstrated to offer a superior balance between compression and accuracy \cite{li2024evaluating}. The decoding stage during LLM inference is typically memory-bound, where the key-value cache (KV cache) overhead often exceeds the size of the model weights \cite{li2024evaluating}. Quantization techniques compress LLMs by converting model weights, activations, or the KV cache, originally in 32-bit floating-point format, into lower-precision data types \cite{zhu-2024-surveymodelcompressionlarge}, e.g., 8-bit integer \cite{dettmers-etal-2022-8bits}. These techniques can be broadly categorized into two types: quantization-aware training (QAT) and post-training quantization (PTQ). QAT requires retraining to mitigate errors introduced by quantization, whereas PTQ facilitates an ad-hoc quantization during inference %facilitates the direct application of a quantized model during inference. Thus, PTQ is especially appealing, as it offers computational efficiency 
without necessitating modifications to the model architecture or training process. Among PTQ, weight-only quantization is the most conventional and widely adopted method \cite{wan2024efficient, zhou2024surveyefficientinferencelarge}, which effectively accelerates matrix multiplications during the decoding stage \cite{li2024evaluating}. Thereby, in this paper, we evaluate the impact of weight-only PTQ quantization (\S\ref{subsec:quantization}) on self-explanations (\S\ref{sec:evaluation}).%), which eliminates the need for retraining to address errors resulting from quantization. \looseness=-1
 % PTQ addresses this by reducing the memory usage of weights, activations, and KV cache using low-precision values \cite{zhu-2024-surveymodelcompressionlarge}.

 \paragraph{Impact of Quantization.}
Recent work has extensively examined the impact of quantization on various capabilities of LLMs. \citet{marchisio-etal-2024-quantization} conduct a thorough analysis of quantized multilingual LLMs, focusing on performance degradation across languages. \citet{goncalves-strubell-2023-understanding, kirsten-etal-2024-bias} explore the emergence of bias in the outputs generated by quantized models. \citet{liu-etal-2024-emergent} find that in-context learning ability gradually declines in heavily quantized LLMs. \citet{jin-etal-2024-comprehensive} observe that models with 4-bit quantization can still retain the alignment ability. In our work, we explicitly explore the impact of quantization on self-explanations.\looseness=-1

\paragraph{Self-Explanations.} SEs are generated by LLMs to justify their own decisions. 
Prior work has identified several SE types, including prompting-based feature attribution explanations \cite{huang2023largelanguagemodelsexplain}, counterfactual explanations \cite{wang-etal-2025-fitcf}, redaction explanations \cite{madsen-etal-2024-self, doi2025investigatingtraininggeneralizationfaithful}, and natural language explanations \cite{villaarenas2024anchoredalignmentselfexplanationsenhancement}. %Among these, we investigate the two most widely adopted free-text SEs: natural language explanations and counterfactual explanations. 
Previous research has inspected various aspects of SEs: \citet{huang2023largelanguagemodelsexplain} compare prompting-based feature attribution with perturbation-based feature importance. \citet{madsen-etal-2024-self} explore the faithfulness of SEs via self-consistency. \citet{randl-etal-2024-self} examine how SEs correlate with human judgments and internal model dynamics. We extend this line of research by investigating the impact of quantization on SE quality and faithfulness. 

\section{Experimental Setup}
We evaluate the impact of quantization on SE by examining two representative \textbf{free-text} explanation types (\S\ref{subsec:methods}). Specifically, we compare full-precision LLMs (\S\ref{subsec:models}) with their quantized counterparts employing different quantization techniques and bit-widths (\S\ref{subsec:quantization}) across five datasets (\S\ref{subsec:datasets}).

\subsection{Self-Explanations}
\label{subsec:methods}
The experimental investigation focuses on two well-established types of SEs in the explainability literature \cite{madsen-etal-2024-self, agarwal2024faithfulnessvsplausibilityunreliability, villaarenas2024anchoredalignmentselfexplanationsenhancement, wang-etal-2025-cross, monteiro-paes-etal-2025-multi}: \textit{natural language explanations} and \textit{counterfactual examples} (Figure~\ref{fig:example}).

\textbf{Natural Language Explanations (NLEs)} are free-text explanations of predictions made by the model and can be easily understood by humans \cite{camburu-2018-esnli,wiegreffe-etal-2021-measuring}. To minimize the confounding effects of quantization that also affect in-context learning capabilities \cite{liu-etal-2024-emergent}, we generate NLEs in a zero-shot setting using \texttt{ZeroCoT} \cite{kojima-2022-zero-shot-reasoners} which elicits step-by-step reasoning from LLMs by simply adding ``Let's think step by step'' before each answer, without requiring any hand-crafted few-shot examples.
% with the predict-then-explain pattern \cite{kumar-talukdar-2020-nile}

\textbf{Counterfactual Examples (CFEs)} refer to minimally edited inputs that result in different model predictions \cite{miller-2019-explanation, madsen-2022-survey, wang-etal-2026-parallel, wang2026iflipiterativefeedbackdrivencounterfactual}, which can be used to understand the black-box nature of models in a contrastive manner \cite{wu-etal-2021-polyjuice, nguyen-etal-2024-llms}.
Analogous to NLEs, we generate CFEs in a zero-shot setting using \texttt{FIZLE} \cite{bhattacharjee-etal-2024-zero}. \texttt{FIZLE} employs a two-stage process: it begins with prompting LLMs to extract salient keywords from the input, which are then employed to guide the generation of counterfactual examples.

\subsection{Datasets}
\label{subsec:datasets}
Our study employs five widely recognized datasets\footnote{Dataset examples and label distributions are detailed in Appendix~\ref{app:dataset}. \data{eSNLI} is used for both SE types, \data{AG News} and \data{IMDb} for CFEs, and \data{HealthFC} and \data{ECQA} for NLEs.} to evaluate SEs (\S \ref{subsec:methods}). \data{eSNLI} \cite{camburu-2018-esnli} categorizes premise-hypothesis relationships as \textit{entailment}, \textit{contradiction}, or \textit{neutrality}, with human-annotated NLEs. \data{HealthFC} \cite{vladika-etal-2024-healthfc} is a fact-checking dataset with questions, documents, veracity labels (\textit{true}, \textit{false}, \textit{unknown}), and human explanations. \data{ECQA} \cite{aggarwal-etal-2021-explanations} extends \data{CommonsenseQA} \cite{talmor-etal-2019-commonsenseqa} with human-annotated explanations for multiple-choice answers. \data{AG News} \cite{zhang-2015-agnews} classifies news articles into four categories: \textit{World}, \textit{Sports}, \textit{Business}, and \textit{Sci/Tech}. \data{IMDb} \cite{maas-etal-2011-learning} is a binary sentiment classification benchmark of movie reviews.

\subsection{Quantization Techniques}
\label{subsec:quantization} Building on the prior discussion (\S\ref{sec:background}), we identify three commonly used PTQ techniques applied to the selected LLMs (Appendix~\ref{app:quantization}, Table~\ref{tab:used_model}):\looseness=-1
% [noitemsep,topsep=0pt,leftmargin=*]
\begin{itemize}
    \item \texttt{GPTQ} \cite{frantar-etal-2023-optq} uses a second-order, Hessian-based optimization to quantize weights post-training with minimal accuracy loss;
    \item \texttt{AWQ} \cite{lin-etal-2024-awq} enhances weight quantization by handling activation outliers to preserve model accuracy at low bit-widths; 
    \item Integer quantization \cite{dettmers-etal-2022-8bits} implemented by \sys{BitsAndBytes}\footnote{The implementation of integer quantization provided in \sys{BitsAndBytes} is limited to weight-only quantization: \url{https://github.com/bitsandbytes-foundation/bitsandbytes}.} (\texttt{bib4} and \texttt{bib8}) enables fast and memory-efficient inference by using optimized low-bit kernels. \looseness=-1
\end{itemize}

\subsection{Models}
\label{subsec:models}
We employ six open-source LLMs spanning model sizes from 7B to 72B, drawn from two families: \lm{Llama3} (8B, 70B) \cite{llama3modelcard} and \lm{Qwen2.5} (7B, 14B, 32B, 72B) \cite{qwen2024qwen25technicalreport} (Table~\ref{tab:counterfactual_and_explanation_automatic_evaluation}; Appendix~\ref{app:inference}), across all self-explanations. These models are selected because their corresponding quantized versions are (officially) provided. Additionally, for all experiments, we set the \texttt{temperature=0} during generation to minimize the stochastic nature of LLM outputs, ensuring the reproducibility of our results.

\section{Evaluation}
\label{sec:evaluation}
We assess the impact of quantization on self-explanations from three perspectives: explanation quality, evaluated through \ding{192} automatic evaluation (\S\ref{subsec:automatic_evaluation}), \ding{193} human evaluation (\S\ref{subsec:human_evaluation}), and \ding{194} explanation faithfulness (\S\ref{subsec:faithful}). 

% , and LLM-as-a-Judge evaluation (\S\ref{subsec:laaj})

\subsection{Self-Explanation Quality Evaluation}
\label{subsec:automatic_evaluation}

We assess the self-explanation quality using automatic metrics evaluating NLE plausibility (resemblance to human-annotated explanations; \S\ref{subsubec:nle}) and CFE performance across validity, fluency, and textual similarity (\S\ref{subsubsec:cfe}). 
All results are averaged over three runs with different seeds (\S\ref{subsec:results_explainability}).

\subsubsection{Natural Language Explanation}
\label{subsubec:nle}
Following \citet{marasovic-etal-2022-shot, wang-etal-2025-cross, hsu-etal-2025-free}, we employ two automatic metrics: 
% two based on similarity that assess NLE plausibility through comparison with human annotations, and one error analysis metric that identifies deficiencies in the generated NLEs. % are employed to assess the quality of generated NLEs: BERTScore, BARTScore, and TIGERScore. 

% \paragraph{Micro-$F_1$} Following the predict-then-explain pattern \cite{kumar-talukdar-2020-nile}, Micro-$F_1$ Score is used to assess task performance on the target dataset.

% \textbf{BERTScore} \cite{zhang2020BERTScore} measures the similarity between the human-annotated NLEs and LLM-generated NLEs by computing pairwise cosine similarities between their contextual embeddings from \lm{BERT} \cite{devlin-etal-2019-bert}.

\textbf{BARTScore} \cite{yuan-2021-bartscore} is a reference-based metric that employs \lm{BART} \cite{lewis-etal-2020-bart} to evaluate generated explanations based on how well they align with human-annotated references. Furthermore, BARTScore performs bidirectional evaluation, assessing both ``generated-to-reference'' and ``reference-to-generated'' directions, thereby offering a more robust assessment. BARTScore measures NLE plausibility based on textual similarity between human NLEs and LLM-generated NLEs. \looseness=-1

\textbf{TIGERScore} \cite{jiang-2024-tigerscore}, in contrast, is a reference-free metric that deploys a fine-tuned \lm{Llama2} \cite{touvron-2023-llama2openfoundation} model to identify errors in the generated explanations in terms of, e.g., coherence, informativeness, and accuracy. For each mistake, TIGERScore assigns a penalty score between $[-5, -0.5]$. High-quality explanations that contain no detected errors receive a score of 0.

\subsubsection{Counterfactual Example}
\label{subsubsec:cfe}
We evaluate the generated counterfactuals using three automated metrics widely adopted in the literature~\cite{ross-etal-2021-explaining, bhan-etal-2023-enhancing, nguyen-etal-2024-llms, wang-etal-2025-truth}.

\paragraph{Label Flipping Rate (LFR)} For a dataset $\mathcal{D}=\{(x_i,y_i)\}_{i=1}^N$ containing $N$ pairs of original inputs $x_i$ and gold labels $y_i$, LFR captures the frequency with which the generated counterfactual $\tilde{x}_i$ alters the original model prediction $\hat{y}_i$ on $x_i$ to a different one $\tilde{y}_i$ \cite{ge-2021-counterfactualevaluationexplainableai, bhattacharjee-etal-2024-zero, wang-etal-2025-fitcf}. The LFR is calculated as:
\begin{equation*}
    LFR = \frac{1}{N}\sum_{i=1}^{N} \mathbb{I} \big(\hat{y}_i \neq \tilde{y}_i\big)
\end{equation*}
where $\mathbb{I}$ is the indicator function, which returns 1 if the condition is true and 0 otherwise.

\paragraph{Perplexity (PPL)} represents the exponential of the average negative log-likelihood computed over a sequence. PPL is a commonly used metric in counterfactual evaluation literature to assess the fluency of generated counterfactuals by measuring the model's predictive accuracy for each word given its preceding context \cite{le2023cococounterfactuals,nguyen-etal-2024-ceval-benchmark}. For a given counterfactual $\tilde{x} = (t_1, t_2, \cdots, t_n)$ and a model parameterized by $\theta$, PPL is computed as follows:\looseness=-1
\begin{equation*}
    PPL(\tilde{x}) = \exp\left\{\frac{1}{n}\sum_{i=1}^{n} \log{p_{\theta}(t_i|t_{<i})}\right\}
\end{equation*}

\paragraph{Textual Similarity (TS)} As counterfactuals $\tilde{x}$ should closely resemble the original inputs $x$, we measure this similarity using the Levenshtein distance $d$ on token level, a standard metric in existing literature \cite{ross-etal-2021-explaining, treviso-etal-2023-crest}:
\begin{equation*}
    TS = \frac{1}{N} \sum_{i=1}^{N} \frac{d(x_i, \tilde{x}_i)}{|x_i|}
\end{equation*}

\subsection{Faithfulness}
\label{subsec:faithful}
In addition to assessing the impact of quantization on SE quality, we also evaluate the effect of quantization on the faithfulness of NLEs and CFEs. This evaluation determines whether the SEs generated by quantized LLMs are still able to truly
reflect their underlying reasoning process \cite{doshivelez2017rigorousscienceinterpretablemachine, jacovi-goldberg-2020-towards}. The faithfulness rate is defined as: 
\begin{equation*}
r_{\text{faith}} = \frac{N_{\text{faithful}}}{N}    
\end{equation*}
where $N$ is the number of evaluated explanations and $N_{\text{faith}}$ denotes the number of faithful explanations. This metric enables direct comparison of faithfulness of NLE and CFE across various quantization methods and bit-widths, with higher values indicating greater faithfulness.

\paragraph{Faithfulness of NLEs.} For NLE faithfulness evaluation, we employ two widely used faithfulness metrics: \textit{biasing features} and \textit{CC-SHAP}.\footnote{The description of the faithfulness metrics is in App.~\ref{app:faithfulness}.} \ding{192} %The counterfactual test determines explanations as unfaithful by inserting words into the original input and checking if predictions change despite these words not being mentioned in the explanation \cite{atanasova-etal-2023-faithfulness}. \ding{193} 
The biasing features metric marks explanations as unfaithful that do not reflect answer biases added in context examples %if suggested answers appended to the prompt influence the chain-of-thought reasoning and predictions 
\cite{turpin2023language}. \ding{193} CC-SHAP identifies unfaithful explanations when the model's input contribution distributions for prediction and reasoning diverge, with values ranging from -1 to +1 \cite{parcalabescu-frank-2024-measuring}. %\looseness=-1

\paragraph{Faithfulness of CFEs.} Following \citet{madsen-etal-2024-self}, we employ the self-consistency check to evaluate whether counterfactual predictions satisfy the targeted labels; counterfactuals meeting this criterion are deemed faithful. In this context, $r_{\text{faith}}$ corresponds to the label flip rate (\S\ref{subsubsec:cfe}) -- the ratio of valid and faithful counterfactuals to all instances
that successfully alter the model prediction to the target label.

% ----------- automatic evaluation ----------
\input{table/automatic_evaluation} 
% -------------------------------------------

\subsection{Human Evaluation}
\label{subsec:human_evaluation}
We further assess the effect of quantization on self-explanation quality (\S\ref{subsec:methods}) by conducting a user study in which participants subjectively evaluate NLEs and CFEs along two dimensions (\S\ref{subsubsec:subjective_ratings}). This analysis identifies explanation qualities that necessitate human judgment, which extend beyond the scope of automatic evaluation metrics (\S\ref{subsec:automatic_evaluation}).

% This investigation examines the extent to which self-explanation quality changes observed through automatic evaluation are consistent with human judgment.

\subsubsection{Subjective Ratings}
\label{subsubsec:subjective_ratings}
% tsai-etal-2024-leveraging,
Following the design of Likert scales for explanation evaluation proposed by \citet{feldhus-etal-2023-interrolang,chiang-lee-2023-large}, and user studies on NLEs and CFEs conducted by \citet{ domnich2024unifyingevaluationcounterfactualexplanations,wang-etal-2025-cross,shailya-etal-2025-trust}, we ask human annotators to evaluate NLEs and CFEs based on the following dimensions, each rated on a 5-point Likert scale ranging from "strongly disagree" (1) to "strongly agree" (5):

\begin{itemize}
    \item \textbf{Trustworthiness}: Evaluate whether the provided explanation is trustworthy and can be relied upon by humans;

    \item \textbf{Coherence}: Assess whether the provided explanation is sensible and clear, and effectively captures the rationale.
    
    % \item \textbf{Insightfulness}: Evaluate whether the explanation offers informative and meaningful insights.
\end{itemize}

\subsubsection{User Study Setup}
\label{subsubsec:setup}
To assess the quality of explanations produced by various LLMs and quantization techniques, we conduct a user study involving $N=48$ participants, who are all native English speakers without necessarily possessing expertise in explainability. 
We randomly sample ($k=30$) dataset indices. For each model-precision pair, the self-explanations generated by the corresponding model in full precision or quantized using different methods are evaluated by at least two human annotators. Our user study focuses on the overlapping dataset between NLEs and CFEs (\data{eSNLI}), and on \lm{Qwen2.5} models of sizes \{7B, 32B, 72B\} to capture a wide range of model scales. We exclude \lm{Qwen2.5-14B} due to its consistently suboptimal performance, even without quantization. Each annotator is assigned 15 explanations, accompanied by two evaluation dimensions (\S\ref{subsubsec:subjective_ratings}), and tasked with assigning appropriate scores based on a given Likert scale.\footnote{Further details about annotator recruitment and annotation guidelines can be found in Appendix~\ref{app:annotation_guideline}.} We report inter-annotator agreements with Krippendorff’s $\alpha$ of $0.71$ for NLEs and $0.64$ for CFEs. \looseness=-1

% \subsection{LLM-as-a-Judge}
% \label{subsec:laaj}
% The adoption of LLMs as evaluators for complex tasks, referred to as ``LLM-as-a-Judge'' (LaaJ), has gained popularity to perform evaluations by assigning corresponding scores \cite{zheng-etal-2023-llm, huang-etal-2024-chatgpt}. In addition to automatic and human evaluation, we investigate how well LLMs can quantitatively assess the explanation quality degradation caused by quantization. For this purpose, we select three open-source LLMs of varying sizes that are commonly used in the literature \cite{gu2025surveyllmasajudge}: \lm{prometheus-8x7b} \cite{kim2024prometheus}, \lm{Gemma3-27B} \cite{gemmateam2025gemma3technicalreport} and \lm{aya-expanse-32b} \cite{dang2024ayaexpansecombiningresearch}.

\section{Results}
\label{sec:results}
\subsection{Automatic Evaluation}
\label{subsec:results_explainability}

\subsubsection{Natural Language Explanations}
\label{subsubsec:nle}
\paragraph{Quality of NLEs.}
Tables~\ref{tab:counterfactual_and_explanation_automatic_evaluation} and \ref{tab:automatic_evaluation_additional} demonstrate that NLE quality reduction \textit{varies more substantially in smaller models but remains less affected in larger models}. Surprisingly, full-precision LLMs do not consistently outperform their quantized counterparts. Furthermore, LLMs with lower precision, e.g., those using \texttt{bib4}, occasionally generate higher-quality NLEs than LLMs with higher precision. In addition, Table~\ref{tab:quality_performance} reveals that, especially for larger models, quantization-induced task performance degradation generally does not contribute to NLE quality degradation, as indicated by weak or even negative correlations.

%the quality of NLEs generated by quantized LLMs deteriorates relative to full-precision LLMs, with the extent of reduction emerging nuanced variation. The quality of NLEs does not consistently improve with increasing model size, which aligns with the findings from \citet{marasovic-etal-2022-shot, chen-etal-2023-zara}. This may be ascribable to the fact that even larger-sized LLMs may experience performance plateaus in zero-shot scenarios \cite{ni2023evaluatingrobustnessinstructionslarge, chowdhery-etal-2024-palm}. Meanwhile, 

\paragraph{Faithfulness of NLEs.}
%\hl{(1) reliability/consistency of different approaches - spearson ranking correlation; (2) flip-flop: faithful to unfaithfulness dominant; (3) what's the best quantization method preserving the faithfulness? (Llama3-8B and Qwen2.5-14B are mostly affected) (4) How the model size affect it? (5) sometimes explanations are more faithful; (6) average drops; (7) confusion metrics for llama70b and qwen32b; add the details in appendix F}
Table~\ref{tab:faithfulness} reveals that quantization generally induces \textit{moderate declines} in NLE faithfulness across all faithfulness metrics (biasing features $\downarrow$ 3.4\%, CC-SHAP $\downarrow$0.05). %; as displayed in Table~\ref{tab:biasing-transitions}). 
Analysis of transition patterns confirms that faithfulness is preserved in the majority of cases (Figure~\ref{fig:faithfulness_7b_bib}, Appendix~\ref{app:faithfulness}; Table~\ref{tab:biasing-transitions}). Notably, compared to \lm{Qwen2.5} models, \lm{Llama3} models are more susceptible to quantization-induced degradation of NLE faithfulness, especially at 4-bit precisions, e.g., \lm{Llama3-8B} with \texttt{GPTQ4} on \data{HealthFC} assessed by CC-SHAP. Thus, although the drop is on average modest, in some configurations, the drop could be substantial, which is not easily predictable. As a result, \textit{practitioners must validate the impact on SE faithfulness per use case}. Moreover, NLEs generated by larger models tend to be more faithful, aligned with the finding from \citet{siegel2025verbositytradeoffsimpactscale}, and larger models show greater robustness to quantization in preserving NLE faithfulness. %(Appendix~\ref{app:faithfulness}).
% \footnote{Figure~\ref{fig:spearman_faithfulness} reveals a moderate correlation between faithfulness measured through counterfactual tests and biasing features (Appendix~\ref{app:faithfulness}).}

%A rank-based comparison (Figure~\ref{fig:ranking}) shows \texttt{integer quantization} preserves faithfulness best on average (\texttt{bib4} / \texttt{bib8}), with \texttt{AWQ} close and \texttt{GPTQ-4/8} generally worse. Most transitions from full precision are faithful→faithful; when flips occur, they more often move faithful→unfaithful (Table~\ref{tab:cf-edit-transitions}, \ref{tab:biasing-transitions}). Method-order Spearman rank correlations indicate strong positive agreement for CT across datasets (\(\rho=0.975,\,p=0.0048\)); in contrast, we observe strong negative relations involving CC-SHAP \hl{(more analysis?)} %(e.g., eSNLI~CC-SHAP vs.\ HealthFC~Bias \(\rho=-0.872\); eSNLI~CC-SHAP vs.\ CT on eSNLI \(\rho=-0.816\)). Notably, \lm{Llama3-70B} shows low CC-SHAP under quantization relative to both full precision and \lm{Llama3-8B} (eSNLI: 0.741→0.278–0.628; HealthFC: 0.409→0.313–0.407). We also observe targeted improvements (e.g., CT for \lm{Llama3-8B} with \texttt{gptq4}, +10\,pp).

\subsubsection{Counterfactual Examples} 
\label{subsubsec:res_cfe}
%the quality of counterfactual examples does not uniformly enhance with larger model sizes, a finding consistent with observations shown by \citet{nguyen-etal-2024-llms, bhattacharjee-etal-2024-zero, li-etal-2024-prompting} and

% \paragraph{The quality and faithfulness of counterfactuals degrade due to quantization.}
Tables~\ref{tab:counterfactual_and_explanation_automatic_evaluation} and \ref{tab:automatic_evaluation_additional} show that quantization negatively affects LLMs' ability to generate CFEs, causing \textit{substantial counterfactual quality degradation} in \textit{validity}, \textit{fluency}, and \textit{textual similarity} (\S\ref{subsec:automatic_evaluation}), with fluency being most affected (on average 6.3\%). This degradation is particularly pronounced for smaller LLMs, whereas larger LLMs demonstrate greater robustness. Moreover, as discussed in Section~\ref{subsec:faithful}, the validity of counterfactuals, measured by LFR, simultaneously reflects their faithfulness (Table~\ref{tab:counterfactual_and_explanation_automatic_evaluation}, Figure~\ref{fig:faithfulness_7b_bib}). We observe that counterfactual faithfulness decreases by an average of 1.5\% under quantization and smaller LLMs exhibit more noticeable faithfulness drops. Counterintuitively, we find that full-precision models may underperform quantized models in generating effective counterfactuals, and larger quantized models sometimes generate lower-quality counterfactuals compared to smaller full-precision models. 

\input{table/faithfulness}

\subsection{Human Evaluation}
\label{subsec:user_study_results}

\begin{figure}[t!]
\centering
\resizebox{\columnwidth}{!}{
\begin{minipage}{\columnwidth}
\includegraphics[width=\columnwidth]{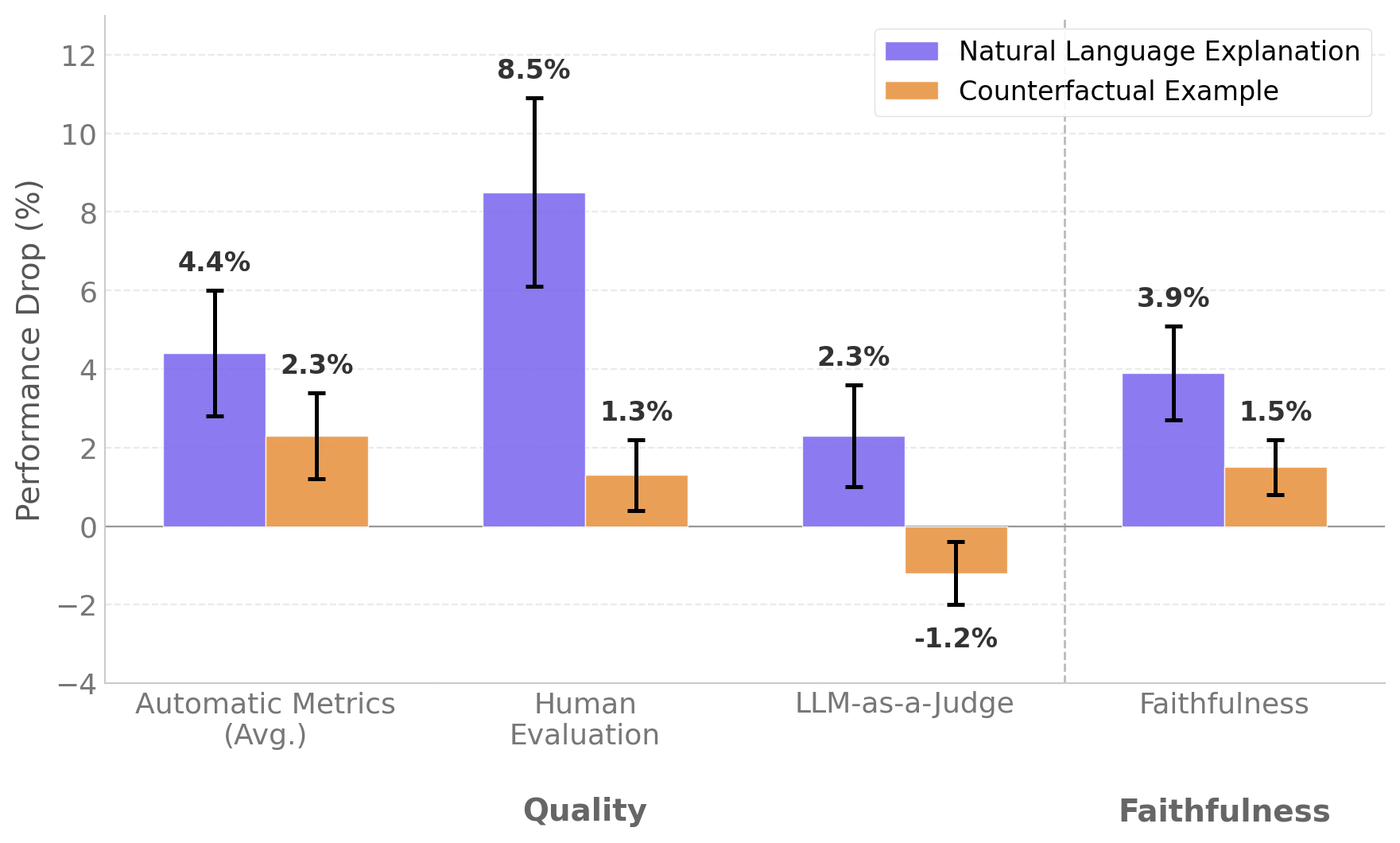}
\end{minipage}
}
\caption{Averaged drop in self-explanation \textit{quality} and \textit{faithfulness} with confidence intervals, as assessed by various automatic evaluation metrics (\S\ref{subsec:automatic_evaluation}), human and LLM-as-a-Judge evaluation dimensions (\S\ref{subsubsec:subjective_ratings}), and faithfulness metrics (\S\ref{subsec:faithful}).} %Automatic metrics \textbf{underestimate} the impact of quantization on NLEs and CFEs compared to the human evaluation, averaged across three models and quantization methods (\S\ref{subsubsec:setup}).}
\label{fig:user_study}
% \vspace{-1em}
\end{figure}

\paragraph{Self-explanations generated by full-precision models are more trustworthy and coherent.} Table~\ref{tab:user_study} presents results from the human evaluation, showing that, overall, NLEs and CFEs generated by full-precision models are perceived as generally more trustworthy and coherent than those generated by quantized models (Figure~\ref{fig:example} and \ref{fig:user_study}). This can be attributed to quantization's effect on confidence calibration: By introducing truncation or rounding, it is harder for the model to capture contextual semantics due to distribution shifts \cite{proskurina-etal-2024-quantization}. As a result, the coherence of generated text is impaired \cite{resendiz2025llmbasedaffectivetextgeneration} and the likelihood of hallucinations increases \cite{li-etal-2024-dawn}, ultimately diminishing annotators' trust in the self-explanations. %\hl{to be checked} Moreover, as shown in Table~\ref{tab:user_study}, \texttt{AWQ} is less effective than \texttt{GPTQ} which further supports the trends observed in the automatic evaluation (\S\ref{subsec:results_explainability}). NLEs and CFEs produced by \texttt{GPTQ4} are occasionally preferred by human annotators over those from \texttt{GPTQ8}, which aligns with findings from the automatic evaluation. 

% ------------------------------------------
% ------------------------------------------

\paragraph{LLM-as-a-Judge evaluation fails to fully capture the impact of quantization.} 

%Figure~\ref{fig:user_study} illustrates that automatic evaluation metrics for CFEs tend to \textit{overestimate} the extent of degradation, while those for NLEs tend to \textit{underestimate} it when compared to user study results. This highlights that automatic metrics may fall short in fully capturing the impact of quantization, particularly for NLEs, where a larger gap exists between automatic evaluation and human evaluation. This discrepancy occurs because most NLE evaluation metrics assess similarity to human-annotated NLEs rather than evaluating fine-grained explanatory attributes (\S\ref{subsubsec:subjective_ratings}). 

To facilitate comparative analysis, an LLM-as-a-Judge (LaaJ) evaluation is conducted on the identical subset of data examples selected for human assessment of trustworthiness and coherence (\S\ref{subsubsec:setup}). Setup details and full correlation results are in Appendix~\ref{app:laaj}. We find a striking discrepancy:  while judge models generally agree with each other (Figure~\ref{fig:laaj_correlation_nle} and \ref{fig:laaj_correlation_cfe}, especially on coherence, with $\kappa$ overall above 0.69), their agreement with human raters is consistently weak or even occasionally negative and is not statistically significant (all $p$-values > 0.05; Figure~\ref{fig:correlation_user_nle} and \ref{fig:correlation_user_cfe}). This misalignment is particularly pronounced for CFEs, where judges systematically diverge from human judgments far more than for NLEs. These findings indicate that, in our setting, LaaJ cannot substitute for human evaluation of quantization effects. One plausible explanation is that the SE quality differences induced by quantization in our setting are subtle (Figure~\ref{fig:user_study}; particularly for CFEs where the full-precision vs. quantized gap is only 1.3\%), and within this narrow margin judges may tend to converge on surface features (e.g., fluency, length) that diverge from the subtle quality signals humans pick up on. Therefore, human evaluation should preferably be considered when subjective judgment is involved in assessing the effect of quantization of SE quality. %Moreover, compared to LaaJ, automatic evaluation is generally more reliable, as these are objective metrics. %this may be attributable to the subtle quality degradation induced by quantization (Figure~\ref{fig:example}). %Judge models cannot easily distinguish the minor differences between SEs generated by full-precision and quantized LLMs (Figure~\ref{fig:user_study}), particularly for CFEs, where the human-evaluated drop is only 1.3\%.

%This difference can be attributed to the plausibility and easier understandability of NLEs \cite{agarwal2024faithfulnessvsplausibilityunreliability} compared to CFEs, which aim to explain model predictions instead of ground-truth labels, thereby increasing the complexity level and making the perception of quality variation due to quantization more difficult.

%Notably, we observe that annotator judgments are sometimes highly polarized - explanations are frequently rated as either ``strongly agree'' or ``strongly disagree'', and in several instances, annotators expressed completely opposing preferences for the same explanation.

\subsection{Discussion} 
\label{subsub:discussion}
\paragraph{Various methods for self-explanations have different sensitivities to quantization.} 
Figure~\ref{fig:user_study} illustrates that quantization impacts self-explanations differently, though it moderately degrades self-explanation quality by up to 8.5\%. NLEs exhibit greater sensitivity, while CFEs demonstrate relative robustness to quantization, with NLE quality degradation being substantially more pronounced (Table~\ref{tab:counterfactual_and_explanation_automatic_evaluation}). Thus, for applications where CFEs are suitable, quantization presents lower risk to explanation quality and faithfulness.

\paragraph{Quantization generally leads to declines in self-explanation quality.} %We observe that explanation quality degradation varies based on model size and the quantization technique applied (Table~\ref{tab:additional_NLE_experiment_llama} to \ref{tab:additional_CFE_experiment_qwen}). 
Table~\ref{tab:counterfactual_and_explanation_automatic_evaluation} displays that no single quantization method invariably outperforms others across all experimental configurations, making it challenging to predict accurately the quantitative impact of quantization on self-explanation quality, since the magnitude of quality degradation depends on the specific \textit{quantization techniques} and \textit{deployed models}. Nevertheless, quantization generally leads to self-explanation quality degradation (Figure~\ref{fig:user_study}; most $p<0.05$), though surprisingly, it can sometimes even improve self-explanation quality. The increase in SE quality from quantization may arise from the reduced entropy of the output distribution and from more consistent, simple language use, as quantization narrows the diversity of LLM outputs \cite{guo-etal-2025-diversity}. Additionally, we observe that \textit{larger quantized models do not consistently generate higher-quality self-explanations than smaller full-precision models}, contradicting the finding of \citet{badshah2024quantifyingcapabilitiesllmsscale}. Furthermore, LLMs with lower-bit precision do not invariably perform worse than those with higher bit precision. These findings may stem from regularization effects \cite{park-etal-2022-robust} or noise \cite{li2024investigating} introduced by quantization, which limits weight precision and may inadvertently enhance self-explanation quality. \looseness=-1

%  and \citet{liu-etal-2024-emergent}

%(Table~\ref{tab:additional_NLE_experiment_llama} to \ref{tab:additional_CFE_experiment_qwen}). , thereby contradicting the findings of \citet{badshah2024quantifyingcapabilitiesllmsscale, liu-etal-2024-emergent}. 

\begin{figure}[t!]
    \centering
    \begin{minipage}{\columnwidth}
        \begin{minipage}[t]{0.5\textwidth}
            \centering
            \includegraphics[width=\textwidth]{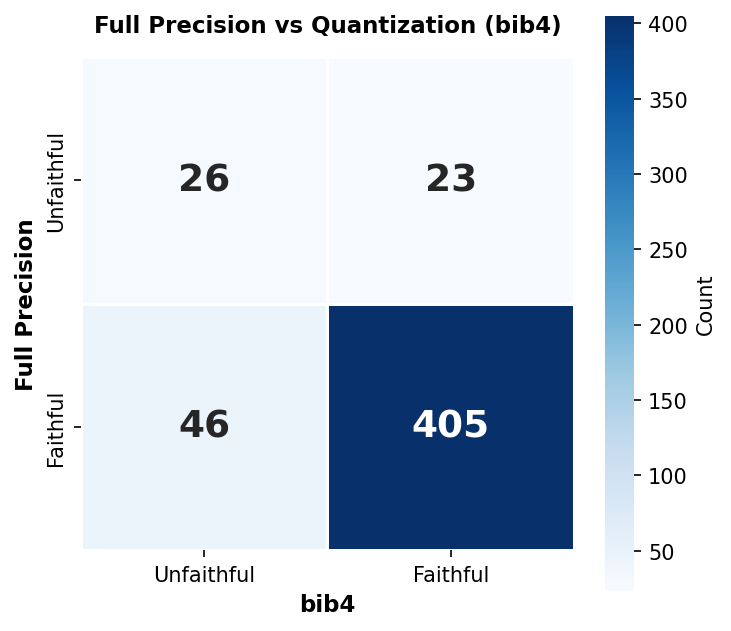}
       \subcaption{Natural language expl.}
        \end{minipage}%
        \hfill%
        \begin{minipage}[t]{0.5\textwidth}
            \centering
            \includegraphics[width=\textwidth]{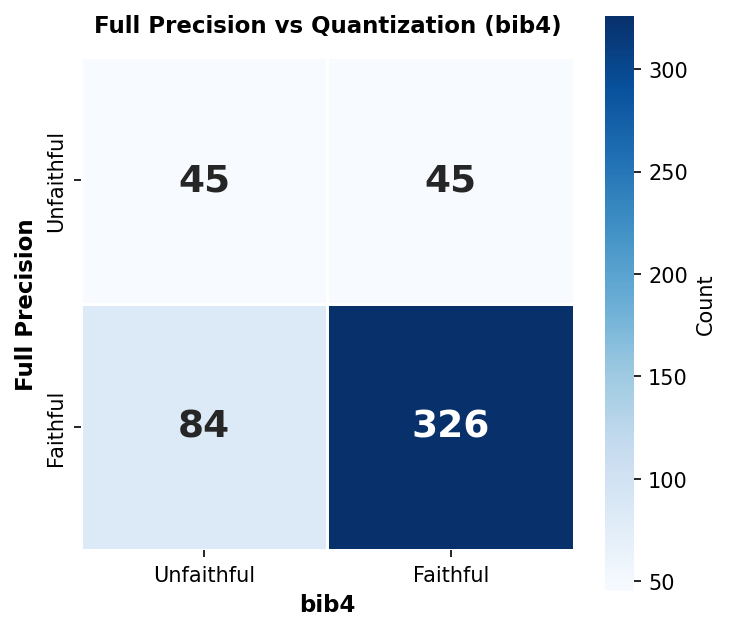}
            \subcaption{Counterfactual example}
        \end{minipage}
    \end{minipage}
    
\caption{Self-explanation faithfulness variation due to quantization for \lm{Qwen2.5-7B} with \texttt{bib4} on \data{eSNLI} measured by biasing feature and self-consistency.}
  \label{fig:faithfulness_7b_bib}
  % \vspace{-1em}
\end{figure}

\paragraph{Self-explanation faithfulness is adversely impacted by quantization.} Figure~\ref{fig:user_study} reveals that, overall, quantization does not notably affect the self-explanation faithfulness, with average degradation of only 1.5\% for CFEs and 3.9\% for NLEs. This minimal impact is evidenced by the fact that the faithfulness of SEs generated by the full-precision and quantized models remains largely unchanged. %This minimal impact is evidenced by the high frequency of SE faithfulness transitions when comparing SEs generated by the full-precision and quantized models: (\textit{faithful} $\rightarrow$ \textit{faithful}) and (\textit{unfaithful} $\rightarrow$ \textit{unfaithful}). 
However, there are more cases with full-precision explanations remaining faithful while quantized versions become unfaithful  (Figure~\ref{fig:faithfulness_7b_bib}), although faithfulness can occasionally be surprisingly enhanced through quantization. A possible assumption is that the model retains core reasoning pathways while discards spurious correlations that lead to unfaithful or lower-quality explanations \cite{mulchandani2025severing}. Moreover, we observe that \textit{self-explanations from larger quantized models are more frequently faithful than those from smaller full-precision models}, as smaller models experience more pronounced faithfulness degradation from quantization (Appendix~\ref{app:faithfulness}).\footnote{This highlights that quantization affects SE quality differently from faithfulness (App.~\ref{app:why}). We hypothesize that under quantization, weights may be more sensitive compared to the decision boundary due to overparametrization, indicating that faithfulness can be better preserved compared to SE quality.} Consequently, \textit{when SE faithfulness is critical, practitioners should consider employing larger quantized models rather than smaller full-precision models}.

%This occurs because smaller models experience more pronounced faithfulness degradation from quantization.
% (on average 9.65\% for NLE)  (on average 7.57\% for NLE)

%However, among the cases where faithfulness does change, quantization more frequently causes degradation, with full-precision explanations remaining faithful while quantized versions become unfaithful, rather than the reverse. Consequently, quantization generally leads to degradation in self-explanation faithfulness, although faithfulness can be sometimes surprisingly enhanced through quantization. 

\paragraph{Ranking of quantization methods based on the extent of degradation.} We assign rankings to quantization methods for each model based on quality changes compared to full-precision models. Subsequently, we calculate the mean ranking across all experimental configurations. We find that no single quantization method consistently outperforms others across \textit{task performance}, self-explanation \textit{quality}, and \textit{faithfulness}. Figure~\ref{fig:ranking} shows that \texttt{GPTQ8}, \texttt{AWQ}, and \texttt{bib8} excel at preserving these metrics, respectively. Moreover, we observe a trade-off among quantization methods between self-explanation (both quality and faithfulness) and task performance preservation. Notably, lower-bit methods can generate explanations with comparable quality to their higher-bit counterparts. %\looseness=-1

\paragraph{Summary.} Quantization leads to degradation in self-explanation quality and faithfulness, with this trend becoming \textbf{more pronounced in smaller models} (Table~\ref{tab:counterfactual_and_explanation_automatic_evaluation} and \ref{tab:faithfulness}).\footnote{To demonstrate the generalizability of our findings to non-free-text SEs, we conduct an additional evaluation of feature attribution \cite{madsen-etal-2024-self} in Appendix~\ref{app:additional_experiments}.} Surprisingly, quantized models occasionally generate self-explanations of higher quality than their full-precision counterparts. Moreover, lower-bit quantization does not necessarily produce inferior self-explanations compared to higher-bit quantization. Although \textbf{quantization effects exhibit variability across models and techniques}, our findings indicate only modest degradation in self-explanation quality and faithfulness (Figure~\ref{fig:user_study} and \ref{fig:faithfulness_7b_bib}), \textbf{rendering quantization a viable compression strategy}. Nevertheless, practitioners should proceed cautiously when deploying quantized LLMs in transparency-critical applications.

\section{Conclusion}
In this work, we examine the impact of quantization on two free-text self-explanation types concerning explanation quality and faithfulness, employing three quantization techniques across six LLMs of varying sizes. Quantization generally causes degradation in both self-explanation quality and faithfulness. While larger models demonstrate limited robustness to quantization regarding explanation quality, they are more robust in preserving faithfulness. Across our experiments, the impact of quantization on self-explanations is highly context-dependent, and no single quantization method consistently outperforms others across \textit{task performance}, explanation \textit{quality}, and \textit{faithfulness}. Our user study further reveals that quantization reduces the coherence and trustworthiness of self-explanations. This heterogeneity suggests practitioners should empirically test multiple quantization strategies for their specific use case rather than assuming a one-size-fits-all solution. Nevertheless, the modest explanation quality and faithfulness degradation indicates that quantized models retain their competence for self-explanation and does not undermine quantization's viability as a model compression strategy.

\section*{Limitations}
% Our experimental work is confined to English-language datasets. Consequently, the effectiveness in other languages may not be comparable. Extending experiments to the multilingual setting is considered as future work.

In our experiments, we extensively compare full-precision models with different quantized versions in 4-bit and 8-bit formats. Lower-bit quantization, such as 1-bit or 2-bit, is not included in our study. Moreover, following \citet{singh2025interpretingeffectsquantizationllms}, the scope of our experiments is limited to post-training quantization (PTQ) techniques. The rationale for focusing on PTQ is twofold: PTQ facilitates an ad-hoc quantization during inference and it offers computational efficiency without necessitating modifications to the model architecture or training process. %Investigating the impact of weight-activation quantization, KV cache compression, or quantization-aware training techniques on self-explanations is counted as future work.

Although quantization can simultaneously affect other model capabilities, we argue that disentangling the impact of quantization from other confounding factors is infeasible, due to the black-box nature of LLMs. Consistent with prior work across multiple domains, e.g., model calibration \cite{singh2025interpretingeffectsquantizationllms}, multilinguality \cite{marchisio-etal-2024-quantization}, and alignment \cite{jin-etal-2024-comprehensive}, which similarly does not disentangle confounding factors, we adopt established experimental protocols while focusing on patterns that demonstrate notable divergence from full-precision models.

\section*{Ethics Statement}
The conducted user study was ethically approved by the Ethics Committee of Faculty IV of Technische Universit\"at Berlin. The participants in our user studies were compensated at or above the minimum wage in accordance with the standards of our host institutions’ regions. The annotation took each annotator 45 minutes on average.

\section*{Author Contributions}
Author contributions are listed according to the CRediT taxonomy as follows:
\begin{itemize}[noitemsep,topsep=0pt,leftmargin=*]
    \item QW: Writing, idea conceptualization, experiments and evaluations, formal analysis, visualization.
    \item NF: Writing – review \& editing, supervision, idea conceptualization.
    \item PA: Writing – review \& editing and idea conceptualization of the comparison between LLM-as-a-Judge evaluation and human evaluation.
    \item FS: NLE faithfulness evaluation.
    \item SO: Writing – review \& editing.
    \item SM: Supervision, review \& editing, and funding acquisition.
    \item VS: Funding acquisition.
\end{itemize}

\section*{Acknowledgment}
We sincerely thank \textbf{Martin Tutek} and CopeNLU members, including \textbf{Jingyi Sun}, \textbf{Nadav Borenstein}, and \textbf{Sekh Mainul Islam}, for their thorough review of an early draft and the anonymous reviewers of EMNLP 2025 and ACL 2026 for their helpful and rigorous feedback.
This work has been supported by the Federal Ministry of Research, Technology and Space (BMFTR) as part of the projects BIFOLD 24B and VERANDA (16KIS2047).

\bibliography{custom}
% \bibliography{anthology,custom}

% \clearpage

\appendix
\section{Dataset Information}
\label{app:dataset}
\subsection{Dataset Examples}
\begin{figure*}[t!]
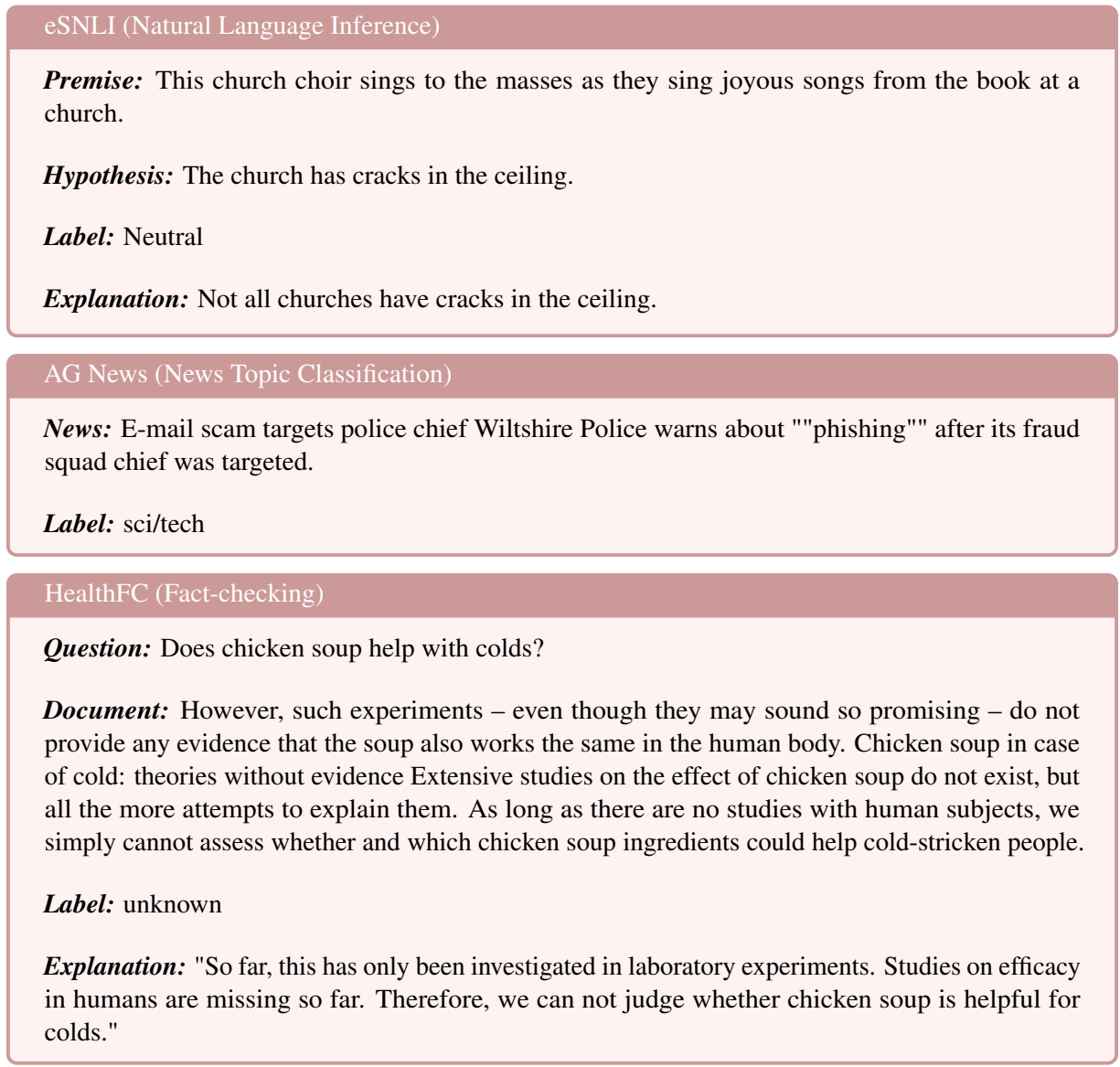

    
    \centering

    \begin{tcolorbox}[colback=pink!20!white, colframe=pink!80!black, title=eSNLI (Natural Language Inference)]
    \textbf{\textit{Premise:}} This church choir sings to the masses as they sing joyous songs from the book at a church.

    \bigskip

    \textbf{\textit{Hypothesis:}} The church has cracks in the ceiling.

    \bigskip

    \textbf{\textit{Label:}} Neutral

    \bigskip

    \textbf{\textit{Explanation:}} Not all churches have cracks in the ceiling.
    \end{tcolorbox}

    \begin{tcolorbox}[colback=pink!20!white, colframe=pink!80!black, title=AG News (News Topic Classification)]
    \textbf{\textit{News:}} E-mail scam targets police chief Wiltshire Police warns about ``phishing'' after its fraud squad chief was targeted.

    \bigskip

    \textbf{\textit{Label:}} sci/tech
    \end{tcolorbox}

    \begin{tcolorbox}[colback=pink!20!white, colframe=pink!80!black, title=HealthFC (Fact-checking)]
    \textbf{\textit{Question:}} Does chicken soup help with colds?

    \bigskip
    \textbf{\textit{Document:}} However, such experiments – even though they may sound so promising – do not provide any evidence that the soup also works the same in the human body. Chicken soup in case of cold: theories without evidence Extensive studies on the effect of chicken soup do not exist, but all the more attempts to explain them. As long as there are no studies with human subjects, we simply cannot assess whether and which chicken soup ingredients could help cold-stricken people.

    \bigskip

    \textbf{\textit{Label:}} unknown

    \bigskip

    \textit{\textbf{Explanation:}} "So far, this has only been investigated in laboratory experiments. Studies on efficacy in humans are missing so far. Therefore, we can not judge whether chicken soup is helpful for colds."
    \end{tcolorbox}
    
    \caption{Examples from \data{eSNLI}, \data{AG News} and \data{HealthFC}.}
    \label{fig:dataset_example}
\end{figure*}

Figure~\ref{fig:dataset_example} presents examples from the \data{eSNLI}, \data{AG News}, and \data{HealthFC} datasets.

\subsection{Label Distribution}
\begin{figure}[!htbp]
  \centering
  \begin{subfigure}{0.85\columnwidth}
    \centering
    \centering
\includegraphics[width=\linewidth]{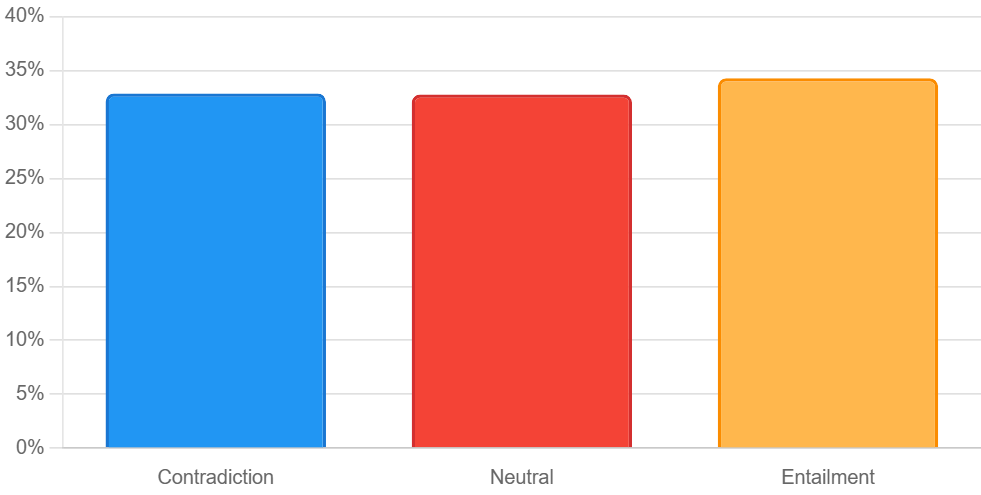}
\caption{\data{eSNLI}}
  \end{subfigure}

  \hfill
  \begin{subfigure}{0.85\columnwidth}
    \centering
    \centering
\includegraphics[width=\linewidth]{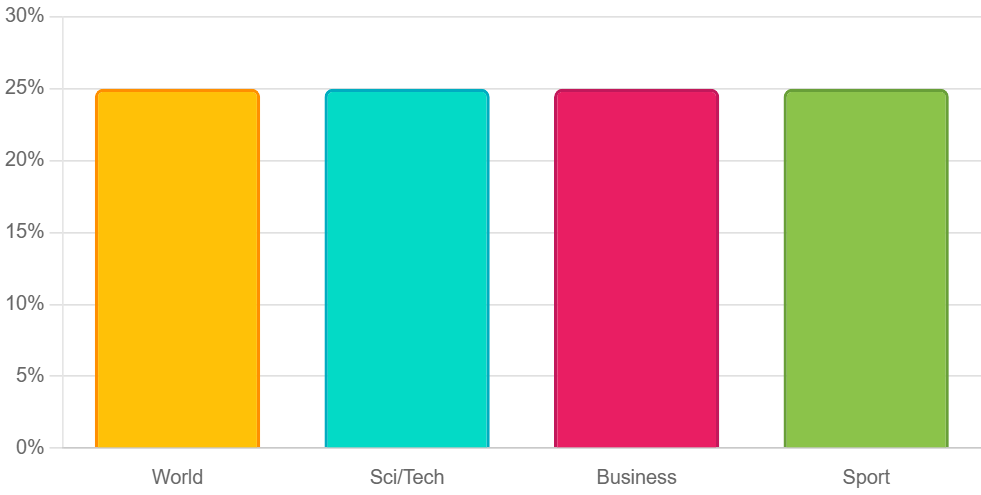}
\caption{\data{AG News}}
\label{fig:ag_news_distribution}
  \end{subfigure}
  \hfill
  \begin{subfigure}{0.85\columnwidth}
\centering
\includegraphics[width=\linewidth]{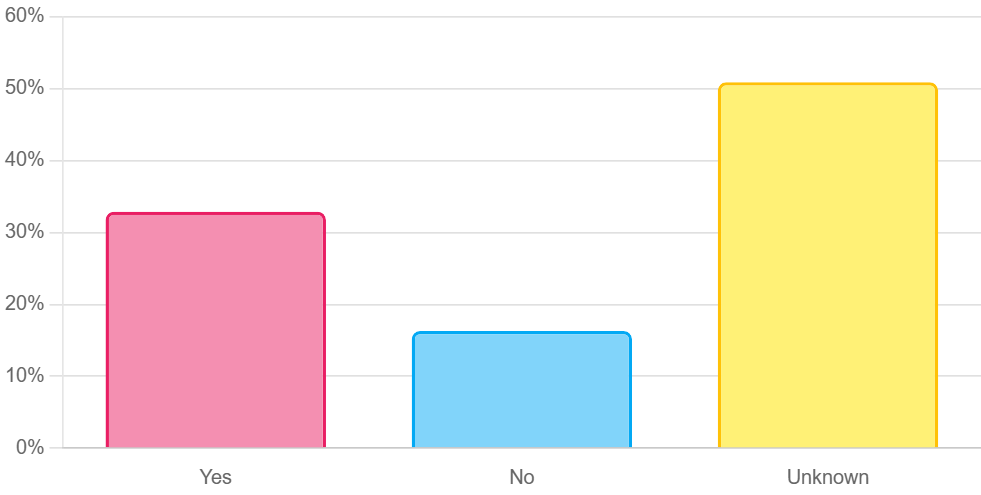}
\caption{\data{HealthFC}}
\label{fig:healthfc_label_distribution}

  \end{subfigure}

  \caption{Label distributions of \data{eSNLI}, \data{AG News} and \data{HealthFC}.}
  \label{fig:app_label_distribution}
\end{figure}

Label distributions of \data{eSNLI}, \data{AG News}, and \data{HealthFC} are shown in Figure~\ref{fig:app_label_distribution}.

% --------------------------------------------------

% --------------------------------------------------

\section{Quantization Method}
\label{app:quantization}
We further provide a detailed overview of three selected quantization methods employed in our experiments (\S\ref{subsec:quantization}).

\paragraph{GPTQ.} \texttt{GPTQ} is a post-training quantization technique that compresses a large language model by reducing its weights to a low precision (typically 4-bit) without needing to retrain the model \cite{frantar-etal-2023-optq}. It works layer-by-layer and group-by-group, solving a local least-squares optimization problem for each set of weights. Crucially, it uses second-order information (Hessian estimates) to intelligently decide which weights can be approximated (quantized) with the least impact on the model's overall output accuracy.

\paragraph{AWQ.} \texttt{AWQ} is an Activation-aware Weight Quantization technique that compresses Large Language Models to low precision by prioritizing accuracy \cite{lin-etal-2024-awq}. It uses a calibration dataset to find salient channels (groups of weights) that are highly sensitive to the model's activations and scales up these critical weights before quantization to protect them from accuracy loss when their precision is reduced.

\paragraph{BitSandBytes} \sys{BitSandBytes} \cite{dettmers-etal-2022-8bits} identifies and isolates outliers, which are model weights or data points with values significantly deviating from the norm. To maintain high precision, these outliers are preserved in 16-bit floating-point format. The remaining non-outliers (standard-range values) are efficiently quantized to 4- or 8-bit integers.

\section{Models \& Inference Time}
\label{app:inference}
\input{table/model}
Table~\ref{tab:used_model} presents details of all LLMs used in our experiments (\S\ref{subsec:models}), including model sizes, quantization approaches and corresponding URLs from the Hugging Face Hub. All models were directly obtained from the Hugging Face repository. All experiments were conducted using A100 or H100 GPUs. Explanation generation across the entire dataset, including both natural language explanations (NLEs) and counterfactual examples (CFEs), can be completed within 10 hours.

\section{Prompting-based Feature Attribution}
\label{app:additional_experiments}
\paragraph{Motivation.} In this paper, we specifically focus on two representative \textit{free-text} self-explanations: natural language explanations (NLEs) and counterfactual examples (CFEs) (\S\ref{subsec:methods}). The selection of self-explanations is motivated by two considerations: (1) their widespread adoption in the literature \cite{madsen-etal-2024-self, agarwal2024faithfulnessvsplausibilityunreliability, wang-etal-2025-fitcf}; and (2) their free-text nature, which facilitates human comprehension and enables more straightforward quantification of quantization-induced degradation through user studies. Furthermore, NLEs and CFEs necessitate multi-dimensional evaluation, whereas feature attribution and redaction-based explanations generally require assessment only along task performance and faithfulness dimensions rather than text quality dimensions. Additionally, the other explanation methods largely depend on the type of technique instead of the underlying model, indicating that quantization may not have large effects on them.

\input{table/attribution}

\paragraph{Setup for Feature Attribution Evaluation.} In order to further show the generalizability of our findings, we conduct a small-scale experiments on prompt-based feature attribution \cite{madsen-etal-2024-self}, which is a not necessarily faithful approximation of model internal-based feature attribution, on \data{eSNLI} with \lm{Qwen2.5}-\{7,14,32,72B\} and \lm{Llama3}-\{8,70B\} with full-precision, AWQ, GPTQ4/8, and bib4/8, while the degradation of task performance is reported in Table~\ref{tab:task_performance}.

\paragraph{Results.} The findings derived from Table~\ref{tab:attribution} align with those reported in the main body of the paper. Quantization negatively impacts feature attribution faithfulness, as evidenced by the average faithfulness decline across all models (\S\ref{subsubsec:nle}). Notably, no single quantization method outperforms the others or better preserves feature attribution faithfulness (\S\ref{subsub:discussion}). Lastly, consistent with the findings for NLEs and CFEs, feature attributions from smaller models exhibit more pronounced faithfulness degradation under quantization.

\section{Results with Additional Datasets}
\label{app:automatic_evaluation}

\input{table/automatic_evaluation_nle}
Table~\ref{tab:automatic_evaluation_additional} displays the automatic evaluation results of natural language explanations on \data{ECQA} and counterfactual examples on \data{IMDb}.

\section{Annotation}
\label{app:annotation_guideline}

% \subsection{Annotation Guideline}
Figure~\ref{fig:annotation_guideline} displays the annotation guideline that we provide to human annotators. NLEs and CFEs are presented to annotators in the form of questionnaires. We use the Crowdee\footnote{\url{https://www.crowdee.com/}} crowdsourcing platform to recruit annotators, distribute the questionnaires, and store their responses. A total of 48 annotators were recruited, all of whom are native English speakers without requiring specific expertise in explainable AI (XAI). Each annotator was given 15 explanations, along with two evaluation dimensions (\S\ref{subsubsec:subjective_ratings}). Each explanation was evaluated by at least two annotators. 

\input{table/user_study}

Table~\ref{tab:user_study} summarizes the observed self-explanation degradation in terms of trustworthiness and coherence.

% % ------------------------------------------
% \begin{figure}[t!]
% \centering
% \resizebox{\columnwidth}{!}{
% \begin{minipage}{\columnwidth}
% \includegraphics[width=\columnwidth]{figure/human_vs_automatic.png}
% \end{minipage}
% }
% \caption{Self-explanation quality drop as measured by both automatic and human evaluation. We compute the average extent of quality reduction, as assessed by various automatic evaluation metrics (\S\ref{subsubec:nle}, \S\ref{subsubsec:cfe}) and human evaluation dimension (\S\ref{subsubsec:subjective_ratings}).} %Automatic metrics \textbf{underestimate} the impact of quantization on NLEs and CFEs compared to the human evaluation, averaged across three models and quantization methods (\S\ref{subsubsec:setup}).}
% \label{fig:user_study}
% \end{figure}
% % ------------------------------------------

\section{LLM-as-a-Judge Evaluation}
\label{app:laaj}
\input{table/judge}
\subsection{Setup} 
The adoption of LLMs as evaluators for complex tasks, referred to as ``LLM-as-a-Judge'', has gained popularity to perform evaluations by assigning quality scores in accordance with human intuition \cite{zheng-etal-2023-llm, huang-etal-2024-chatgpt}. In addition to automatic and human evaluation, we investigate how well LLMs can quantitatively assess the explanation quality degradation caused by quantization. For this purpose, we select three open-source LLMs of varying sizes that are commonly used in the literature \cite{gu2025surveyllmasajudge}: \lm{DeepSeek-R1} \cite{deepseekai2025deepseekr1incentivizingreasoningcapability}, \lm{Gemma3-27B} \cite{gemmateam2025gemma3technicalreport} and \lm{GPT-OSS-120B} \cite{openai2025gptoss120bgptoss20bmodel}. Judge models assess the trustworthiness and coherence of self-explanations generated by LLMs with varying levels of precision (\S\ref{subsubsec:subjective_ratings}).

% ------------------------------------------
\begin{figure*}[t!]
\centering
\resizebox{\textwidth}{!}{
\begin{minipage}{\columnwidth}
\includegraphics[width=\columnwidth]{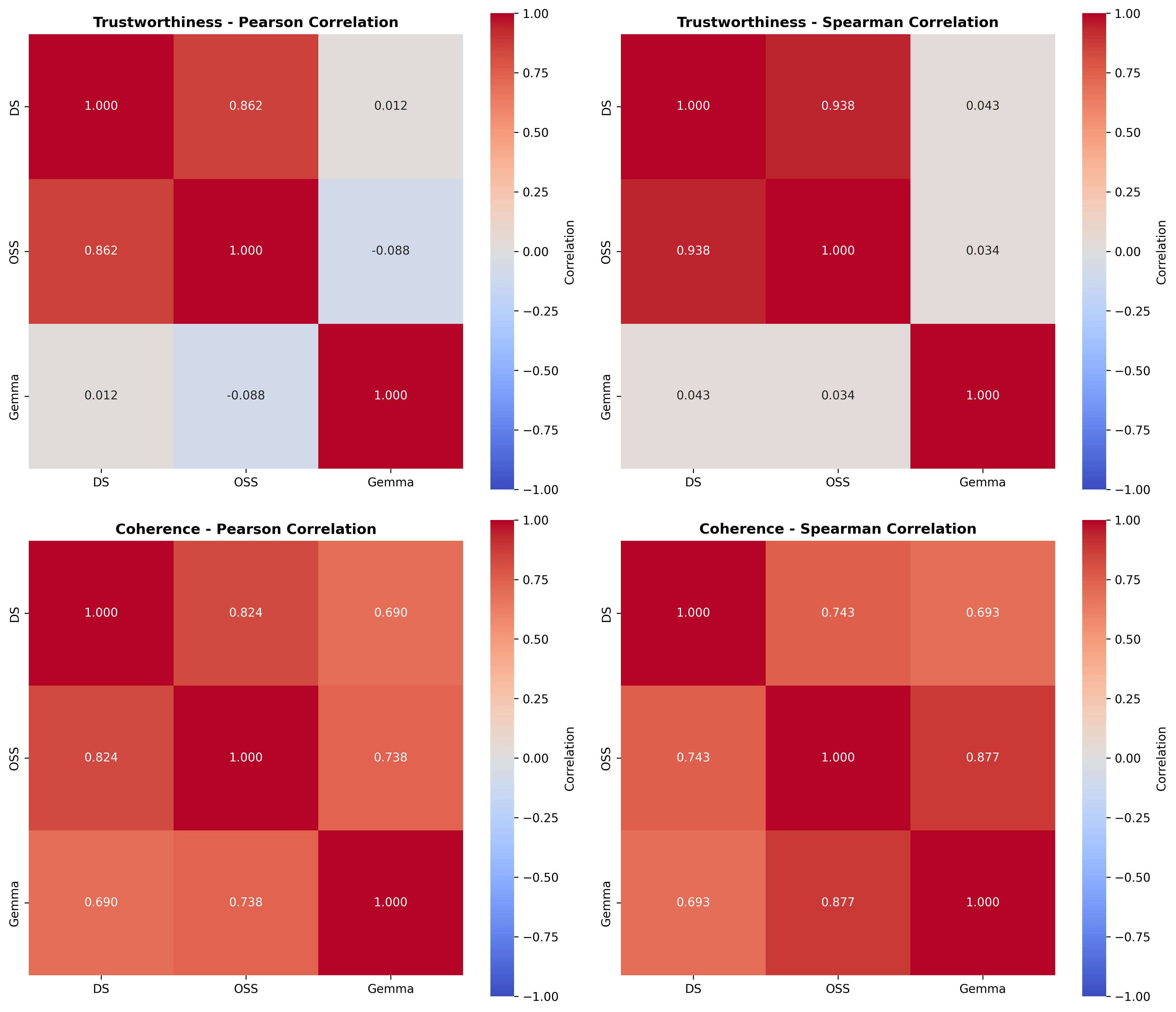}
\end{minipage}
}
\caption{Pearson and Spearman correlation heatmaps for \lm{DeepSeek-R1} (DS), \lm{GPT-OSS-120B} (OSS), and \lm{Gemma3-27B} (Gemma) for \textbf{natural language explanations} evaluated on \textit{trustworthiness} and \textit{coherence}.} 
\label{fig:laaj_correlation_nle}
\end{figure*}
% ------------------------------------------

% ------------------------------------------
\begin{figure*}[t!]
\centering
\resizebox{\textwidth}{!}{
\begin{minipage}{\columnwidth}
\includegraphics[width=\columnwidth]{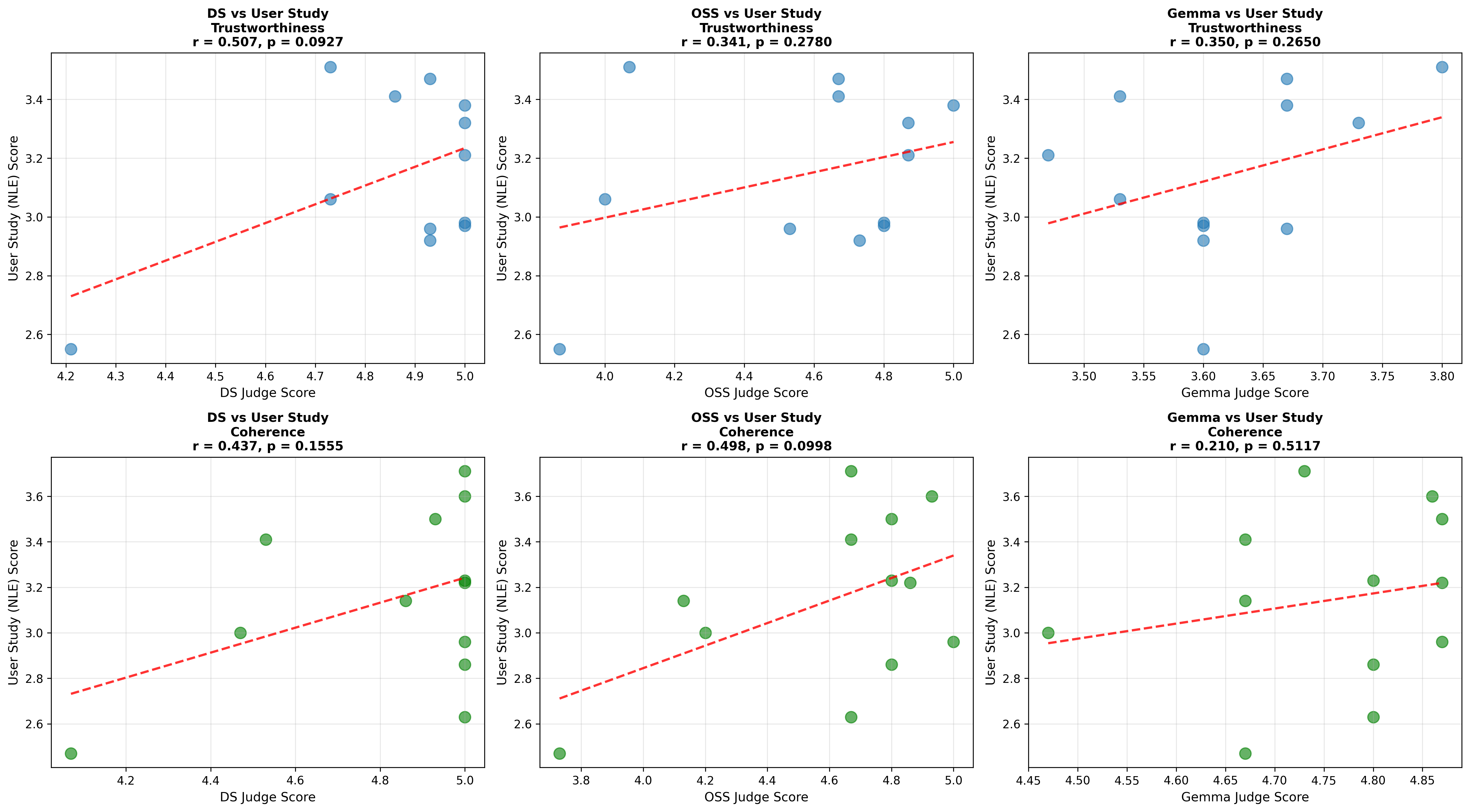}
\end{minipage}
}
\caption{Correlations and significance between judge models (\lm{DeepSeek-R1} (DS), \lm{GPT-OSS-120B} (OSS), and \lm{Gemma3-27B} (Gemma)) and user study for \textbf{natural language explanations} evaluated on \textit{trustworthiness} and \textit{coherence}.} 
\label{fig:correlation_user_nle}
\end{figure*}
% ------------------------------------------

% ------------------------------------------
\begin{figure*}[t!]
\centering
\resizebox{\textwidth}{!}{
\begin{minipage}{\columnwidth}
\includegraphics[width=\columnwidth]{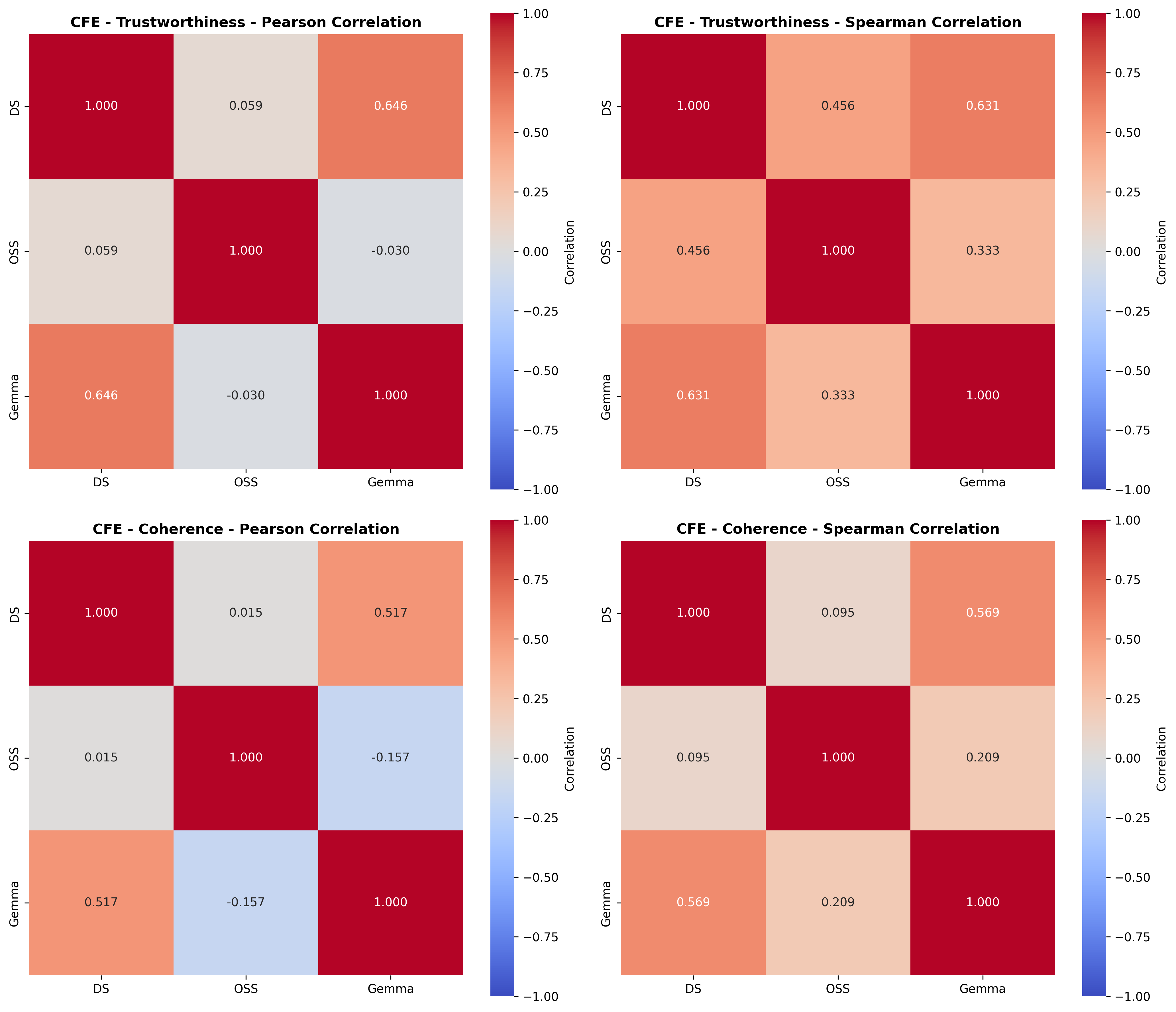}
\end{minipage}
}
\caption{Pearson and Spearman correlation heatmaps for \lm{DeepSeek-R1} (DS), \lm{GPT-OSS-120B} (OSS), and \lm{Gemma3-27B} (Gemma) for \textbf{counterfactual explanations} evaluated on \textit{trustworthiness} and \textit{coherence}.} 
\label{fig:laaj_correlation_cfe}
\end{figure*}
% ------------------------------------------

% ------------------------------------------
\begin{figure*}[t!]
\centering
\resizebox{\textwidth}{!}{
\begin{minipage}{\columnwidth}
\includegraphics[width=\columnwidth]{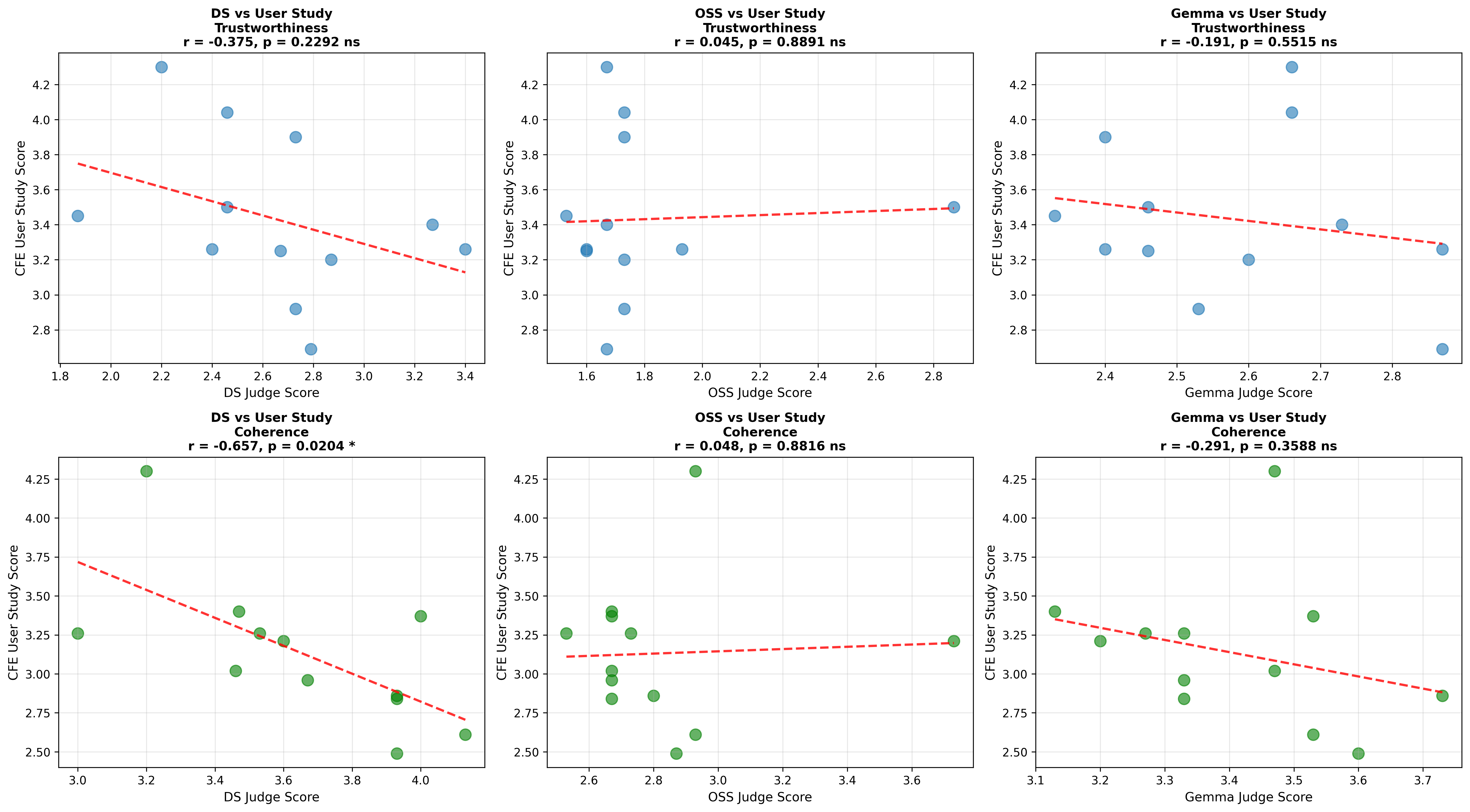}
\end{minipage}
}
\caption{Correlations between judge models (\lm{DeepSeek-R1} (DS), \lm{GPT-OSS-120B} (OSS), and \lm{Gemma3-27B} (Gemma)) and user study for \textbf{counterfactual explanations} evaluated on \textit{trustworthiness} and \textit{coherence}.} 
\label{fig:correlation_user_cfe}
\end{figure*}
% ------------------------------------------

\subsection{Results} 
The LLM-as-a-Judge evaluation results for data examples selected for human evaluation (\S\ref{subsubsec:setup}) are displayed in Table~\ref{tab:judge}.

\subsubsection{Natural Language Explanations} 

\paragraph{Inter Judge Model Agreement.} The within-category correlation analysis reveals notable differences in judge agreement across the two evaluation metrics (Figure~\ref{fig:laaj_correlation_nle}). For \textbf{trustworthiness}, \lm{DeepSeek-R1} and \lm{GPT-OSS-120B} judges demonstrate strong agreement with a Pearson correlation of 0.862 and Spearman correlation of 0.938, both highly significant ($p < 0.001$). However, \lm{Gemma3-27B} shows essentially no correlation with either \lm{DeepSeek-R1} ($r$ = 0.012) or \lm{GPT-OSS-120B} ($r$ = -0.088) when evaluating trustworthiness. The pattern shifts considerably for \textbf{coherence}, where all three judges show moderate to strong agreement with each other. \lm{DeepSeek-R1} and \lm{GPT-OSS-120B} maintain solid agreement ($r$ = 0.824, $\rho$ = 0.743), while \lm{Gemma3-27B} now correlates moderately well with \lm{DeepSeek-R1} ($r$ = 0.690) and strongly with \lm{GPT-OSS-120B} ($\rho$ = 0.877). This indicates that while judges can reach reasonable consensus on what constitutes coherent output, they diverge substantially on trustworthiness assessments, with \lm{Gemma3-27B} being the outlier.

\paragraph{Correlation with the User Study.} Figure~\ref{fig:correlation_user_nle} shows the correlation between judge models and the user study. None of the correlations are statistically significant (all $p$-values > 0.05), indicating weak to moderate alignment between automated judge models and human user study evaluations. The highest correlation is \lm{GPT-OSS-120B} for \textit{Coherence} ($r$ = 0.498, $p$ = 0.100), which approaches but doesn't reach significance. \lm{DeepSeek-R1} shows the highest correlation for \textit{Trustworthiness} ($r$ = 0.507, $p$ = 0.093), also approaching significance. However, the Spearman correlations are even weaker, particularly for trustworthiness.

\subsubsection{Counterfactual Explanations}
\paragraph{Inter Judge Model Agreement.} Figure~\ref{fig:laaj_correlation_cfe} reveals that the CFE judgments show dramatically different correlation patterns compared to the NLE judgments. \lm{DeepSeek-R1} and \lm{GPT-OSS-120B} judges show essentially no correlation with each other for either metric. The only statistically significant correlation is \lm{DeepSeek-R1} vs \lm{Gemma3-27B} for trustworthiness ($r$ = 0.646, $p$ = 0.023). However, \lm{GPT-OSS-120B} shows no meaningful correlation with either \lm{DeepSeek-R1} or \lm{Gemma3-27B} across both metrics. These findings highlight that judge alignment is task-dependent and cannot be assumed to generalize across different explanation types.

\paragraph{Correlation with the User Study.} Figure~\ref{fig:correlation_user_cfe} shows that all three judge models demonstrate weak or negative correlations with human evaluation, raising concerns about using these judge models for evaluating quantization's impact on counterfactual quality without careful calibration.

\begin{figure*}[h!]
    
    \centering

    \begin{tcolorbox}[colback=purple!20!white, colframe=purple!80!white, title=Annotation Guideline for Human Evaluation]

    \textbf{\#\#\# User Study Description}:
    
    Dear participants,
    
    Thanks for attending our user study. This study focuses on evaluating model-generated explanations. We present two types of explanations:
    \begin{itemize}
        \item \textbf{Counterfactual Example (CFE):} A minimally edited version of the input text that results in a change in the model’s prediction. 
        \item \textbf{Natural Language Explanation (NLE)}: A textual justification generated by the model to explain its decision-making process for a given input.
    \end{itemize}
    
    Each explanation should be evaluated along the following two dimensions. Please assign a score to each dimension on a scale from 1 \textbf{(strongly disagree)} to 5 \textbf{(strongly agree)}.
    \begin{itemize}
        \item \textbf{Trustworthiness}: Evaluate whether the provided explanation is trustworthy and can be relied upon by humans;
    
        \item \textbf{Coherence}: Assess whether the provided explanation is sensible, clear, and coherent, and effectively captures the rationale;
    
    \end{itemize}

    \bigskip
    
    \textbf{\#\#\# Dataset Structure:}
    
    \textbf{e-SNLI} (Stanford Natural Language Inference): Each example consists of a premise and a hypothesis. The task is to determine the relationship between the two, categorizing it as either Entailment, Contradiction, or Neutral based on the information in the premise.
    \bigskip

    \textbf{e-SNLI Example}: \{example\}
    \bigskip
    
    \textbf{Explanation}: \{example explanation\}
    \bigskip
    
    \textit{\textbf{Rating:}}
    \bigskip
    
    \textbf{Trustworthiness}: \{score\}
    \bigskip
    
    \textbf{Coherence}: \{score\}
    \bigskip

    \textit{Entailment} means the hypothesis must be true if the premise is true. \textit{Contradiction} means the hypothesis must be false if the premise is true. \textit{Neutral} means the hypothesis might be true, or might not — we can’t tell just from the premise.

    \bigskip

    \textbf{\#\#\# User Study Instruction:}
    
    You will be provided with 15 instances to evaluate. For the \textbf{counterfactual example} evaluation, each instance includes a premise–hypothesis pair. Your task is to evaluate only the quality of the \textbf{premise} according to the two dimensions described above. For the \textbf{natural language explanation} evaluation, each instance also includes a premise–hypothesis pair, along with a \textbf{model-generated justification} in natural language. In this case, your task is to evaluate the provided justification based on the same two dimensions.

    \end{tcolorbox}
    
    \caption{Annotation Guideline.
    }
    \label{fig:annotation_guideline}
\end{figure*}

\section{Task Performance}

\input{table/task_performance}
Table~\ref{tab:task_performance} illustrates the task performance of various quantization methods applied to deployed models on the \data{eSNLI} and \data{HealthFC} datasets.

\input{table/performance-quality}
Table~\ref{tab:quality_performance} displays the Spearman correlation coefficient between task performance and natural language explanation quality, with values averaged across different quantization methods for each individual model. We find that, overall, the correlation between task performance degradation and self-explanation quality degradation is rather weakly positive and occasionally weakly negative. This indicates that task performance degradation contributes to self-explanation quality degradation to some extent.

\section{Quantization Method Ranking}
\begin{figure*}[t!]
  \centering

  \begin{subfigure}{\columnwidth}
    \centering
    \centering
\includegraphics[width=\linewidth]{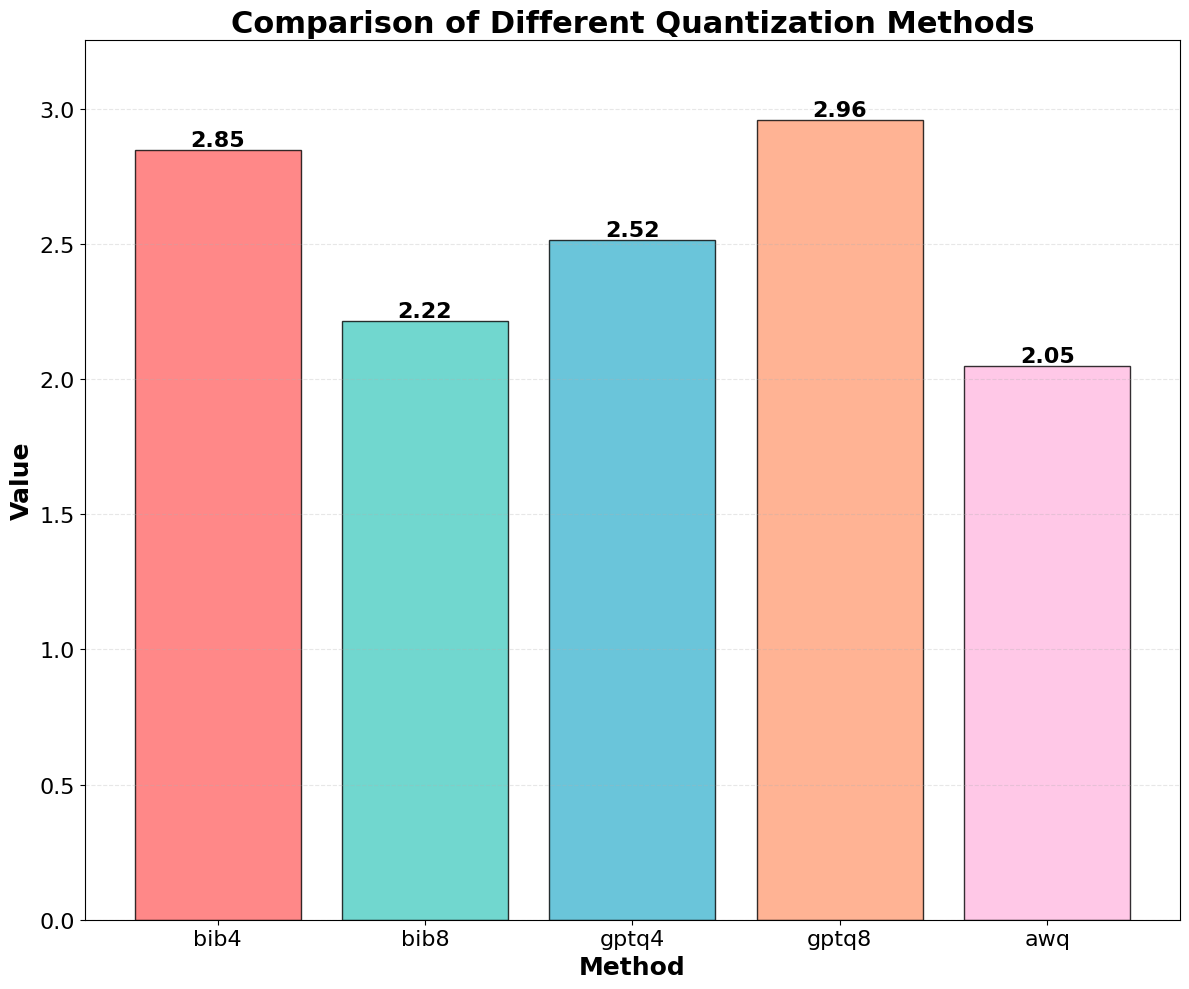}
\caption{Self-explanation quality}
\label{fig:quality_ranking}
  \end{subfigure}
  \hfill
  \begin{subfigure}{\columnwidth}
\centering
\includegraphics[width=\linewidth]{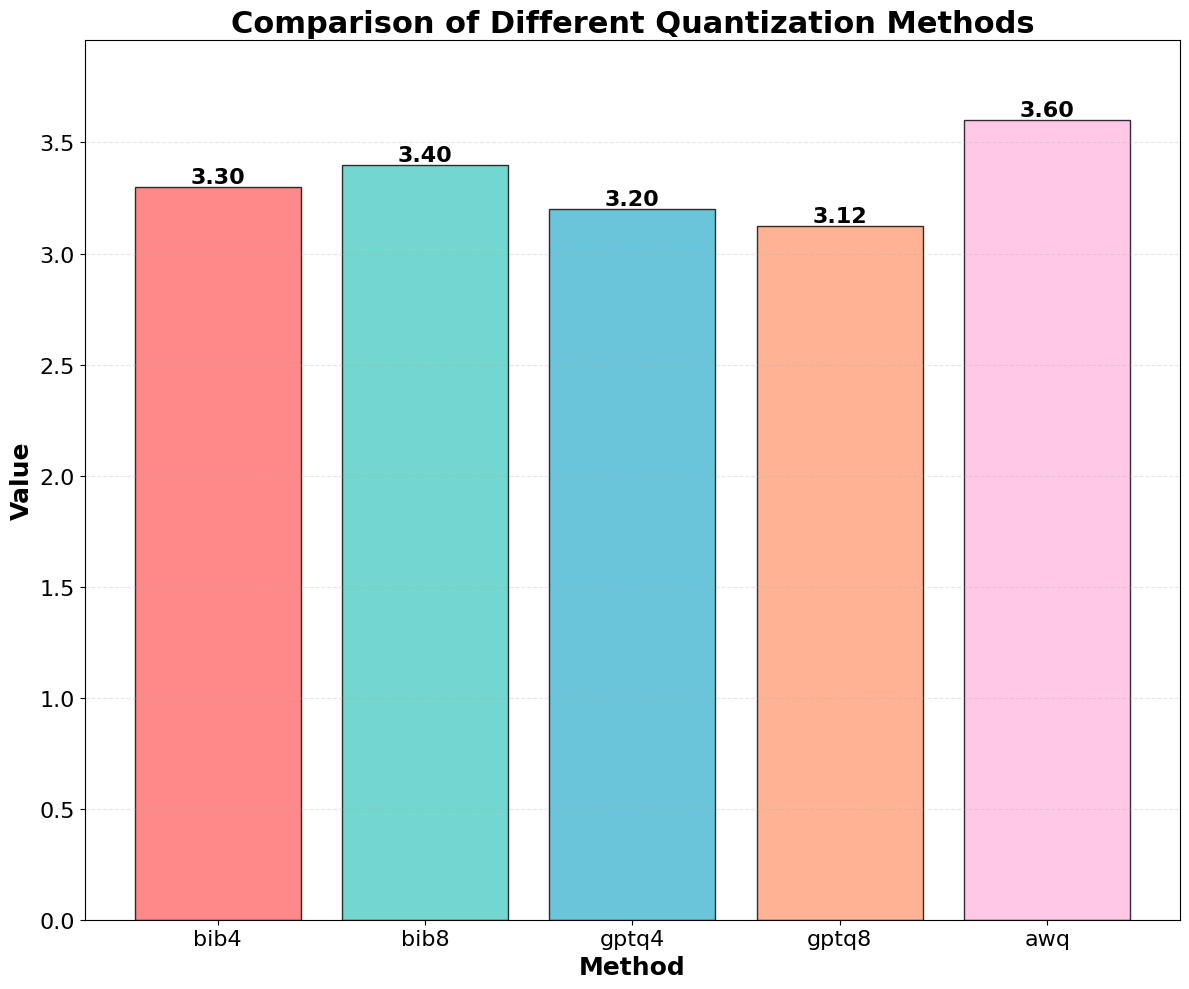}
\caption{Task performance}
\label{fig:performance_ranking}

  \end{subfigure}
  \begin{subfigure}{\columnwidth}
    \centering
    \centering
\includegraphics[width=\linewidth]{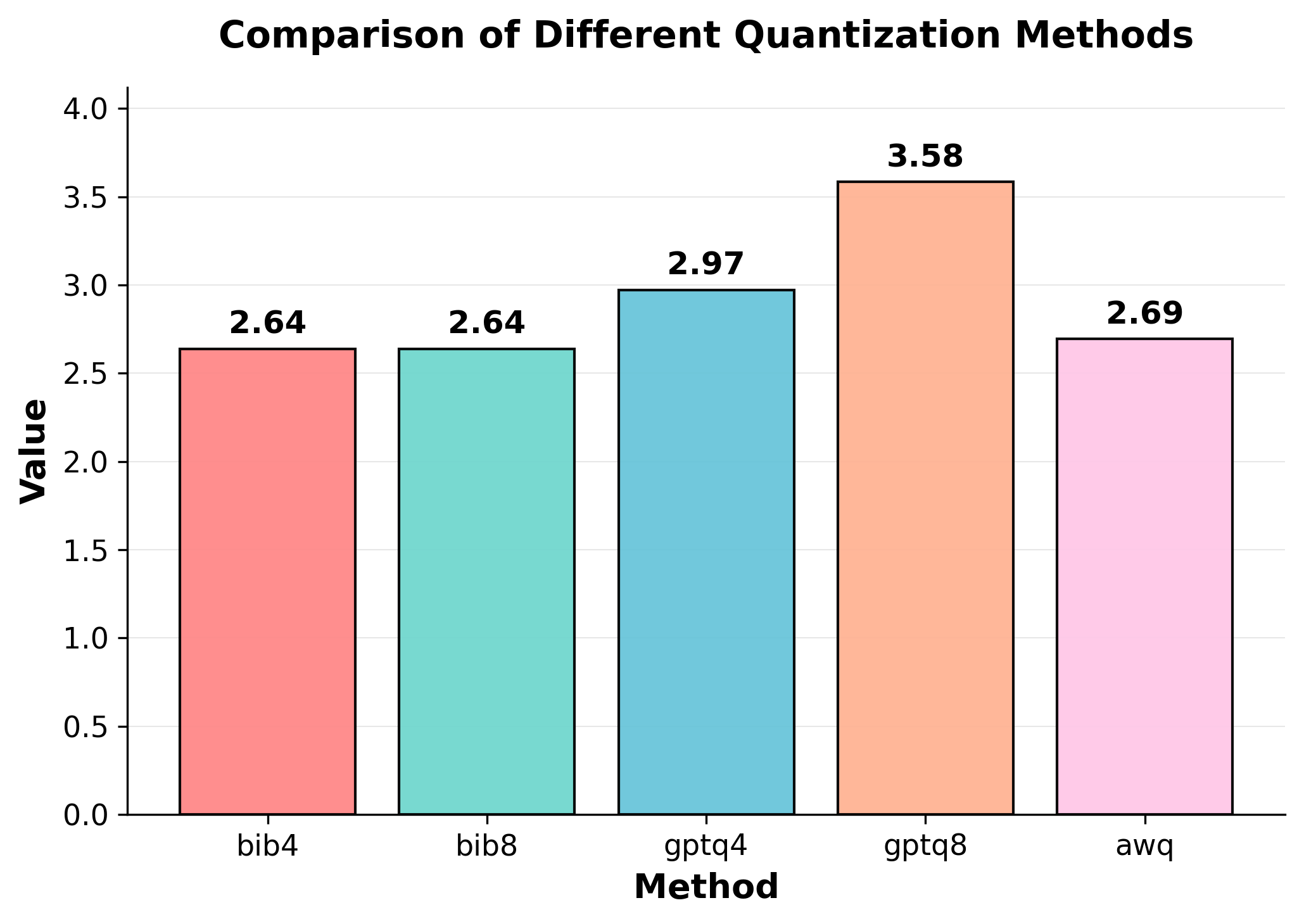}
\caption{Faithfulness}
\label{fig:faithfulness_ranking}
  \end{subfigure}
  \hfill
  \caption{Quantization method ranking.}
  \label{fig:ranking}
\end{figure*}

We assign rankings to quantization methods for each model based on their results (Table~\ref{tab:counterfactual_and_explanation_automatic_evaluation}, Table~\ref{tab:task_performance}) and calculate the mean ranking across all experimental configurations. Figure~\ref{fig:ranking} shows the quantization method ranking based on self-explanation quality and task performance, respectively. Quantization methods demonstrating superior preservation of full-precision LLM capabilities are assigned lower ranking values. We observe that while \texttt{AWQ} is optimal for preserving self-explanation quality and \texttt{GPTQ8} is suboptimal, \texttt{GPTQ8} is optimal for preserving task performance while \texttt{AWQ} is suboptimal. Furthermore, no quantization method can simultaneously be optimal in preserving self-explanation quality and task performance.

% --------------------------------------------------

\section{Faithfulness}
\label{app:faithfulness}

\subsection{Faithfulness Metrics}
We detail the specific faithfulness metrics utilized for evaluating natural language explanations in the subsequent discussion.

% \paragraph{Counterfactual Test.} \citet{atanasova-etal-2023-faithfulness} involve training a model to execute counterfactual interventions by introducing new words into the LLM input. The criterion for assessing explanation unfaithfulness is defined as follows: A change in the LLM's prediction resulting from the intervention, coupled with the absence of the inserted counterfactual terms in the original explanation, constitutes an unfaithful explanation.

\paragraph{Biasing Features.} \citet{turpin2023language} examine the faithfulness of Chain-of-Thought (CoT) explanations that appear before the answer. Their methodology relies on introducing biasing features, such as ``Suggested Answer'' or ``Answer is always A'' in few-shot learning, or stereotype-inducing input edits, to the context. Unfaithfulness is established when the model's answer changes due to the bias, but the explanation does not explicitly state the bias as the reason for the decision (e.g., not generating a phrase like ``Because you suggested A.'').

\paragraph{CC-SHAP.} CC-SHAP assess the self-consistency of LLM explanations \cite{parcalabescu-frank-2024-measuring}. It works by using SHAP values to compare how a model's input contributes to generating the predicted answer versus generating the explanation. The core idea is that a highly consistent explanation should rely on the same important input tokens as the prediction, allowing the method to measure the alignment between the input's importance for the answer and its importance for the explanation, all without needing to edit or perturb the model's input.

\subsection{Results}

\input{table/biasing_transitions}

\begin{figure*}[t!]
  \centering

    \begin{subfigure}{\textwidth}
    \centering
    \centering
\includegraphics[width=\linewidth]{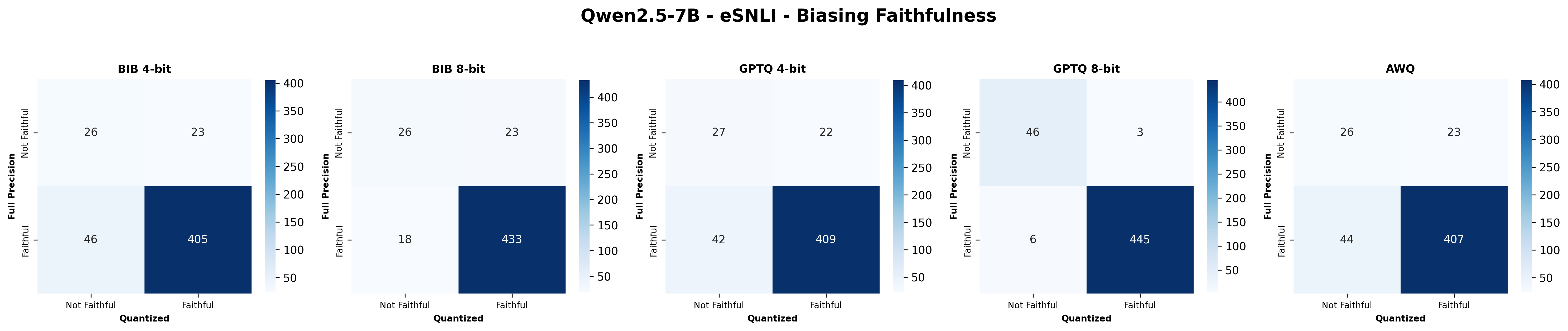}
\caption{\data{eSNLI}}
  \end{subfigure}
  
  \hfill

    \begin{subfigure}{\textwidth}
    \centering
    \centering
\includegraphics[width=\linewidth]{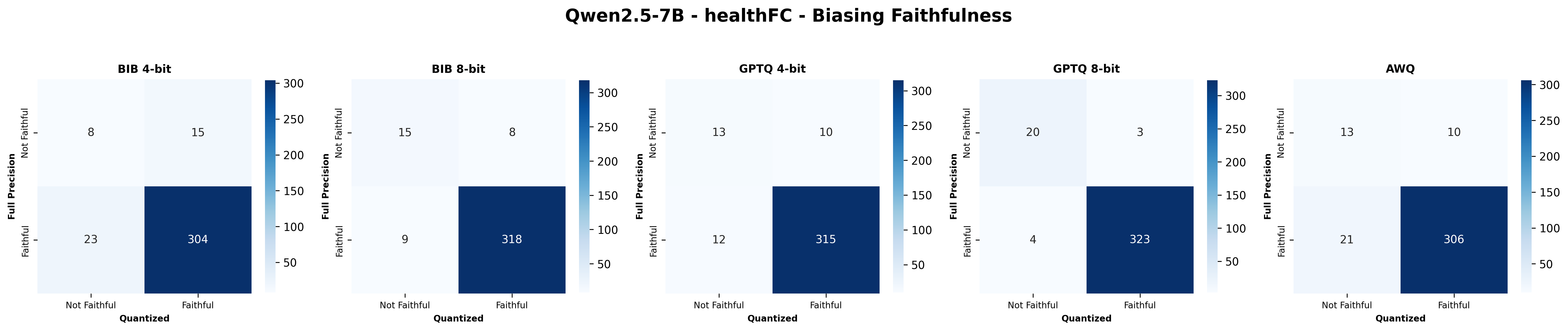}
\caption{\data{HealthFC}}

  \end{subfigure}
    \hfill

  \begin{subfigure}{\textwidth}
    \centering
    \centering
\includegraphics[width=\linewidth]{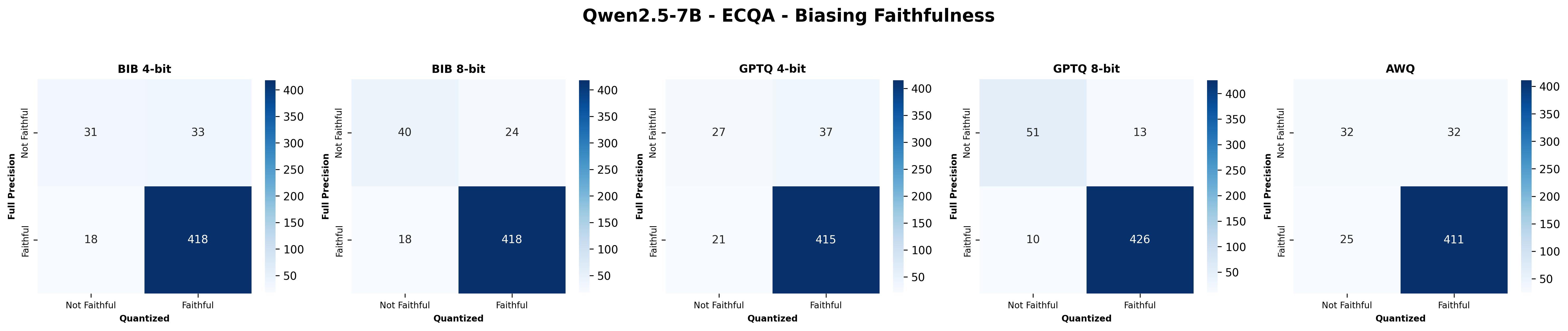}
\caption{\data{ECQA}}

  \end{subfigure}
  
  \hfill

    \caption{Faithfulness Variation \lm{Qwen2.5-7B} (\S\ref{subsub:discussion})}
  \label{fig:faith_7b}
\end{figure*}

\begin{figure*}[t!]
  \centering

    \begin{subfigure}{\textwidth}
    \centering
    \centering
\includegraphics[width=\linewidth]{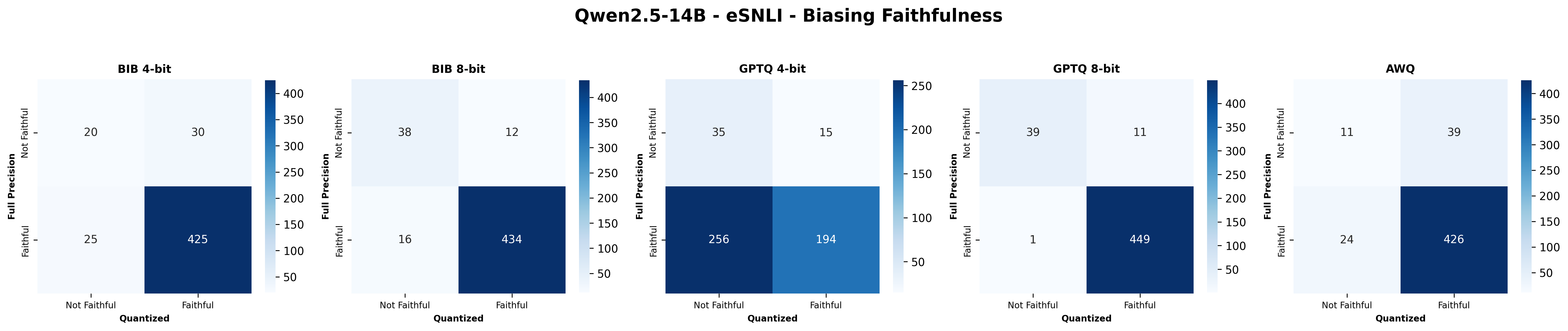}
\caption{\data{eSNLI}}
  \end{subfigure}
  
  \hfill

    \begin{subfigure}{\textwidth}
    \centering
    \centering
\includegraphics[width=\linewidth]{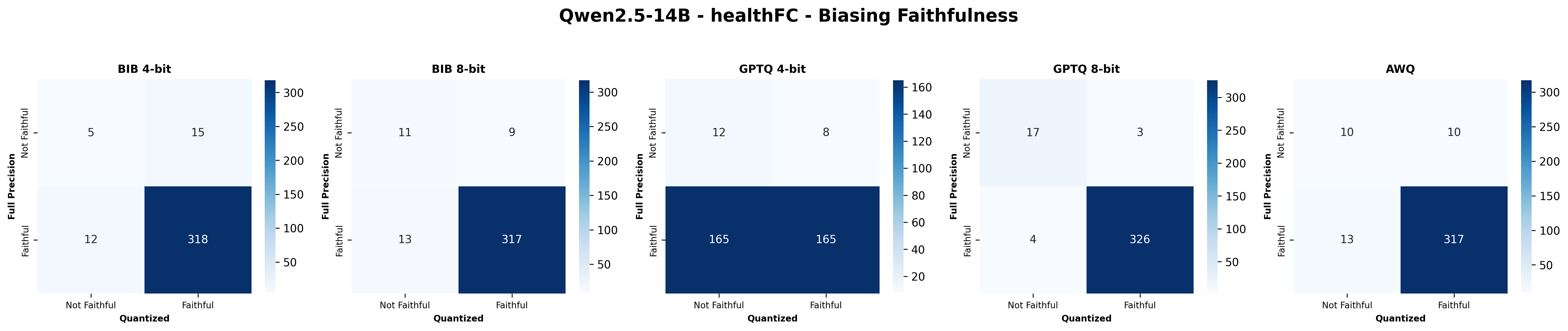}
\caption{\data{HealthFC}}

  \end{subfigure}
  
  \hfill

  \begin{subfigure}{\textwidth}
    \centering
    \centering
\includegraphics[width=\linewidth]{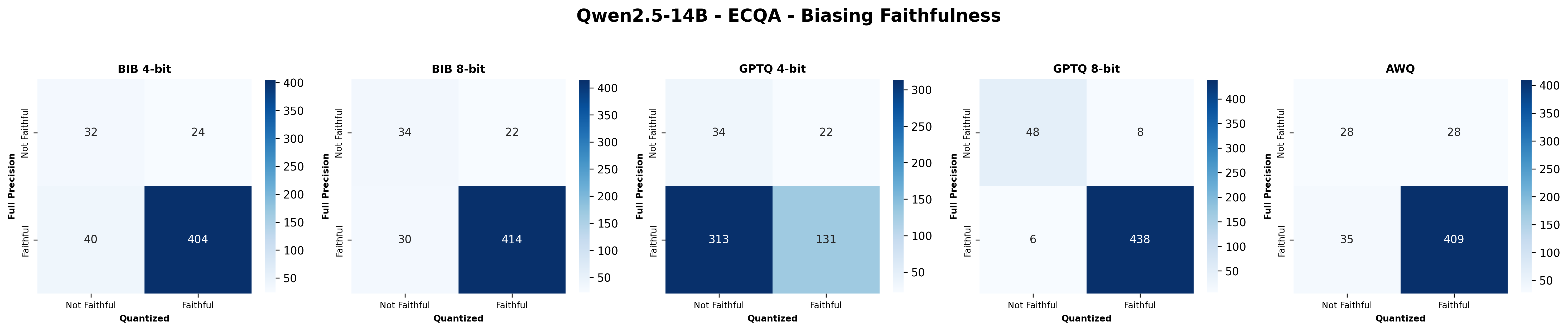}
\caption{\data{ECQA}}

  \end{subfigure}

    \caption{Faithfulness Variation \lm{Qwen2.5-14B} (\S\ref{subsub:discussion})}
  \label{fig:faith_14b}
\end{figure*}

\begin{figure*}[t!]
  \centering

    \begin{subfigure}{\textwidth}
    \centering
    \centering
\includegraphics[width=\linewidth]{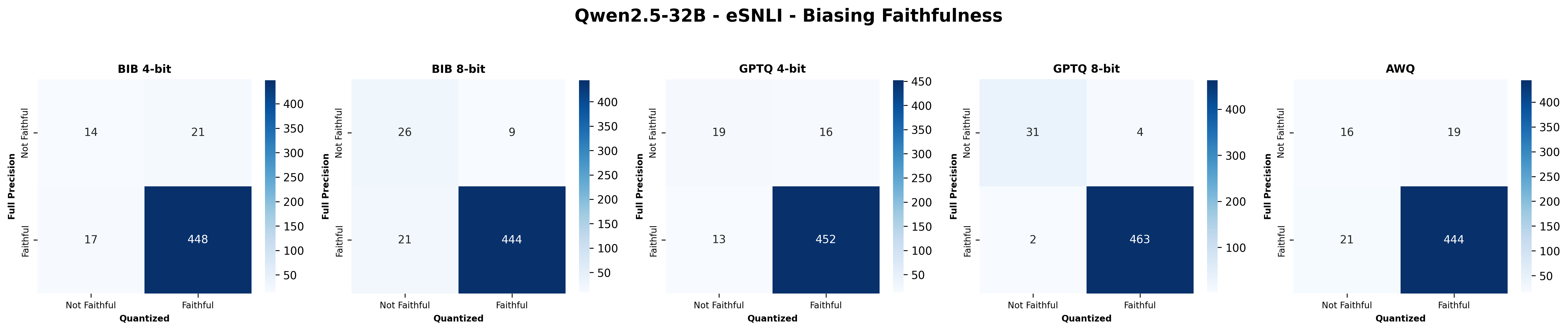}
\caption{\data{eSNLI}}
  \end{subfigure}
  
  \hfill

    \begin{subfigure}{\textwidth}
    \centering
    \centering
\includegraphics[width=\linewidth]{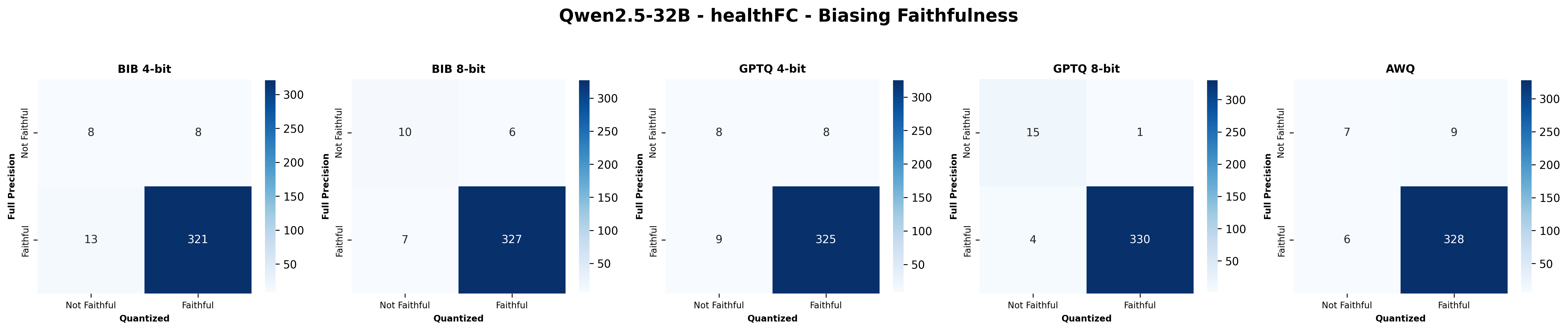}
\caption{\data{HealthFC}}

  \end{subfigure}
  
  \hfill

  \begin{subfigure}{\textwidth}
    \centering
    \centering
\includegraphics[width=\linewidth]{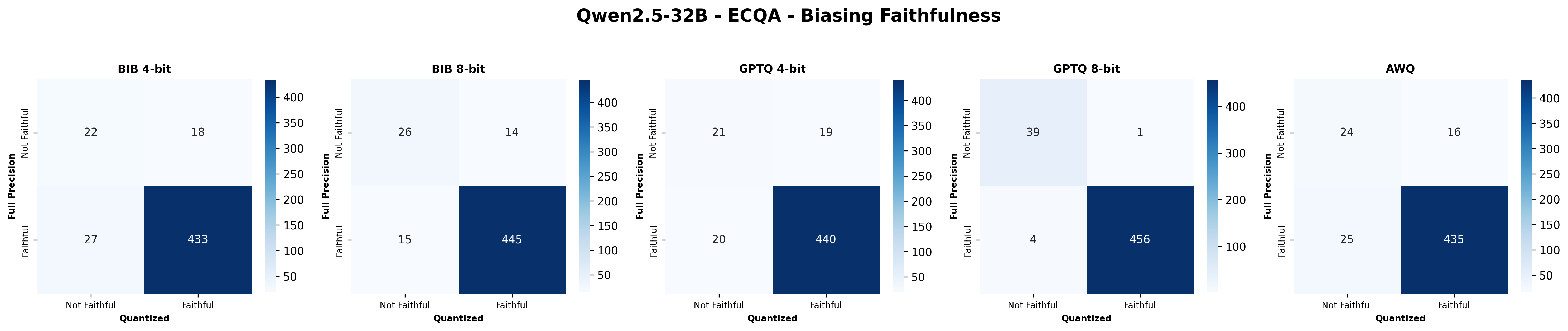}
\caption{\data{ECQA}}

  \end{subfigure}

    \caption{Faithfulness Variation \lm{Qwen2.5-32B} (\S\ref{subsub:discussion})}
  \label{fig:faith_32b}
\end{figure*}

\begin{figure*}[t!]
  \centering

    \begin{subfigure}{\textwidth}
    \centering
    \centering
\includegraphics[width=\linewidth]{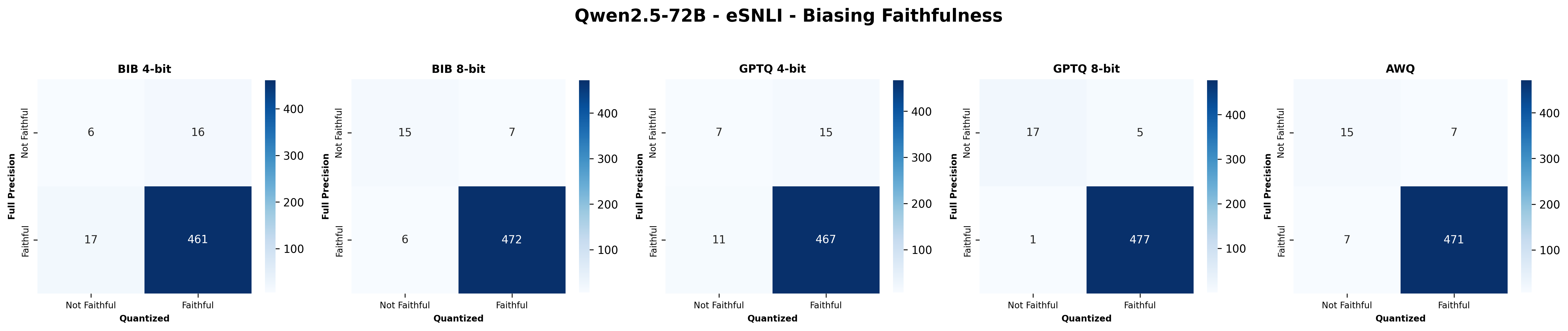}
\caption{\data{eSNLI}}
  \end{subfigure}
  
  \hfill

    \begin{subfigure}{\textwidth}
    \centering
    \centering
\includegraphics[width=\linewidth]{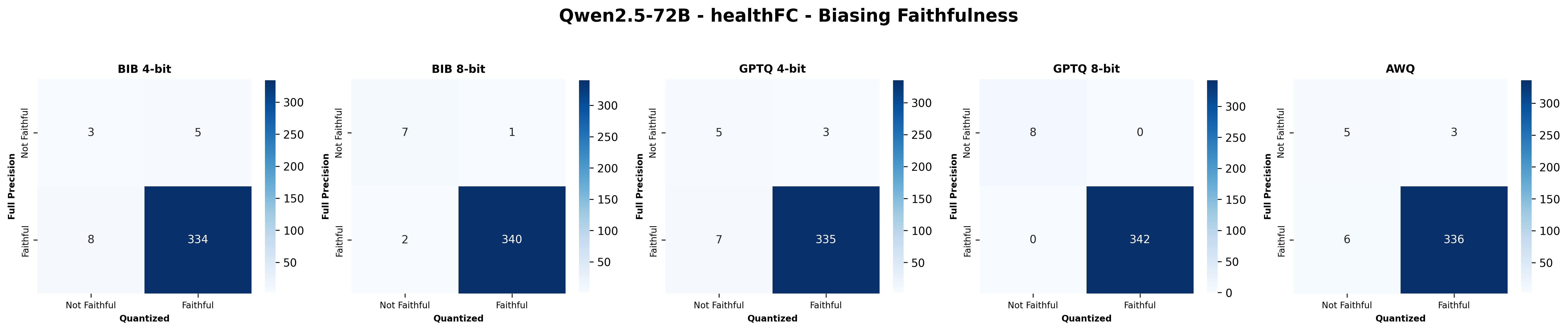}
\caption{\data{HealthFC}}

  \end{subfigure}
  
  \hfill

  \begin{subfigure}{\textwidth}
    \centering
    \centering
\includegraphics[width=\linewidth]{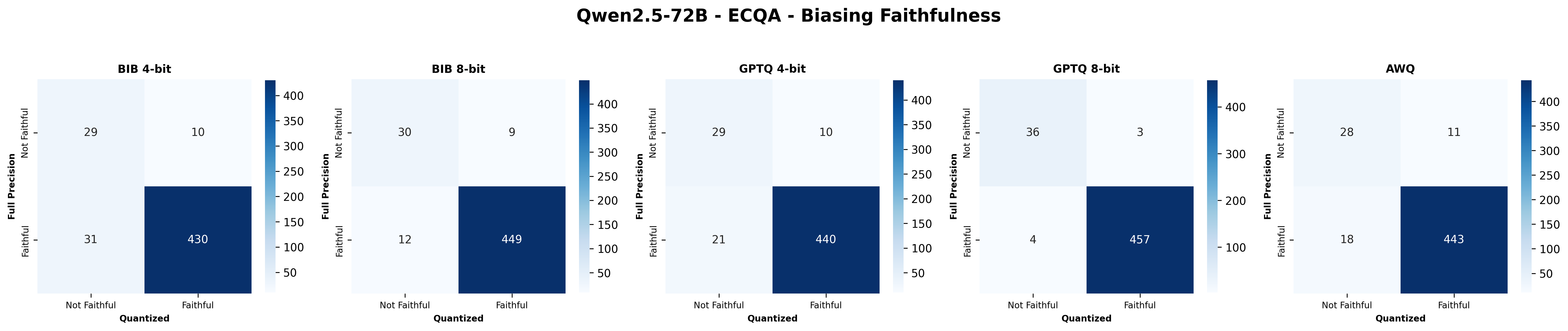}
\caption{\data{ECQA}}

  \end{subfigure}

    \caption{Faithfulness Variation \lm{Qwen2.5-72B} (\S\ref{subsub:discussion})}
  \label{fig:faith_72b}
\end{figure*}

\begin{figure*}[t!]
  \centering

    \begin{subfigure}{\textwidth}
    \centering
    \centering
\includegraphics[width=\linewidth]{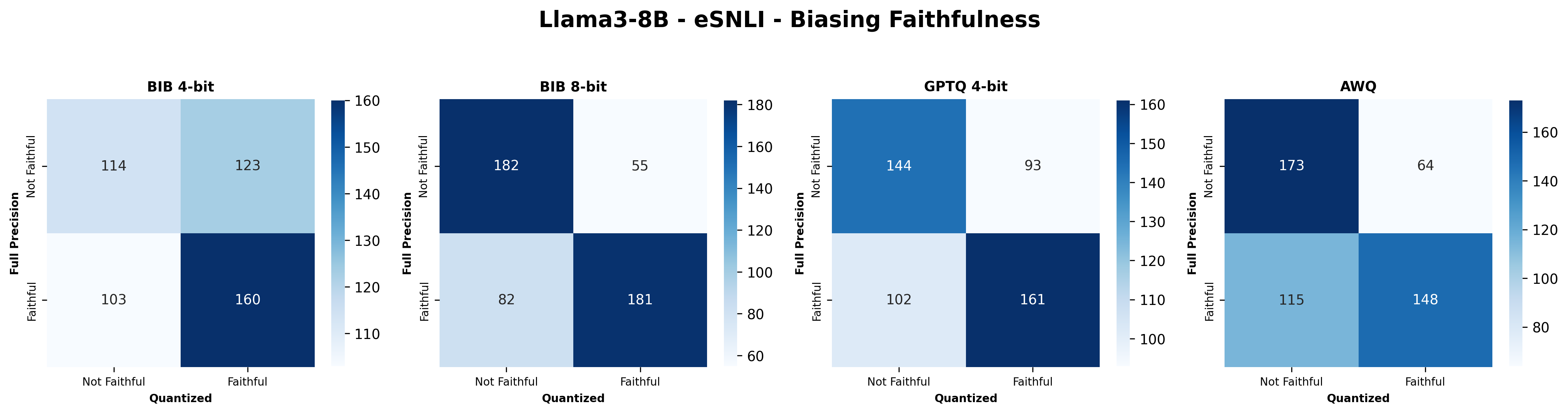}
\caption{\data{eSNLI}}
  \end{subfigure}
  
  \hfill

    \begin{subfigure}{\textwidth}
    \centering
    \centering
\includegraphics[width=\linewidth]{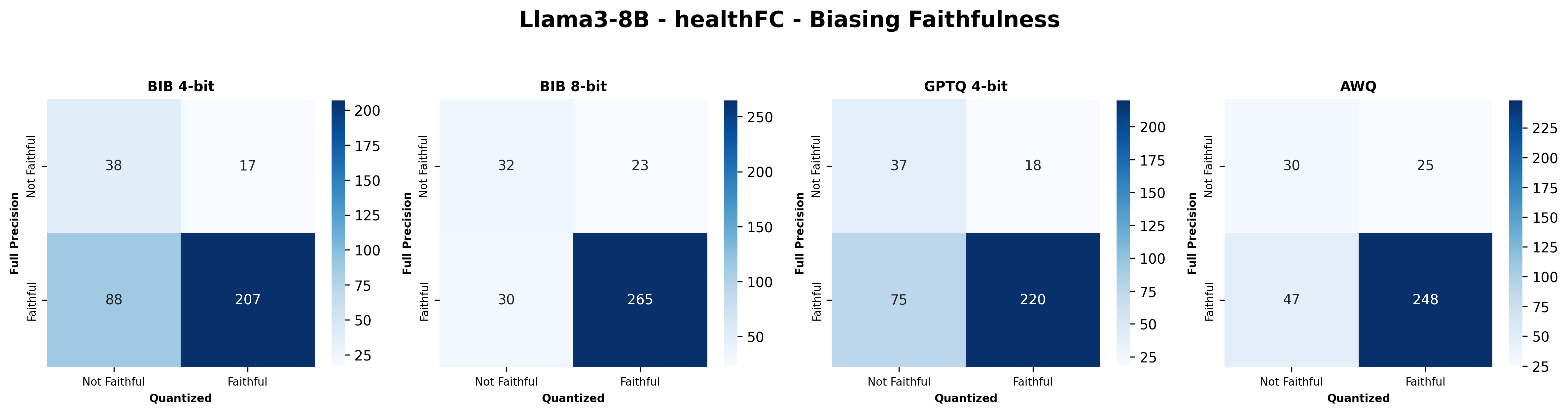}
\caption{\data{HealthFC}}

  \end{subfigure}
  
  \hfill

  \begin{subfigure}{\textwidth}
    \centering
    \centering
\includegraphics[width=\linewidth]{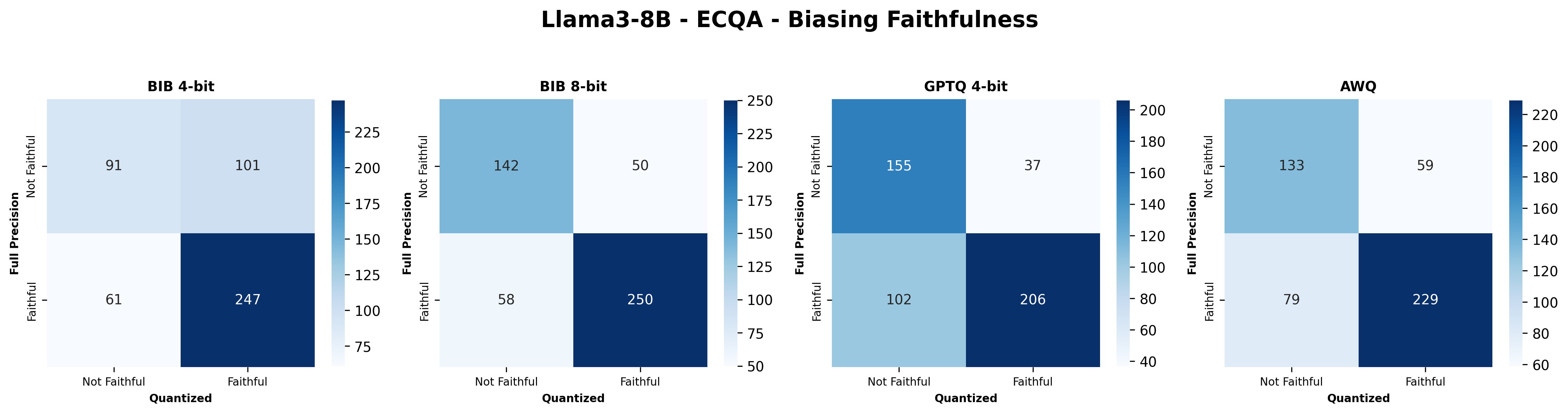}
\caption{\data{ECQA}}

  \end{subfigure}

    \caption{Faithfulness Variation \lm{Llama3-8B} (\S\ref{subsub:discussion})}
  \label{fig:faith_8b}
\end{figure*}

\begin{figure*}[t!]
  \centering

    \begin{subfigure}{\textwidth}
    \centering
    \centering
\includegraphics[width=\linewidth]{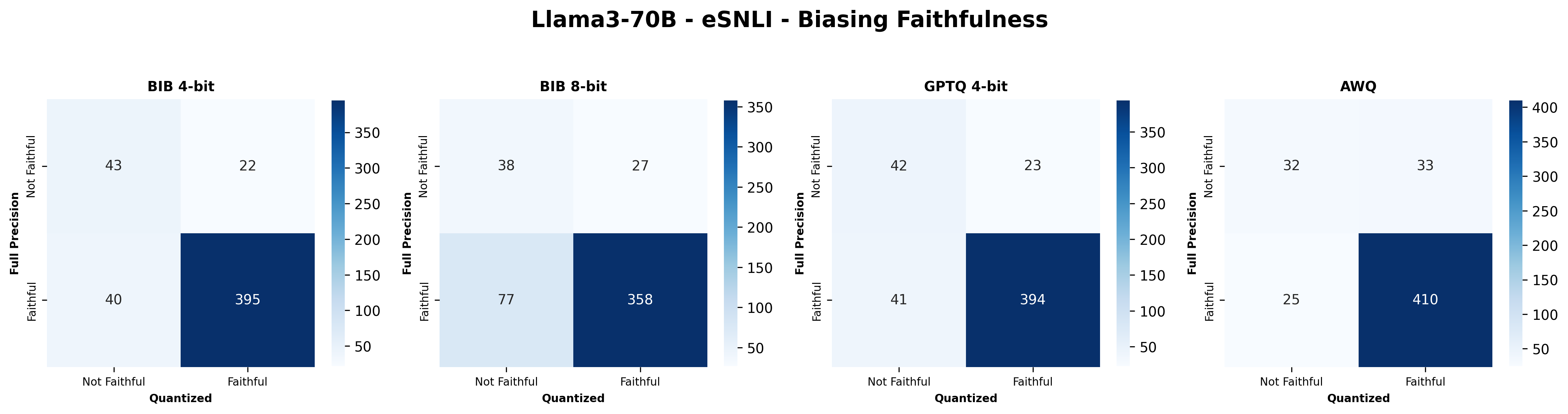}
\caption{\data{eSNLI}}
  \end{subfigure}
  
  \hfill

    \begin{subfigure}{\textwidth}
    \centering
    \centering
\includegraphics[width=\linewidth]{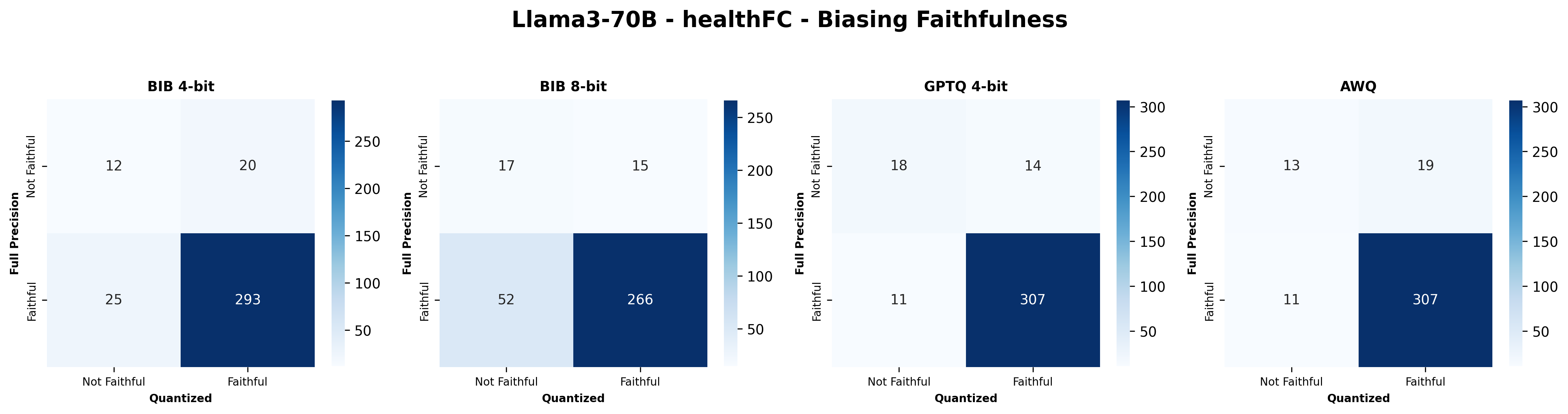}
\caption{\data{HealthFC}}

  \end{subfigure}
  
  \hfill

  \begin{subfigure}{\textwidth}
    \centering
    \centering
\includegraphics[width=\linewidth]{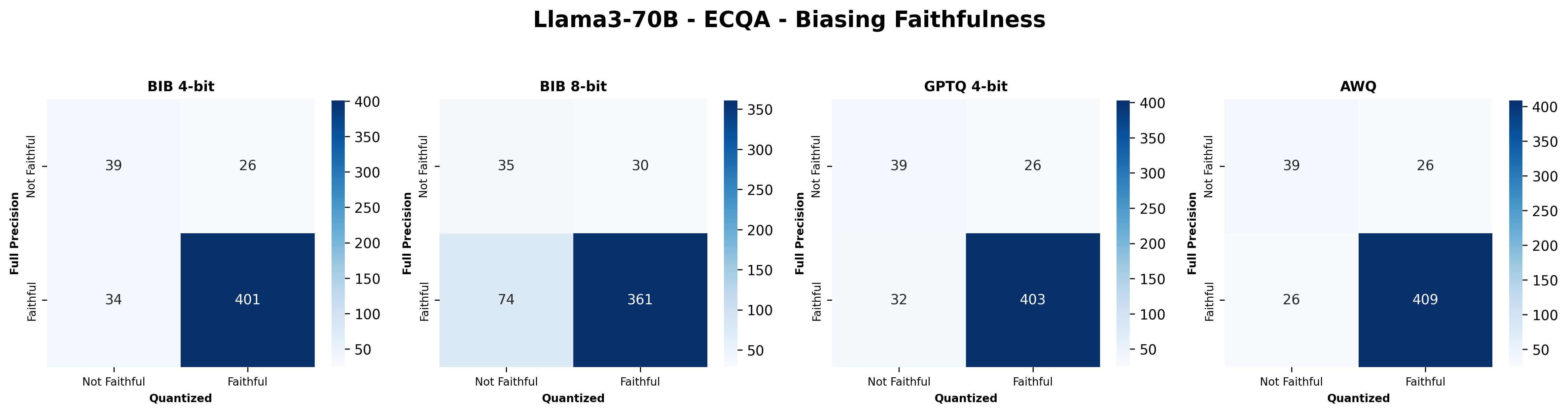}
\caption{\data{ECQA}}

  \end{subfigure}

    \caption{Faithfulness Variation \lm{Llama3-70B} (\S\ref{subsub:discussion})}
  \label{fig:faith_70b}
\end{figure*}

Table~\ref{tab:biasing-transitions} shows the natural language explanation faithfulness measured by biasing features. We observe that faithfulness is largely preserved, as evidenced by the predominant portion of instances maintaining their original state (faithful $\rightarrow$ faithful and unfaithful $\rightarrow$ unfaithful). Nevertheless, a greater number of cases exist in which natural language explanations become unfaithful due to quantization. More fine-grained faithfulness transitions are shown in Figure~\ref{fig:faith_7b}, \ref{fig:faith_14b}, \ref{fig:faith_32b}, \ref{fig:faith_72b}, \ref{fig:faith_8b} and \ref{fig:faith_70b}.

% Figure~\ref{fig:spearman_faithfulness} further shows the Spearmann correlation between employed faithfulness matrices (\S\ref{subsec:faithful}). We find that faithfulness as measured by the counterfactual test is moderately correlated with that measured by biasing features, while CC-SHAP produces divergent faithfulness assessments relative to the other two metrics.

% --------------------------------------------------
% --------------------------------------------------

% --------------------------------------------------
% --------------------------------------------------

% --------------------------------------------------
% --------------------------------------------------

% --------------------------------------------------
% --------------------------------------------------

% --------------------------------------------------
% --------------------------------------------------

% --------------------------------------------------
% --------------------------------------------------

% \begin{figure*}[t!]
%     \centering
%     \begin{minipage}{\textwidth}
%         \begin{minipage}[t]{0.5\textwidth}
%             \centering
%             \includegraphics[width=\textwidth]{figure/faithfulness/spearman_eSNLI.png}
%        \subcaption{\data{eSNLI}}
%         \end{minipage}%
%         \hfill%
%         \begin{minipage}[t]{0.5\textwidth}
%             \centering
%             \includegraphics[width=\textwidth]{figure/faithfulness/spearman_healthFC.png}
%             \subcaption{\data{HealthFC}}
%         \end{minipage}
%     \end{minipage}
    
% \caption{Spearman correlation matrices across all faithfulness metrics on \data{eSNLI} and \data{HealthFC}.}
%   \label{fig:spearman_faithfulness}
% \end{figure*}

\section{Quantization Affects Explanation Quality Differently From Faithfulness?}
\label{app:why}
Self-explanation quality and faithfulness are separate phenomena. We provide the following hypothesis regarding the different behavior of quality and faithfulness under quantization, i.e., under quantization, weights may be more sensitive compared to the decision boundary due to overparametrization, indicating that faithfulness can be better preserved compared to text quality. However, rigorously validating this hypothesis would constitute a substantial independent investigation beyond the scope of the present work.

Self-explanation quality, e.g., coherence or trustworthiness, depends heavily on the language modeling head and the token-by-token generation process, i.e., the precise logit distributions over vocabulary. These distributions are highly sensitive to small perturbations in weight values, because the softmax over a large vocabulary amplifies even minor shifts in pre-softmax scores. Quantization introduces exactly this kind of uniform noise across weights, degrading the fine-grained distributional properties that make text read coherent and well-structured \cite{resendiz2025llmbasedaffectivetextgeneration}.

Faithfulness, by contrast, reflects whether an explanation causally tracks the model's actual decision-making process, contingent upon the decision boundary in the representation space. Quantization perturbs weights but tends to preserve the topology of these boundaries, particularly in overparameterized models, where numerous weight configurations yield functionally equivalent decision surfaces. Consequently, faithfulness exhibits lower sensitivity to quantization than explanation quality.

\end{document}

%% file: table/automatic_evaluation.tex
\begin{table*}[t!]
    \centering
    \renewcommand*{\arraystretch}{1}
    
    \footnotesize
    \resizebox{\textwidth}{!}{%
        \begin{tabular}{c|c|cc|cc||ccc|ccc}

        \toprule[1.5pt]
          \multirow{2}{*}{\textbf{Model}} & \textbf{Preci} & \multicolumn{2}{c|}{\textbf{NLE (\data{eSNLI})}} & \multicolumn{2}{c||}{\textbf{NLE (\data{HealthFC})}} & \multicolumn{3}{c|}{\textbf{CFE (\data{eSNLI})}} & \multicolumn{3}{c}{\textbf{CFE (\data{AG News})}}\\
         
         & \textbf{-sion}  & BARTScore $\uparrow$ & TIGERScore $\uparrow$ & BARTScore $\uparrow$ & TIGERScore $\uparrow$ & LFR $\uparrow$ & PPL $\downarrow$ &TS $\downarrow$ & LFR $\uparrow$ & PPL $\downarrow$ & TS $\downarrow$\\
         % & \textbf{-sion}  & Score $\uparrow$ & Score $\uparrow$ & Score $\uparrow$ & Score $\uparrow$ & &  & \\

         \midrule

        \centering \multirow{5}{*}{\rotatebox[origin=r]{90}{\scriptsize{\lm{Qwen2.5-7B}}}} & \cellcolor[HTML]{F2CFC2}full
             & \cellcolor[HTML]{F2CFC2}-6.56 & \cellcolor[HTML]{F2CFC2}\textbf{-0.13}  & \cellcolor[HTML]{F2CFC2}-4.41 & \cellcolor[HTML]{F2CFC2}\textbf{-0.34} &  \cellcolor[HTML]{F2CFC2}64.80\% & \cellcolor[HTML]{F2CFC2}94.34 & \cellcolor[HTML]{F2CFC2}\textbf{0.37} & \cellcolor[HTML]{F2CFC2}35.40\% & \cellcolor[HTML]{F2CFC2}74.87 & \cellcolor[HTML]{F2CFC2}\textbf{0.53}\\
        & bib4
             & -6.69 & -0.22 & -4.35 &  \underline{-0.64} & 67.40\% & \textbf{\underline{64.99}} & 0.57 & 36.00\% & 95.26 & 0.61 \\
        & bib8
             & -6.56 & -0.24  & -7.84 &  -0.94 & 65.40\% & 88.85 & \underline{0.37} & \textbf{\underline{42.00\%}} & \textbf{\underline{72.97}} & 0.61 \\
        & gptq4
             & \textbf{\underline{-6.53}} & -0.51 & -4.29 & -0.65 & \textbf{\underline{68.60\%}} & 92.39 & 0.57 & 34.80\% & 81.52 &  \underline{0.57}\\
        & gptq8
          & -6.60 & -0.16 & -6.02 & -0.74 & 66.00\% & 79.48 & 0.39 & 31.20\% & 79.45 & 0.57 \\
        & awq
          & -6.59 & \underline{-0.13} & \textbf{\underline{-4.26}} & -0.93 & 67.60\% & 99.97 & 0.35 & 38.40\% & 80.52 & 0.57 \\

        \midrule

        % Qwen2.5-14B
         \centering \multirow{5}{*}{\rotatebox[origin=r]{90}{\scriptsize{\lm{Qwen2.5-14B}}}} & \cellcolor[HTML]{F2CFC2}full
           & \cellcolor[HTML]{F2CFC2}\textbf{-6.52} & \cellcolor[HTML]{F2CFC2}-0.25 & \cellcolor[HTML]{F2CFC2}-4.25 & \cellcolor[HTML]{F2CFC2}-0.39 & \cellcolor[HTML]{F2CFC2}67.20\% & \cellcolor[HTML]{F2CFC2}\textbf{90.67} & \cellcolor[HTML]{F2CFC2}0.54 & \cellcolor[HTML]{F2CFC2}38.40\% & \cellcolor[HTML]{F2CFC2}102.83 & \cellcolor[HTML]{F2CFC2}0.66\\
        
        & bib4
             & -6.73 & -0.24 & -4.30 & -0.29 & \textbf{\underline{67.60\%}} & 93.03 & 0.59 & \textbf{\underline{42.20\%}} & 133.47 & 0.78 \\
        & bib8
             & -6.59 & \textbf{\underline{-0.21}} & -4.34 & \textbf{\underline{-0.24}} & 64.40\% & 93.54 & 0.53 & 34.60\% & 99.28 & 0.63 \\
        & gptq4
         & -6.59 & -0.25 & \textbf{\underline{-4.23}} & -0.41 & 64.40\% & 91.45 & 0.53 & 39.40\% & 99.97 & 0.66 \\
        & gptq8
          & \underline{-6.54} & -0.30 & -4.27 & -0.31 & 63.60\% & 92.58 & 0.51 & 36.80\% & 95.11 & 0.66 \\
        & awq
          & -6.57 & -0.81 & -4.33 & -0.48 & 67.60\% & \underline{91.44} & \textbf{\underline{0.47}} & 39.20\% & \textbf{\underline{84.55}} & \textbf{\underline{0.61}} \\

         \midrule

        % Qwen2.5-32B
         \centering \multirow{5}{*}{\rotatebox[origin=r]{90}{\scriptsize{\lm{Qwen2.5-32B}}}} & \cellcolor[HTML]{F2CFC2}full
            & \cellcolor[HTML]{F2CFC2}-7.68 & \cellcolor[HTML]{F2CFC2}-0.44 & \cellcolor[HTML]{F2CFC2}-8.17 & \cellcolor[HTML]{F2CFC2}-2.68 & \cellcolor[HTML]{F2CFC2}64.20\% & \cellcolor[HTML]{F2CFC2}87.07 & \cellcolor[HTML]{F2CFC2}0.43 & \cellcolor[HTML]{F2CFC2}31.40\% & \cellcolor[HTML]{F2CFC2}83.13 & \cellcolor[HTML]{F2CFC2}0.49\\

        & bib4
             & -10.50 & -1.28  & \textbf{\underline{-6.06}} & \textbf{\underline{-1.61}} & \textbf{\underline{64.80\%}} & \textbf{\underline{79.35}} & 0.43 & \textbf{\underline{39.20\%}} & 80.93 & 0.56 \\
        & bib8
             & -8.77 & -0.70 & -9.07 & -3.00 & 64.00\% & 86.11 & 0.41 & 34.00\% & 82.82 & 0.49 \\
        & gptq4
             & \textbf{\underline{-6.61}} & -0.56 & -6.62 & -1.88 & 63.60\% & 87.07 & \textbf{\underline{0.41}}  & 33.40\% & 81.48 & \textbf{\underline{0.48}} \\
        & gptq8
          & -7.91 & -0.90 & -8.54 & -2.86 & 63.60\% & 90.03 & 0.43 & 35.40\% & 84.81 & 0.49 \\
        & awq
          & -9.91 & \textbf{\underline{-0.27}} & -8.97 & -3.27 & 63.40\% & 92.82 & 0.41 & 37.20\% & \textbf{\underline{80.60}} & 0.52 \\

        \midrule

        % Qwen2.5-72B
         \centering \multirow{5}{*}{\rotatebox[origin=r]{90}{\scriptsize{\lm{Qwen2.5-72B}}}} & \cellcolor[HTML]{F2CFC2}full
         & \cellcolor[HTML]{F2CFC2}-6.52 & \cellcolor[HTML]{F2CFC2}-0.47 & \cellcolor[HTML]{F2CFC2}-4.21 & \cellcolor[HTML]{F2CFC2}-0.58 &  \cellcolor[HTML]{F2CFC2}61.20\% & \cellcolor[HTML]{F2CFC2}117.85 & \cellcolor[HTML]{F2CFC2}0.39 & \cellcolor[HTML]{F2CFC2}28.80\% & \cellcolor[HTML]{F2CFC2}96.45 & \cellcolor[HTML]{F2CFC2}0.43\\
        
        & bib4
             & -6.54 & \textbf{\underline{-0.39}} & -4.21 & -0.58 & 63.60\% & 120.36 & 0.44 & 25.40\% & 93.03 & 0.45 \\
        & bib8
            &\textbf{\underline{-6.51}} & -0.52 & -4.22 & \textbf{\underline{-0.57}} & 61.60\% & \textbf{\underline{112.60}} & 0.40 & \textbf{\underline{31.20\%}} & 91.51 & 0.44 \\
        & gptq4
            & -6.52 & -0.47 & \textbf{\underline{-4.17}} & -0.60 & 61.40\% & 119.59 & 0.39 & 29.60\% & 96.05 & 0.44 \\
        & gptq8
          & -6.55 & -0.53 & -4.20 & -0.65 & 61.20\% & 115.94 & 0.41 & 26.40\% & 96.47 & 0.45 \\
        & awq
          & -6.52 & -0.51  & -4.22 & -0.66 & \textbf{\underline{64.00\%}} & 119.61 & \textbf{\underline{0.37}} & 24.60\% & \textbf{\underline{88.97}} & \textbf{\underline{0.42}} \\   

         \midrule

        % llama3-8b 
         \centering \multirow{5}{*}{\rotatebox[origin=c]{90}{\scriptsize{\lm{Llama3-8B}}}} & \cellcolor[HTML]{F2CFC2}full
             & \cellcolor[HTML]{F2CFC2}-6.62 & \cellcolor[HTML]{F2CFC2}-0.37 & \cellcolor[HTML]{F2CFC2}-4.29 & \cellcolor[HTML]{F2CFC2}-1.01 & \cellcolor[HTML]{F2CFC2}66.00\% & \cellcolor[HTML]{F2CFC2}\textbf{62.26} & \cellcolor[HTML]{F2CFC2}\textbf{0.41} & \cellcolor[HTML]{F2CFC2}48.80\% & \cellcolor[HTML]{F2CFC2}\textbf{42.88} & \cellcolor[HTML]{F2CFC2}1.54\\
        % \cdashline{2-8}
        & bib4
             & \textbf{\underline{-6.60}} & -0.54 & -4.65 & -1.30 & 63.60\% & 81.80 & 0.42  & 48.40\% & \underline{46.01} & 1.38 \\
        & bib8
            & -6.66 & -0.37  & \textbf{\underline{-4.26}} & \textbf{\underline{-0.95}} & 63.40\% & \underline{72.17} & \underline{0.41} & 51.60\% & 46.95 & 1.60 \\
        & gptq4
             & -7.42 & -0.29  & -4.36 & -1.28 & \textbf{\underline{67.60\%}} & 76.55 & 0.54 & \textbf{\underline{65.20\%}} & 62.61 & 1.60 \\
        & awq
          & -6.76 & \textbf{\underline{-0.26}}  & -4.43 & -1.31 & 67.60\% & 75.21 & 0.53 & 54.00\% & 79.87 & \textbf{\underline{1.21}} \\

         \midrule

        % llama3-70b 
        \centering \multirow{5}{*}{\rotatebox[origin=c]{90}{\scriptsize{\lm{Llama3-70B}}}} & \cellcolor[HTML]{F2CFC2}full
             & \cellcolor[HTML]{F2CFC2}\textbf{-6.58} & \cellcolor[HTML]{F2CFC2}-0.16  & \cellcolor[HTML]{F2CFC2}-4.48 & \cellcolor[HTML]{F2CFC2}-1.04  & \cellcolor[HTML]{F2CFC2}64.80\% & \cellcolor[HTML]{F2CFC2}\textbf{80.77} & \cellcolor[HTML]{F2CFC2}\textbf{0.41} & \cellcolor[HTML]{F2CFC2}49.00\% & \cellcolor[HTML]{F2CFC2}\textbf{101.97} & \cellcolor[HTML]{F2CFC2}0.40 \\
        & bib4
             & -6.62 & -0.71  & -4.44 & -1.56 & 62.40\% & 139.80 & 0.47 & 49.60\% & 130.05 & 0.48 \\
        & bib8
             & -6.76 & -0.26  & \textbf{\underline{-4.18}} & -1.30 & \textbf{\underline{68.80\%}} & 102.23 & 0.45 & 46.20\% & 140.24 & \textbf{\underline{0.25}} \\
        & gptq4
             & -6.62 & \textbf{\underline{-0.14}}  & -4.32 & -1.12 & 63.27\% & \underline{94.48} & 0.46 & \textbf{\underline{53.20\%}} & 115.03 & 0.52 \\
        & awq
          & \underline{-6.61}  & -0.15  & -4.54 & \textbf{\underline{-1.01}} & 64.80\% & 114.89 & \underline{0.45} & 47.40\% & \underline{106.90} & 0.42 \\
        
        \toprule[1.5pt]
        \end{tabular}
    }
    \caption{Automatic evaluation results of NLEs and CFEs generated by \lm{Llama3} (8B, 70B) and \lm{Qwen2.5} (7B, 14B, 32B, 72B) models with full precision and different quantization methods. For NLEs, we use \texttt{ZeroCoT} on \data{eSNLI} and \data{HealthFC}, evaluated by \textit{BARTScore} and \textit{TIGERScore}. For CFEs, we use \texttt{FIZLE} on \data{eSNLI} and \data{AG News}, evaluated by \textit{label flip rate (LFR)}, \textit{perplexity (PPL)}, and \textit{text similarity (TS)}. \textbf{Bold} values indicate the best-performing approach for each model, while \underline{underlined} values denote the best-performing quantization method. Table~\ref{tab:automatic_evaluation_additional} further provides results for \data{ECQA} and \data{IMDb}.
    }
    \label{tab:counterfactual_and_explanation_automatic_evaluation}
    % \vspace{-1em}
\end{table*}

%% file: table/faithfulness.tex
\begin{table}[t!]
    \centering
    \renewcommand*{\arraystretch}{1}
    \footnotesize
    \resizebox{\columnwidth}{!}{%
        \begin{tabular}{>{\centering\arraybackslash}p{0.3cm}|c|cc|cc|cc}
        \toprule[1.5pt]
         \multirow{2}{*}{\rotatebox[origin=c]{90}{\textbf{\tiny{Model}}}} & \textbf{\scriptsize{Preci}} & \multicolumn{2}{c|}{\textbf{\data{eSNLI}}} & \multicolumn{2}{c|}{\textbf{\data{HealthFC}}} & \multicolumn{2}{c}{\textbf{\data{ECQA}}}\\
         & \textbf{\scriptsize{-sion}} & \textbf{Bias} & \textbf{CC-SHAP} & \textbf{Bias} & \textbf{CC-SHAP} & \textbf{Bias} & \textbf{CC-SHAP}\\

         \midrule
    
           % ---------------- Qwen2.5-7B ----------------
        \multirow{6}{*}{\scriptsize{\rotatebox[origin=r]{90}{\lm{Qwen2.5-7B}}}}
          & \cellcolor[HTML]{F2CFC2}full  & \cellcolor[HTML]{F2CFC2}90.20 & \cellcolor[HTML]{F2CFC2}0.843 & \cellcolor[HTML]{F2CFC2}\textbf{93.43} & \cellcolor[HTML]{F2CFC2}0.772 & \cellcolor[HTML]{F2CFC2}0.872 & \cellcolor[HTML]{F2CFC2}0.039\\
          & bib4   & 85.60 & \textbf{\underline{0.852}} & 91.14 & 0.765 & 0.902 & \underline{\textbf{0.107}}\\
          & bib8   & \textbf{\underline{91.20}} & 0.849 & \underline{93.14} & 0.770 & 0.884 &  0.010\\
          & gptq4  & 86.20 & 0.840 & 92.86 & \textbf{\underline{0.774}} & \textbf{\underline{0.904}} & 0.021\\
          & gptq8  & 89.60 & 0.844 & \underline{93.14} & 0.768 & 0.878 & 0.033\\
          & awq    & 85.00 & 0.821 & 92.29 & 0.764 & 0.886 & -0.033\\
        \midrule
    
        % ---------------- Qwen2.5-14B ----------------
        \multirow{6}{*}{\scriptsize{\rotatebox[origin=c]{90}{\lm{Qwen2.5-14B}}}}
          & \cellcolor[HTML]{F2CFC2}full  & \cellcolor[HTML]{F2CFC2}90.00 & \cellcolor[HTML]{F2CFC2}0.819 & \cellcolor[HTML]{F2CFC2}94.29 & \cellcolor[HTML]{F2CFC2}0.804 & \cellcolor[HTML]{F2CFC2}0.888 & \cellcolor[HTML]{F2CFC2}0.170\\
          & bib4   & 91.00 & 0.832 & \textbf{\underline{95.14}} & 0.797 & 0.856 & \underline{\textbf{0.237}} \\
          & bib8   & 89.20 & 0.790 & 93.14 & 0.803 & 0.872 & 0.196\\
          & gptq4  & 41.80 & \textbf{\underline{0.895}} & 49.43 & \textbf{\underline{0.900}} & 0.306 & 0.164 \\
          & gptq8  & 92.00 & 0.820 & 94.00 & 0.808 & \textbf{\underline{0.892}} & 0.160 \\
          & awq    & \textbf{\underline{93.00}} & 0.845 & 93.43 & 0.771 & 0.874 & 0.153 \\
        \midrule
    
        % ---------------- Qwen2.5-32B ----------------
        \multirow{6}{*}{\scriptsize{\rotatebox[origin=c]{90}{\lm{Qwen2.5-32B}}}}
          & \cellcolor[HTML]{F2CFC2}full  & \cellcolor[HTML]{F2CFC2}93.00 & \cellcolor[HTML]{F2CFC2}0.835 & \cellcolor[HTML]{F2CFC2}95.43 & \cellcolor[HTML]{F2CFC2}0.808 & \cellcolor[HTML]{F2CFC2}\textbf{0.920} & \cellcolor[HTML]{F2CFC2}\textbf{0.553}\\
          & bib4   & \textbf{\underline{93.80}} & 0.839 & 94.00 & 0.815 & 0.902 &  0.298\\
          & bib8   & 90.60 & \textbf{\underline{0.842}} & 95.14 & 0.805 & \underline{0.918} & \underline{0.440}\\
          & gptq4  & 93.60 & \textbf{\underline{0.842}} & 95.14 & \textbf{\underline{0.820}} & \underline{0.918} & 0.379\\
          & gptq8  & 93.40 & 0.835 & 94.57 & 0.813 & 0.914 & 0.401\\
          & awq    & 92.60 & 0.840 & \textbf{\underline{96.29}} & 0.806 & 0.902 & 0.330 \\
        \midrule
        
        % ---------------- Qwen2.5-72B ----------------
        \multirow{6}{*}{\scriptsize{\rotatebox[origin=c]{90}{\lm{Qwen2.5-72B}}}}
          & \cellcolor[HTML]{F2CFC2}full  & \cellcolor[HTML]{F2CFC2}95.60 & \cellcolor[HTML]{F2CFC2}0.876 & \cellcolor[HTML]{F2CFC2}\textbf{97.71} & \cellcolor[HTML]{F2CFC2}\textbf{0.826} & \cellcolor[HTML]{F2CFC2}\textbf{0.922} & \cellcolor[HTML]{F2CFC2}\textbf{0.797}\\
          & bib4   & 95.40 & \textbf{\underline{0.884}} & 96.86 & 0.821 & 0.880 & 0.675 \\
          & bib8   & 95.80 & 0.874 & 97.43 & \underline{0.824} & 0.916 & \underline{0.714} \\
          & gptq4  & \textbf{\underline{96.40}} & 0.875 & 96.57 & 0.817 & 0.900 & 0.692\\
          & gptq8  & \textbf{\underline{96.40}} & 0.880 & \textbf{\underline{97.71}} & 0.821 & \underline{0.920} & 0.663\\
          & awq    & 95.60 & 0.875 & 96.86 & 0.813 & 0.908 & 0.548 \\
        \midrule
                  
        % ---------------- Llama3-8B ----------------
        \multirow{5}{*}{\scriptsize{\rotatebox[origin=c]{90}{\lm{Llama3-8B}}}}
          & \cellcolor[HTML]{F2CFC2}full  & \cellcolor[HTML]{F2CFC2}52.60 & \cellcolor[HTML]{F2CFC2}0.742 & \cellcolor[HTML]{F2CFC2}\textbf{84.30} & \cellcolor[HTML]{F2CFC2}0.745 & \cellcolor[HTML]{F2CFC2}0.616 & \cellcolor[HTML]{F2CFC2}0.946\\
          & bib4   & \textbf{\underline{56.60}} & \textbf{\underline{0.776}} & 64.00 & \textbf{\underline{0.749}} & \textbf{\underline{0.696}} & \underline{\textbf{0.994}}\\
          & bib8   & 47.20 & 0.712 & \underline{82.30} & 0.736 & 0.600 & 0.954\\
          & gptq4  & 50.80 & 0.651 & 68.00 & 0.584 & 0.486 & 0.711\\
          & awq    & 42.40 & 0.682 & 78.00 & 0.567 & 0.576 & 0.993\\
        \midrule
    
        % ---------------- Llama3-70B ----------------
        \multirow{5}{*}{\scriptsize{\rotatebox[origin=c]{90}{\lm{Llama3-70B}}}}
          & \cellcolor[HTML]{F2CFC2}full  & \cellcolor[HTML]{F2CFC2}87.00 & \cellcolor[HTML]{F2CFC2}\textbf{0.741} & \cellcolor[HTML]{F2CFC2}90.86 & \cellcolor[HTML]{F2CFC2}\textbf{0.409} & \cellcolor[HTML]{F2CFC2}\textbf{0.870} & \cellcolor[HTML]{F2CFC2}\textbf{0.895}\\
          & bib4   & 83.40 & 0.278 & 89.43 & 0.313 & 0.854 & 0.834\\
          & bib8   & 77.00 & \underline{0.628} & 80.29 & 0.319 & 0.782 & \underline{0.886}\\
          & gptq4  & 83.40 & 0.512 & 91.71 & \underline{0.407} & 0.858 & 0.523 \\
          & awq    & \textbf{\underline{88.60}} & 0.469 & \textbf{\underline{93.14}} & 0.368 & \textbf{\underline{0.870}} & 0.658 \\
        
        \toprule[1.5pt]
        \end{tabular}
    }
    \caption{Faithfulness rate (in \%) of natural language explanation evaluated using biasing features (Bias), and CC-SHAP values on \data{eSNLI}, \data{HealthFC}, and \data{ECQA} across various quantization configurations. \textbf{Bold} values indicate the best-performing approach for each model, while \underline{underlined} values denote the best-performing quantization method.}
    \label{tab:faithfulness}
    % \vspace{-1em}
\end{table}

%% file: table/model.tex
\begin{table*}[t!]
    \centering
    \resizebox{\textwidth}{!}{%
        \begin{tabular}{rcccc}

        \toprule
        \textbf{Name}& \textbf{Citation} & \textbf{Size} & \textbf{Precision} & \textbf{Link}\\

        % \midrule

        % \lm{Vicuna} & \cite{chiang-lee-2023-large} & 13B & full & \url{https://huggingface.co/lmsys/vicuna-13b-v1.5}\\

        % \lm{Vicuna} & \cite{chiang-lee-2023-large} & 13B & GPTQ4 & \url{https://huggingface.co/TheBloke/vicuna-13B-v1.5-GPTQ}\\

        % \lm{Vicuna} & \cite{chiang-lee-2023-large} & 13B & GPTQ8 & \scriptsize{\url{https://huggingface.co/TheBloke/vicuna-13B-v1.5-GPTQ/tree/gptq-8bit--1g-actorder_True}}\\

        % \lm{Vicuna} & \cite{chiang-lee-2023-large} & 13B & AWQ & \url{https://huggingface.co/TheBloke/vicuna-13B-v1.5-AWQ}\\

        \midrule
        \lm{Llama3-8B} & \citet{llama3modelcard} & 8B & Full & \url{https://huggingface.co/meta-llama/Meta-Llama-3-8B-Instruct}\\

        \lm{Llama3-8B} & \citet{llama3modelcard} & 2B & GPTQ4 & \url{https://huggingface.co/TechxGenus/Meta-Llama-3-8B-Instruct-GPTQ}\\

        \lm{Llama3-8B} & \citet{llama3modelcard} & 2B & AWQ & \url{https://huggingface.co/TechxGenus/Meta-Llama-3-8B-Instruct-AWQ}\\

        \midrule

        \lm{Llama3-70B} & \citet{llama3modelcard} & 70B & full & \url{https://huggingface.co/meta-llama/Meta-Llama-3-70B-Instruct}\\

        \lm{Llama3-70B} & \citet{llama3modelcard} & 11B & GPTQ4 & \url{https://huggingface.co/TechxGenus/Meta-Llama-3-70B-Instruct-GPTQ}\\

        \lm{Llama3-70B} & \citet{llama3modelcard} & 11B & AWQ & \url{https://huggingface.co/TechxGenus/Meta-Llama-3-70B-Instruct-AWQ}\\
        
        \midrule

        \lm{Qwen2.5-7B} & \citet{qwen2024qwen25technicalreport} & 7B & Full & \url{https://huggingface.co/Qwen/Qwen2.5-7B-Instruct}\\
        \lm{Qwen2.5-7B} & \citet{qwen2024qwen25technicalreport} & 2B & AWQ & \url{https://huggingface.co/Qwen/Qwen2.5-7B-Instruct-AWQ}\\
        \lm{Qwen2.5-7B} & \citet{qwen2024qwen25technicalreport} & 2B & GPTQ4 & \url{https://huggingface.co/Qwen/Qwen2.5-7B-Instruct-GPTQ-Int4}\\
        \lm{Qwen2.5-7B} & \citet{qwen2024qwen25technicalreport} & 3B & GPTQ8 & \url{https://huggingface.co/Qwen/Qwen2.5-7B-Instruct-GPTQ-Int8}\\

        \midrule
        
        \lm{Qwen2.5-14B} & \citet{qwen2024qwen25technicalreport} & 14B & Full & \url{https://huggingface.co/Qwen/Qwen2.5-14B-Instruct}\\
        \lm{Qwen2.5-14B} & \citet{qwen2024qwen25technicalreport} & 3B & AWQ & \url{https://huggingface.co/Qwen/Qwen2.5-14B-Instruct-AWQ}\\
        \lm{Qwen2.5-14B} & \citet{qwen2024qwen25technicalreport} & 3B & GPTQ4 & \url{https://huggingface.co/Qwen/Qwen2.5-14B-Instruct-GPTQ-Int4}\\
        \lm{Qwen2.5-14B} & \citet{qwen2024qwen25technicalreport} & 5B & GPTQ8 & \url{https://huggingface.co/Qwen/Qwen2.5-14B-Instruct-GPTQ-Int8}\\

        \midrule
        
        \lm{Qwen2.5-32B} & \citet{qwen2024qwen25technicalreport} & 32B & Full & \url{https://huggingface.co/Qwen/Qwen2.5-32B-Instruct}\\
        \lm{Qwen2.5-32B} & \citet{qwen2024qwen25technicalreport} & 6B & AWQ & \url{https://huggingface.co/Qwen/Qwen2.5-32B-Instruct-AWQ}\\
        \lm{Qwen2.5-32B} & \citet{qwen2024qwen25technicalreport} & 6B & GPTQ4 & \url{https://huggingface.co/Qwen/Qwen2.5-32B-Instruct-GPTQ-Int4}\\
        \lm{Qwen2.5-32B} & \citet{qwen2024qwen25technicalreport} & 10B & GPTQ8 & \url{https://huggingface.co/Qwen/Qwen2.5-32B-Instruct-GPTQ-Int8}\\
        
        \midrule
        \lm{Qwen2.5-72B} & \citet{qwen2024qwen25technicalreport} & 72B & Full & \url{https://huggingface.co/Qwen/Qwen2.5-72B-Instruct}\\
        \lm{Qwen2.5-72B} & \citet{qwen2024qwen25technicalreport} & 12B & AWQ & \url{https://huggingface.co/Qwen/Qwen2.5-72B-Instruct-AWQ}\\
        \lm{Qwen2.5-72B} & \citet{qwen2024qwen25technicalreport} & 12B & GPTQ4 & \url{https://huggingface.co/Qwen/Qwen2.5-72B-Instruct-GPTQ-Int4}\\
        \lm{Qwen2.5-72B} & \citet{qwen2024qwen25technicalreport} & 21B & GPTQ8 & \url{https://huggingface.co/Qwen/Qwen2.5-72B-Instruct-GPTQ-Int8}\\
        
        \bottomrule
        \end{tabular}
        }
    \caption{
    Detailed information about used LLMs in our experiments. 
    }
    \label{tab:used_model}
\end{table*}

%% file: table/attribution.tex
\begin{table*}[!t]
    \centering
        \resizebox{\textwidth}{!}{%
        \begin{tabular}{l|cc|cc|cc|cc|cc|cc}
        \toprule
        \multirow{2}{*}{\textbf{Quantization}} & \multicolumn{2}{c|}{\textbf{\lm{Qwen2.5-7B}}} & \multicolumn{2}{c|}{\textbf{\lm{Qwen2.5-14B}}} & \multicolumn{2}{c|}{\textbf{\lm{Qwen2.5-32B}}} & \multicolumn{2}{c|}{\textbf{\lm{Qwen2.5-72B}}} & \multicolumn{2}{c|}{\textbf{\lm{Llama3-8B}}} & \multicolumn{2}{c}{\textbf{\lm{Llama3-70B}}} \\ \cmidrule{2-13} 
         & \textbf{FR} & \textbf{Diff} & \textbf{FR} & \textbf{Diff} & \textbf{FR} & \textbf{Diff} & \textbf{FR} & \textbf{Diff} & \textbf{FR} & \textbf{Diff} & \textbf{FR} & \textbf{Diff} \\ \midrule
        Full-Precision & 0.362 & 0 & 0.568 & 0 & 0.624 & 0 & 0.562 & 0 & 0.442 & 0 & 0.514 & 0 \\
        AWQ & 0.362 & 0 & 0.6092 & 0.0412 & 0.612 & -0.012 & 0.546 & -0.016 & 0.428 & -0.014 & 0.544 & 0.03 \\
        GPTQ4 & 0.277 & -0.085 & 0.531 & -0.037 & 0.593 & -0.031 & 0.511 & -0.051 & 0.379 & -0.063 & 0.472 & -0.042 \\
        GPTQ8 & 0.283 & -0.079 & 0.552 & -0.016 & 0.599 & -0.025 & 0.52 & -0.042 & - & - & - & - \\
        bit4 & 0.278 & -0.084 & 0.5691 & 0.0011 & 0.604 & -0.02 & 0.538 & -0.024 & 0.3618 & -0.0802 & 0.4609 & -0.0531 \\
        bit8 & 0.344 & -0.018 & 0.552 & -0.016 & 0.62 & -0.004 & 0.554 & -0.008 & 0.436 & -0.006 & 0.481 & -0.033 \\ \midrule
        \textbf{Average} & - & \textbf{-0.0532} & - & \textbf{-0.00534} & - & \textbf{-0.0184} & - & \textbf{-0.0282} & - & \textbf{-0.0408} & - & \textbf{-0.0245} \\ \bottomrule
        \end{tabular}%
        }
    \caption{Faithfulness rate (FR) and the value differences between full-precision and quantized models on \data{eSNLI}.}
    \label{tab:attribution}
\end{table*}

%% file: table/automatic_evaluation_nle.tex
\begin{table*}[t!]
    \centering
    \renewcommand*{\arraystretch}{1}
    
    \footnotesize
    \resizebox{\textwidth}{!}{%
        \begin{tabular}{c|c|ccc|cc||c|c|ccc|cc}

        \toprule[1.5pt]
          \multirow{2}{*}{\textbf{Model}} & \textbf{Preci} & \multicolumn{3}{c|}{\textbf{CFE (\data{IMDb})}} & \multicolumn{2}{c||}{\textbf{NLE (\data{ECQA})}} & \multirow{2}{*}{\textbf{Model}} & \textbf{Preci} & \multicolumn{3}{c|}{\textbf{CFE (\data{IMDb})}} & \multicolumn{2}{c}{\textbf{NLE (\data{ECQA})}}\\
         
         & \textbf{-sion}  & LFR $\uparrow$ & PPL $\downarrow$ & TS $\downarrow$ & BARTScore $\uparrow$ & TIGERScore $\uparrow$ & & \textbf{-sion}  & LFR $\uparrow$ & PPL $\downarrow$ & TS $\downarrow$ & BARTScore $\uparrow$ & TIGERScore $\uparrow$\\

         \midrule

        % Row pair 1: Qwen2.5-7B & Llama3-8B
        \centering \multirow{6}{*}{\rotatebox[origin=r]{90}{\scriptsize{\lm{Qwen2.5-7B}}}} & \cellcolor[HTML]{F2CFC2}full
             & \cellcolor[HTML]{F2CFC2}87.75 & \cellcolor[HTML]{F2CFC2}34.77 & \cellcolor[HTML]{F2CFC2}0.47 & \cellcolor[HTML]{F2CFC2}-4.75 & \cellcolor[HTML]{F2CFC2}\textbf{-0.60}
        & \centering \multirow{6}{*}{\rotatebox[origin=l]{90}{\scriptsize{\lm{Llama3-8B}}}} & \cellcolor[HTML]{F2CFC2}full
             & \cellcolor[HTML]{F2CFC2}84.60 & \cellcolor[HTML]{F2CFC2}32.02 & \cellcolor[HTML]{F2CFC2}0.67 & \cellcolor[HTML]{F2CFC2}-4.44 & \cellcolor[HTML]{F2CFC2}\textbf{-0.73}\\
        & bib4
             & 87.20 & \underline{\textbf{34.11}} & 0.58 & -4.74 & -1.11
        & & bib4
             & 87.00 & 35.58 & \underline{\textbf{0.54}} & \underline{\textbf{-4.42}} & \underline{-0.75}\\
        & bib8
             & \underline{\textbf{88.20}} & 36.80 & 0.47 & \underline{\textbf{-4.71}} & -1.04
        & & bib8
            & 82.40 & \underline{\textbf{31.60}} & 0.68 & -4.43 & \underline{-0.75}\\
        & gptq4
             & 87.80 & 37.60 & 1.00 & -4.82 & -0.97
        & & gptq4
             & 76.40 & 33.57 & 0.77 & -4.55 & -1.07\\
        & gptq8
          & 86.27 & 39.46 & 1.00 & -4.82 & \underline{-0.94}
        & & gptq8
          & -- & -- & -- & -- & --\\
        & awq
          & 86.40 & 38.22 & \underline{\textbf{0.43}} & -4.76 & -1.28
        & & awq
          & 88.20 & 38.39 & \underline{\textbf{0.54}} & -4.45 & -0.87\\

        \midrule

        % Row pair 2: Qwen2.5-14B & Llama3-70B
         \centering \multirow{6}{*}{\rotatebox[origin=r]{90}{\scriptsize{\lm{Qwen2.5-14B}}}} & \cellcolor[HTML]{F2CFC2}full
           & \cellcolor[HTML]{F2CFC2}88.60 & \cellcolor[HTML]{F2CFC2}35.37 & \cellcolor[HTML]{F2CFC2}\textbf{0.47} & \cellcolor[HTML]{F2CFC2}-4.57 & \cellcolor[HTML]{F2CFC2}\textbf{-0.21}
        & \centering \multirow{6}{*}{\rotatebox[origin=l]{90}{\scriptsize{\lm{Llama3-70B}}}} & \cellcolor[HTML]{F2CFC2}full
             & \cellcolor[HTML]{F2CFC2}\textbf{89.29} & \cellcolor[HTML]{F2CFC2}\textbf{38.74} & \cellcolor[HTML]{F2CFC2}\textbf{0.21} & \cellcolor[HTML]{F2CFC2}\textbf{-4.37} & \cellcolor[HTML]{F2CFC2}\textbf{-0.15}\\
        & bib4
             & 88.80 & 35.19 & 0.62 & \underline{\textbf{-4.51}} & -0.33
        & & bib4
             & 86.00 & 46.67 & 0.30 & -4.46 & -1.36\\
        & bib8
             & 88.40 & 36.20 & 0.52 & -4.55 & -0.29
        & & bib8
             & \underline{87.01} & 48.23 & 0.33 & -4.39 & \underline{-0.27}\\
        & gptq4
         & \underline{\textbf{90.00}} & 31.06 & 1.00 & -4.61 & -0.30
        & & gptq4
             & 84.89 & 42.78 & \underline{0.24} & \underline{-4.38} & -0.76\\
        & gptq8
          & \underline{\textbf{90.00}} & \underline{\textbf{30.58}} & 1.00 & -4.59 & \underline{-0.28}
        & & gptq8
          & -- & -- & -- & -- & --\\
        & awq
          & 88.00 & 35.44 & \underline{0.51} & -4.57 & -0.44
        & & awq
          & 86.60 & \underline{40.43} & 0.31 & -4.41 & -0.55\\

         \midrule

        % Row pair 3: Qwen2.5-32B & Qwen2.5-72B
         \centering \multirow{6}{*}{\rotatebox[origin=r]{90}{\scriptsize{\lm{Qwen2.5-32B}}}} & \cellcolor[HTML]{F2CFC2}full
            & \cellcolor[HTML]{F2CFC2}89.92 & \cellcolor[HTML]{F2CFC2}\textbf{37.08} & \cellcolor[HTML]{F2CFC2}\textbf{0.25} & \cellcolor[HTML]{F2CFC2}-4.14 & \cellcolor[HTML]{F2CFC2}\textbf{-0.59}
        & \centering \multirow{6}{*}{\rotatebox[origin=r]{90}{\scriptsize{\lm{Qwen2.5-72B}}}} & \cellcolor[HTML]{F2CFC2}full
         & \cellcolor[HTML]{F2CFC2}\textbf{93.79} & \cellcolor[HTML]{F2CFC2}\textbf{34.59} & \cellcolor[HTML]{F2CFC2}\textbf{0.21} & \cellcolor[HTML]{F2CFC2}\textbf{-3.77} & \cellcolor[HTML]{F2CFC2}-0.41\\
        & bib4
             & 85.00 & \underline{37.23} & 0.38 & -4.29 & -0.98
        & & bib4
             & 83.80 & 39.95 & 0.23 & -3.94 & -0.58\\
        & bib8
             & 82.20 & 39.95 & \underline{0.28} & -4.21 & -0.76
        & & bib8
            & 82.20 & 40.15 & 0.22 & -3.86 & -0.51\\
        & gptq4
             & \underline{\textbf{90.00}} & 40.08 & 1.00 & -4.33 & \underline{-0.61}
        & & gptq4
            & \underline{90.00} & \underline{37.35} & 1.00 & \underline{\textbf{-3.77}} & -0.55\\
        & gptq8
          & \underline{\textbf{90.00}} & 42.15 & 1.00 & -4.22 & -1.32
        & & gptq8
          & \underline{90.00} & 39.82 & 1.00 & -3.82 & \underline{\textbf{-0.39}}\\
        & awq
          & 81.60 & 40.85 & \underline{0.28} & \underline{\textbf{-4.13}} & -0.74
        & & awq
          & 81.20 & 40.08 & \underline{0.21} & -4.13 & -0.43\\
        
        \toprule[1.5pt]
        \end{tabular}
    }
    \caption{Automatic evaluation results of CFEs and NLEs generated by \lm{Llama3} (8B, 70B) and \lm{Qwen2.5} (7B, 14B, 32B, 72B) models with full precision and different quantization methods. We use \texttt{FIZLE} on \data{IMDb}, evaluated by \textit{label flip rate (LFR)}, \textit{perplexity (PPL)}, and \textit{text similarity (TS)}; and \texttt{ZeroCoT} on \data{ECQA}, evaluated by \textit{BARTScore} and \textit{TIGERScore}. \textbf{Bold} values indicate the best-performing approach for each model, while \underline{underlined} values denote the best-performing quantization method.
    }
    \label{tab:automatic_evaluation_additional}
\end{table*}

%% file: table/user_study.tex
\begin{table}[t!]
    \centering
    \renewcommand*{\arraystretch}{1}
    
    \footnotesize
    \resizebox{\columnwidth}{!}{%
        \begin{tabular}{cc|cc|cc}

        \midrule
         \multirow{2}{*}{\textbf{Model}} & \multirow{2}{*}{\textbf{Metric}} & \multicolumn{2}{c|}{\textbf{NLE}} & \multicolumn{2}{c}{\textbf{CFE}}\\

          &  & \textbf{Trust.}  & \textbf{Cohere.}  &\textbf{ Trust.}  & \textbf{Cohere.}  \\

         \midrule

        \multirow{4}{*}[0.25em]{\rotatebox[origin=c]{90}{\tiny{\lm{Qwen2.5-7B}}}} & \cellcolor[HTML]{F2CFC2}full
           & \cellcolor[HTML]{F2CFC2}\textbf{3.47} & \cellcolor[HTML]{F2CFC2}\textbf{3.22} & \cellcolor[HTML]{F2CFC2}\textbf{3.90} & \cellcolor[HTML]{F2CFC2}\textbf{3.40} \\
        & gptq4
           & 3.06 & 3.14  & 3.20 & 2.84 \\
        & gptq8
        & 2.96 & 2.63 & 3.26 & 2.86  \\
        & awq
        & 2.98 & 2.96  & 
        2.69 & 2.49  \\

        \midrule

         \multirow{4}{*}[0.45em]{\rotatebox[origin=c]{90}{\tiny{\lm{Qwen2.5-32B}}}} & \cellcolor[HTML]{F2CFC2}full
           & \cellcolor[HTML]{F2CFC2}3.41 & \cellcolor[HTML]{F2CFC2}3.41 & \cellcolor[HTML]{F2CFC2}\textbf{3.50} & \cellcolor[HTML]{F2CFC2}3.21  \\
        & gptq4
           & 3.32 & \textbf{3.60}  & 3.25 & 2.96   \\
        & gptq8
        & \textbf{3.51} & 3.00 & 3.45 & \textbf{3.26}  \\
        & awq
         & 2.55 & 2.47 & 3.26 & 3.26 \\

        \midrule
        
         \multirow{4}{*}[0.45em]{\rotatebox[origin=c]{90}{\tiny{\lm{Qwen2.5-72B}}}} & \cellcolor[HTML]{F2CFC2}full
           & \cellcolor[HTML]{F2CFC2}\textbf{3.38} & \cellcolor[HTML]{F2CFC2}3.50 &  \cellcolor[HTML]{F2CFC2}\textbf{4.30} & \cellcolor[HTML]{F2CFC2}\textbf{4.30}   \\
        & gptq4
           & 3.21 & 3.23 & 3.04 & 2.61   \\
        & gptq8
         & 2.97 & 2.86 & 2.92 & 3.02 \\
        & awq
       & 2.92 & \textbf{3.71} & 3.40 & 3.37  \\     
        
        \bottomrule
        \end{tabular}
    }
    \caption{User study results for generated NLEs and CFEs on \data{eSNLI}, evaluated based on Trustworthiness (Trust.) and Coherence (Cohere.).
    }
    \label{tab:user_study}
\end{table}

%% file: table/judge.tex
\begin{table*}[t!]
    \centering
    \renewcommand*{\arraystretch}{1}
    
    \footnotesize
    \resizebox{\textwidth}{!}{%
        \begin{tabular}{cc|ccc|ccc|ccc|ccc|}

        \midrule
         \textbf{Model} & \textbf{Precision} & \multicolumn{6}{c|}{\textbf{NLE}} & \multicolumn{6}{c|}{\textbf{CFE}}\\
         
          \multicolumn{2}{c|}{\textbf{Metric}} & \multicolumn{3}{c}{\textbf{Trustworthiness}}  & \multicolumn{3}{c|}{\textbf{Coherence}}  & \multicolumn{3}{c}{\textbf{Trustworthiness}}  & \multicolumn{3}{c|}{\textbf{Coherence}}  \\

          \midrule

          \multicolumn{2}{c|}{\textbf{Judge Model}} & DS & OSS & Gemma & DS & OSS & Gemma & DS & OSS & Gemma & DS & OSS & Gemma \\

         \midrule

        \multirow{4}{*}[0.25em]{\rotatebox[origin=c]{90}{\tiny{\lm{Qwen2.5-7B}}}} & \cellcolor[HTML]{F2CFC2}full & 4.93 & 4.67 & 3.67 & 5.00 & 4.86 & 4.87 & 2.73 & 1.73 & 2.40 & 3.47 & 2.67 & 3.13 \\

        & gptq4
           & 4.73 & 4.00 & 3.53 & 4.86 & 4.13 & 4.67 & 2.87 & 1.73 & 2.60 & 3.93 & 2.67 & 3.33 \\
        & gptq8
        & 4.93 & 4.53 & 3.67 & 5.00 & 4.67 & 4.80 & 3.40 & 1.93 & 2.87 & 3.93 & 2.80 & 3.73  \\
        & awq
        & 5.00 & 4.80 & 3.60 & 5.00 & 5.00 & 4.87 & 2.79 & 1.67 & 2.87 & 3.93 & 2.87 & 3.60 \\

        \midrule

         \multirow{4}{*}[0.45em]{\rotatebox[origin=c]{90}{\tiny{\lm{Qwen2.5-32B}}}} & \cellcolor[HTML]{F2CFC2}full
           & 4.86 & 4.67 & 3.53 & 4.53 & 4.67 & 4.67 & 2.46 & 2.87 & 2.46 & 3.60 & 3.73 & 3.20 \\
        & gptq4
           & 5.00 & 4.87 & 3.73 & 5.00 & 4.93 & 4.86 & 2.67 & 1.60 & 2.46 & 3.67 & 2.67 & 3.33  \\
        & gptq8
        & 4.73 & 4.07 & 3.80 & 4.47 & 4.20 & 4.47 & 1.87 & 1.53 & 2.33 & 3.00 & 2.73 & 3.27 \\
        & awq
         & 4.21 & 3.87 & 3.60 & 4.07 & 3.73 & 4.67 & 2.40 & 1.60 & 2.40 & 3.53 & 2.53 & 3.33\\

        \midrule
        
         \multirow{4}{*}[0.45em]{\rotatebox[origin=c]{90}{\tiny{\lm{Qwen2.5-72B}}}} & \cellcolor[HTML]{F2CFC2}full & 5.00 & 5.00 & 3.67 & 4.93 & 4.80 & 4.87 & 2.20 & 1.67 & 2.66 & 3.20 & 2.93 & 3.47  \\
        & gptq4
           & 5.00 & 4.87 & 3.47 & 5.00 & 4.80 & 4.80 & 2.46 & 1.73 & 2.66 & 4.13 & 2.93 & 3.53 \\
        & gptq8
         & 5.00 & 4.80 & 3.60 & 5.00 & 4.80 & 4.80 & 2.73 & 1.73 & 2.53 & 3.46 & 2.67 & 3.47 \\
        & awq
       & 4.93 & 4.73 & 3.60 & 5.00 & 4.67 & 4.73 & 3.27 & 1.67 & 2.73 & 4.00 & 2.67 & 3.53\\     
        
        \bottomrule
        \end{tabular}
    }
    \caption{LLM-as-a-Judge evaluation on data examples selected for the user study using \lm{DeepSeek-R1} (DS), \lm{GPT-OSS-120B} (OSS), and \lm{Gemma3-27B} (Gemma).
    }
    \label{tab:judge}
\end{table*}

%% file: table/task_performance.tex
\begin{table*}[t!]
    \centering
    \setlength{\extrarowheight}{3pt}
    
    \footnotesize
    \begin{tabular}{c|cccc|c|cccc}
        \toprule[1.5pt]
        \textbf{Model} & \textbf{Precision} & \textbf{\data{eSNLI}} & \textbf{\data{HealthFC}} & & \textbf{Model} & \textbf{Precision} & \textbf{\data{eSNLI}} & \textbf{\data{HealthFC}} \\
        \midrule
        \centering \multirow{6}{*}{\rotatebox[origin=r]{90}{\lm{Qwen2.5-7B}}} 
            & \cellcolor[HTML]{F2CFC2}full & \cellcolor[HTML]{F2CFC2}\textbf{87.40}\% & \cellcolor[HTML]{F2CFC2}\textbf{51.43\%} 
            & & \centering \multirow{6}{*}{\rotatebox[origin=r]{90}{\lm{Qwen2.5-14B}}} 
            & \cellcolor[HTML]{F2CFC2}full & \cellcolor[HTML]{F2CFC2}69.50\% & \cellcolor[HTML]{F2CFC2}\textbf{45.14\%} \\
        & bib4 & 84.80\% & 36.57\% 
            & & & bib4 & 63.90\% & 40.29\% \\
        & bib8 & 86.70\% & 36.57\% 
            & & & bib8 & 67.70\% & 40.29\% \\
        & gptq4 & 86.80\% & 38.86\% 
            & & & gptq4 & \textbf{\underline{78.50}\%} & 36.57\% \\
        & gptq8 & \underline{87.30}\% & \underline{39.71\%} 
            & & & gptq8 & 67.90\% & 39.42\% \\
        & awq & 86.10\% & 36.86\% 
            & & & awq & 70.00\% & \underline{46.29\%} \\
        
        \midrule
        
        \centering \multirow{6}{*}{\rotatebox[origin=r]{90}{\lm{Qwen2.5-32B}}} 
            & \cellcolor[HTML]{F2CFC2}full & \cellcolor[HTML]{F2CFC2}88.40\% & \cellcolor[HTML]{F2CFC2}42.57\% 
            & & \centering \multirow{6}{*}{\rotatebox[origin=r]{90}{\lm{Qwen2.5-72B}}} 
            & \cellcolor[HTML]{F2CFC2}full & \cellcolor[HTML]{F2CFC2}78.80\% & \cellcolor[HTML]{F2CFC2}51.43\% \\
        & bib4 & 89.50\% & 42.86\% 
            & & & bib4 & \textbf{\underline{82.00}\%} & 53.71\% \\
        & bib8 & 88.10\% & \textbf{\underline{47.43\%}} 
            & & & bib8 & 78.90\% & \textbf{\underline{54.00\%}} \\
        & gptq4 & 87.90\% & 41.43\% 
            & & & gptq4 & 78.00\% & 53.42\% \\
        & gptq8 & 88.60\% & 42.29\% 
            & & & gptq8 & 78.10\% & 52.09\% \\
        & awq & \textbf{\underline{90.60}\%} & 39.43\% 
            & & & awq & 77.70\% & 52.28\% \\

        \midrule

        \centering \multirow{5}{*}{\rotatebox[origin=r]{90}{\lm{Llama3-8B}}} & \cellcolor[HTML]{F2CFC2}full & \cellcolor[HTML]{F2CFC2}34.53\% & \cellcolor[HTML]{F2CFC2}59.43\% & & \centering \multirow{5}{*}{\rotatebox[origin=r]{90}{\lm{Llama3-70B}}} & \cellcolor[HTML]{F2CFC2}full & \cellcolor[HTML]{F2CFC2}60.46\% & \cellcolor[HTML]{F2CFC2}\textbf{67.14\%} \\
        
        & bib4 & \textbf{\underline{37.44\%}}& 39.43\% & & & bib4 & 62.36\% & 63.71\%\\
        & bib8 & 33.73\% & 59.43\% & & & bib8 & 38.44\% & 58.29\% \\
        & gptq4 & 27.33\% & \textbf{\underline{63.14\%}} & & & gptq4 & \textbf{\underline{64.26\%}} & 61.71\% \\
        & awq & 32.03\% & 62.00\% & & & awq & 64.26\% & \underline{66.00\%} \\

        \toprule[1.5pt]
    \end{tabular}
    
    \caption{Task performance of all deployed models with different data type precisions on \data{eSNLI} and \data{HealthFC}.}
    \label{tab:task_performance}
\end{table*}

%% file: table/performance-quality.tex
\begin{table}[t!]
    \centering
    \setlength{\extrarowheight}{3pt}
        \begin{tabular}{c|cc}
            \toprule[1.5pt]
            \textbf{Model} & \textbf{\data{eSNLI}} & \textbf{\data{HealthFC}}\\

            \midrule

            \lm{Qwen2.5-7B} & 0.13 & -0.01\\
            \lm{Qwen2.5-14B} & -0.15 & 0.04 \\
            \lm{Qwen2.5-32B} & -0.16 & 0.08\\
            \lm{Qwen2.5-72B} & 0.13 & 0.03\\
            \lm{Llama3-8B} & 0.02 & 0.05\\
            \lm{Llama3-70B} & 0.11 & -0.11\\
            
            \toprule[1.5pt]
        \end{tabular}
    % }
    \caption{Spearman correlation between the task performance and natural language explanation quality.}
    \label{tab:quality_performance}
\end{table}

%% file: table/biasing_transitions.tex
\begin{table*}[t!]
    \centering
    \setlength{\extrarowheight}{3pt}
    % \resizebox{\textwidth}{!}{%
        \begin{tabular}{l|cccc}
            \toprule[1.5pt]
            \textbf{Quantization} &
            \textbf{$\mathrm{F}\!\to\!\mathrm{N}$ (Degrade)} &
            \textbf{$\mathrm{F}\!\to\!\mathrm{F}$ (Maintain)} &
            \textbf{$\mathrm{N}\!\to\!\mathrm{N}$ (Maintain)} &
            \textbf{$\mathrm{N}\!\to\!\mathrm{F}$ (Improve)} \\
            \midrule
            AWQ  &  7.11\% & 80.78\% & 7.20\% & 4.92\% \\
            GPTQ4 & 16.88\% & 70.32\% & 7.76\% & 5.05\% \\
            GPTQ8 &  4.57\% & 83.54\% & 8.52\% & 3.37\% \\
            bib4  &  8.20\% & 79.91\% & 5.52\% & 6.37\% \\
            bib8  &  5.57\% & 82.54\% & 8.24\% & 3.65\% \\
            \midrule
            \textbf{Average} & \textbf{8.47\%} & \textbf{79.42\%} & \textbf{7.45\%} & \textbf{4.67\%} \\
            \toprule[1.5pt]
        \end{tabular}
    % }
    \caption{Faithfulness transition rates (in \%) measured by biasing features.}
    \label{tab:biasing-transitions}
\end{table*}